\documentclass[12pt]{report}
\usepackage[utf8]{inputenc}
\usepackage{amsmath}
\usepackage{graphicx}
\graphicspath{ {./images/} }
\usepackage[font=footnotesize,labelfont=bf]{caption}
\usepackage[font=footnotesize]{subcaption}
\usepackage{setspace}

\usepackage[a4paper, margin=3cm, top=2.5cm, bottom=2.5cm]{geometry}
\usepackage[table,xcdraw]{xcolor}
\usepackage{color}
\usepackage[all]{nowidow}

\usepackage[colorinlistoftodos]{todonotes}
\usepackage{pdflscape}
\usepackage{afterpage}
\usepackage{pdfpages}

\usepackage[style=ieee,backref=true]{biblatex}
\usepackage[hidelinks]{hyperref}
\usepackage{nameref}
\usepackage[nottoc]{tocbibind}
\usepackage{csquotes}
\usepackage{multirow}

\usepackage[acronym, toc]{glossaries}
\makenoidxglossaries % use TeX to sort
\setacronymstyle{long-short}
\newacronym{SIDE}{SIDE}{single image depth estimation}
\newacronym{SSIDE}{SSIDE}{semantic single image depth estimation}
\newacronym{LIDAR}{LIDAR}{light detection and ranging}
\newacronym{RADAR}{RADAR}{radio detection and ranging}
\newacronym{SONAR}{SONAR}{sound navigation and ranging}
\newacronym{IMU}{IMU}{inertial measurement unit}
\newacronym{GPS}{GPS}{global positioning system}
\newacronym{FCRN}{FCRN}{fully convolutional residual network}
\newacronym{ReLU}{ReLU}{rectified linear unit}
\newacronym{IoU}{IoU}{intersection over union}
\newacronym{AFOV}{AFOV}{angular field of view}

\newacronym{MAPE}{MAPE}{mean absolute percentage error}
\newacronym{MSPE}{MSPE}{mean square percentage error}
\newacronym{RMSE}{RMSE}{root mean square error}
\newacronym{RMSElog}{RMSElog}{root mean square error in log space}
\newacronym{SILog}{SILog}{scale-invariant error}

\newacronym{MRF}{MRF}{markov random field}
\newacronym{CRF}{CRF}{conditional random field}
\newacronym{SfM}{SfM}{structure from motion}
\newacronym{CNN}{CNN}{convolutional neural networks}

\newglossarystyle{mystyle}{%
  \setglossarystyle{long}%
     {\begin{longtable}[l]{p{4cm}p{\glsdescwidth}}}%
     {\end{longtable}}%
}
\begin{document}

    \pagenumbering{gobble}

\begin{titlepage}
    \centering
    ~\\~\\
    \includegraphics[scale = 1.5]{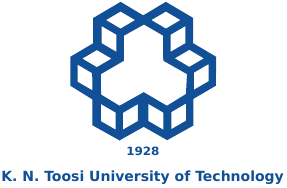}\\
	\small Faculty of Electrical Engineering\\~\\~\\~\\
	\large A Master’s Thesis in Mechatronics Engineering\\
	\LARGE Predicting Depth from Semantic Segmentation using 
	Game Engine Dataset\\~\\~\\~\\
	\small by\\
	\large Mohammad Amin Kashi\\~\\
	\small Supervisor\\
	\large Hamid D. Taghirad\\~\\~\\
	\small Summer 2020
\end{titlepage}

\begin{figure}[h]
    \centering
    \includegraphics[width=\linewidth]{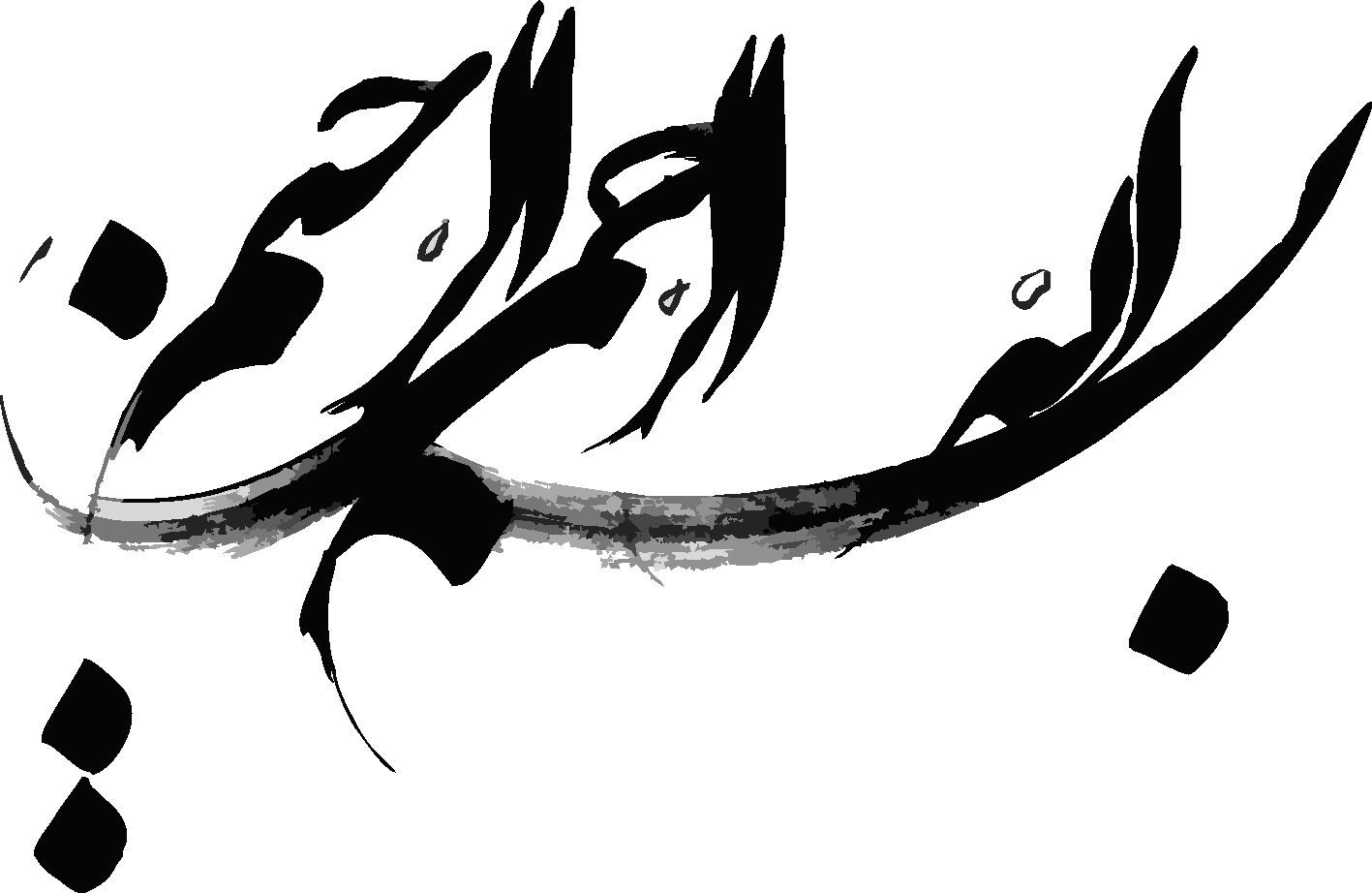}
	\end{figure}
\newpage



بسمه تعالی

| | تأییدیه هیأت داوران | 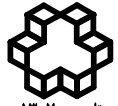 |
|---|---|---|
| شماره:<br>تاریخ: | | تاسیس ۱۳۰۷<br>دانشگاه صنعتی خواجه نصیرالدین طوسی |

هیأت داوران پس از مطالعه پایان‌نامه و شرکت در جلسه دفاع از پایان‌نامه تهیه شده تحت عنوان :

**پیش بینی عمق از تقسیم‌بندی معنایی با استفاده از داده‌های موتور بازی سازی (Predicting Depth from Semantic Segmentation using Game Engine Dataset)**

توسط آقای **محمد امین کاشی** ، صحت و کفایت تحقیق انجام شده را برای اخذ درجه کارشناسی ارشد رشته **مکاترونیک** در تاریخ ۱۳۹۹/۶/۲۴ مورد تأیید قرار می‌دهند.

| | امضاء | جناب آقای دکتر حمید رضا تقی راد | ۱- استاد راهنمای اول |
|---|---|---|---|
| | امضاء | - | ۲- استاد راهنمای دوم |
| | امضاء | - | ۳- استاد مشاور |
| | امضاء | جناب آقای دکتر مهدی علیاری | ۴- ممتحن داخلی |
| | امضاء | جناب سرکار خانم دکتر شهره کسائی | ۵- ممتحن خارجی |
| | امضاء | جناب آقای دکتر مهدی علیاری | ۶- نماینده تحصیلات تکمیلی دانشکده |

9/10/1387-4216

| :شماره | اظهارنامه دانشجو | 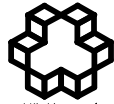 |
|---|---|---|
| :تاریخ | | تاسیس ۱۳۰۷<br>دانشگاه صنعتی خواجه نصیرالدین طوسی |

اینجانب **محمد امین کاشی** دانشجوی کارشناسی‌ارشد رشته **مکاترونیک** دانشکده **برق** دانشگاه صنعتی خواجه نصیرالدین طوسی گواهی می‌نمایم که تحقیقات ارائه شده در پایان‌نامه با عنوان

بیش بینی عمق از تقسیم‌بندی معنایی با استفاده از داده‌های موتور بازی سازی (Predicting Depth from Semantic Segmentation using Game Engine Dataset)

با راهنمایی استاد محترم جناب آقای دکتر حمید رضا تقی راد، توسط شخص اینجانب انجام شده و صحت واصالت مطالب نگارش شده در این پایان‌نامه مورد تأیید می‌باشد، و در مورد استفاده از کار دیگر محققان به مرجع مورد استفاده اشاره شده است. بعلاوه گواهی می‌نمایم که مطالب مندرج در پایان نامه تا کنون برای دریافت هیچ نوع مدرک یا امتیازی توسط اینجانب یا فرد دیگری در هیچ جا ارائه نشده است و در تدوین متن پایان‌نامه چارچوب (فرمت) مصوب دانشگاه را بطور کامل رعایت کرده‌ام.

امضاء دانشجو:

تاریخ:
1399/7/20

۲۹/۱۰/۱۳۸۷-۴۲۱۷

بسمه تعالی

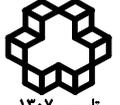

| شماره : | عنوان فرم | |
|---|---|---|
| تاریخ : | حق طبع و نشر و مالکیت نتایج | |

اینجانب **محمد امین کاشی** دانشــجوی دوره **کارشناســی ارشد** مهندسیِ **مکاترونیک** دانشــکده **بـرق** دانشـگاه صـنعتی خواجه نصیر الدین طوسی گواهی می‌نمایم که تحقیقات ارائه شده در این پایان نامه توسط شخص اینجانب انجام شـــده و صحت و اصالت نگارش شده مورد تایید می‌باشد و در موارد استفاده از کار دیگر محققان به مرجـع مـورد اسـتفاده اشـاره شده است. به علاوه گواهی می‌نمایم که مطالب مندرج در پایان نامه تاکنون برای دریـافت هیچ نـوع مـدرک یـا امتیـازی توسط اینجانب یا فرد دیگری در هیچ جا ارائه نشده است و در تدوین متن پایان نامه چارچوب مصــوب دانشگاه را بطــور کامل رعایت کرده‌ام.

کلیه حقوق مادی و معنوی این اثر فقط متعلق به دانشگاه صنعتی خواجه نصیرالدین طوسـی می‌باشــد و بـدون اجـازه کتبی دانشگاه قابل واگذاری وبهره برداری نیست. همچنین استفاده از اطلاعات و نتایج موجود در پایـان نامـه بـدون ذکـر مرجع مجاز نمی باشد.

امضا دانشجو:

تاریخ:
1399/7/20



\chapter*{Acknowledgements}

I would like to express my gratitude to my supervisor Prof. Hamid D. Taghirad for the useful comments, remarks, and engagement through the learning process of this master thesis. I would also like to thank Advanced Robotics and Automated System (ARAS) team members for their comments on my work. Furthermore, I would like to thank my friends, specially Erfan Khaniki, for their encouragement during the research process.

Finally, I must express my very profound gratitude to my parents, my brother, and my sister for providing me with unfailing support and continuous encouragement throughout my years of study and through the process of researching and writing this thesis. This accomplishment would not have been possible without them. Thank you.
\newline
\newline
\newline
\newline
Tehran, September 26, 2020
\newline
Mohammad Amin Kashi

    \chapter*{Abstract}
\label{sec:engAbstract}

Depth perception is fundamental for robots to understand the surrounding environment. As the view of cognitive neuroscience, visual depth perception methods are divided into three categories, namely binocular, active, and pictorial. The first two categories have been studied for decades in detail. However, research for the exploration of the third category is still in its infancy and has got momentum by the advent of deep learning methods in recent years.
In cognitive neuroscience, it is known that pictorial depth perception mechanisms are dependent on the perception of seen objects. Inspired by this fact, in this thesis, we investigated the relation of perception of objects and depth estimation convolutional neural networks. For this purpose, we developed new network structures based on a simple depth estimation network that only used a single image at its input. Our proposed structures use both an image and a semantic label of the image as their input. We used semantic labels as the output of object perception. The obtained results of performance comparison between the developed network and original network showed that our novel structures can improve the performance of depth estimation by 52\% of relative error of distance in the examined cases.
Most of the experimental studies were carried out on synthetic datasets that were generated by game engines to isolate the performance comparison from the effect of inaccurate depth and semantic labels of non-synthetic datasets. It is shown that particular synthetic datasets may be used for training of depth networks in cases that an appropriate dataset is not available. Furthermore, we showed that in these cases, usage of semantic labels improves the robustness of the network against domain shift from synthetic training data to non-synthetic test data.

\textbf{Keywords}: Depth estimation, pictorial depth cues, semantic segmentation, synthetic dataset, deep convolutional neural network.

    \clearpage
    \pagenumbering{roman}
    \tableofcontents
    \listoffigures
    \listoftables
    \printnoidxglossary[type=acronym, nonumberlist, title=Symbols and Acronyms, style=mystyle, nogroupskip=true]
    \clearpage
    \pagenumbering{arabic} % restart page numbers at one, now in arabic style
    \chapter{Introduction}
\section{Preface}
\label{ch:introduction}

%=============================== scope of the problem ===============================

%===why the overall subject area of the research is important.
Depth perception is fundamental for robots to understand the surrounding environment. This perception guides robots to know where the objects in their environment are. Various sensors including \gls{SONAR}, \gls{LIDAR}, \gls{RADAR}, and camera can be used to accommodate a robot with depth perception. Among these sensors, cameras are the cheapest and widely available in various locations such as factories, streets, shopping centers and systems such as mobile phones, cars, drones, surveillance camera systems. Furthermore, cameras are used for different purposes such as security, traffic monitoring, mapping, retail analysis, and industrial automation. Therefore, images of cameras may be widely used for depth estimation in different applications.

%===Review of the pertinent literature to orient the reader. Identification of the gap in the literature that the current research was intended to address.
Due to its importance, depth estimation from images has received a large amount of attention in the literature. These researches are inspired by studying the mechanisms underlying human vision, as the whole computer vision is inspired by the human vision. Therefore, proposed algorithms may be distinguished into the same categories as that of the depth perception mechanisms of human vision.

\section{Taxonomy of Depth Estimation Algorithms}
In cognitive neuroscience, different sources of visual information about distance and depth are called depth cues and are divided into three kinds. The first and second kinds are called binocular cues and active cues, respectively. The third kind is called pictorial cues, which are used for \gls{SIDE}\footnote{This classification is borrowed from \cite{brenner_depth_2018}. In another classification depth cues are divided into three kinds of oculomotor cues, monocular cues, and binocular cues \cite{goldstein_sensation_2016}.}. Using this classification, mechanisms of depth perception may be classified by their source of information.\footnote{In some papers, the term \emph{monocular depth estimation} was used to describe the estimation of depth from pictorial cues. This designation is ambiguous because \emph{monocular depth estimation} also applies to depth estimation from active cues. Therefore, only the term \gls{SIDE} is considered in this thesis.}

\subsection{Binocular Depth Cues}
Binocular cues include image disparity and orientation of optical sensors. Since optical sensors are fixed relative to each other in a stereo camera, the later cue has not been investigated in computer vision. In contrast, the binocular disparity has been extremely studied (for reviews of works done from the mid-70's up to 2020 refer to \cites{barnard_computational_1982,dhond_structure_1989, scharstein_taxonomy_2002, zhou_review_2020}). Since the relation of disparity map and depth map is known, in many of the proposed algorithms only estimation of disparity map is addressed.

\subsection{Active Depth Cues}
Active cues include accommodation information and motion parallax information. Several works proposed algorithms that use accommodation information to estimate the depth of a scene (e.g. \cites{kim_video_2016, hazirbas_deep_2018, maximov_focus_2020}). For this purpose, a sequence of images is captured from the scene, while the amount of accommodation is changed. Relative to accommodation information, motion parallax information has a more active research area (for reviews of related works refer to \cites{triggs_bundle_2000, oliensis_critique_2000, hartley_multiple_2003, ozyesil_survey_2017}). Depth estimation algorithms related to motion parallax are called \gls{SfM}. Motion parallax is a special case of parallax\footnote{Difference or change in the apparent position or direction of an object as seen from two different points \cite{noauthor_parallax_nodate}} in which different views are obtained by motion at different moments. Another example of parallax occurs when a binocular vision system is used, while different views are obtained by two eyes or cameras at the same moment.

\subsection{Pictorial Depth Cues}
Between the three kinds of depth cues, the third kind contains the highest number of cues. Pictorial cues are:

\begin{itemize}
    \item \label{itm:PictorialCue1} Occlusion: Nearby objects occlude farther objects.
    \item \label{itm:PictorialCue2} Relative position: By knowing the relative position of an object to another object that its depth is known, depth estimation of the first object becomes possible.
    \item \label{itm:PictorialCue3} Size of objects: Image size of a specific object varies with distance.
    \item \label{itm:PictorialCue4} Density of textures' gradient: It varies with distance.
    \item \label{itm:PictorialCue5} Image quality: Contrast of an image decreases with distance because of light diffusion by particles in the air.
\end{itemize}
Until the last decade, few studies investigated the depth estimation using pictorial cues. With the widespread use of deep learning methods, more and more researches have been focused on pictorial cues.

Some works tried to estimate the depth from a specific cue in special situations. \citeauthor*{aslantas_depth_2007} proposed a numerical model that estimates the depth of an object, provided the defocused image and the sharp image of the object are available \cite{aslantas_depth_2007}. The parameters of the model should be recalculated for different cameras. In \cite{joglekar_depth_2011} the \emph{relative position} cue is used to estimate the distance of a vehicle on a road. The proposed method first detects the contact point of the vehicle on the road, then by assuming a planar road and a known camera height and using geometry rules the distance of the contact point is calculated and is reported as the distance of the vehicle. Another representative research on this area is the use of a haze removal method as a tool for estimating the depth of a scene \cite{ke_wang_combining_2014}.

By imposing the burden of extracting the pictorial cues on \gls{CNN}, the majority of works are focused to directly estimate the depth. \citeauthor*{eigen_depth_2014} proposed the first successful \gls{CNN} to estimate depth from a single image \cite{eigen_depth_2014}. They designed a two-stage architecture; the first stage makes a coarse and global prediction, while the second stage enhances this prediction locally. They also used a scale-invariant loss function to train the proposed architecture regardless of scale. This loss function becomes a standard metric in performance evaluation of depth estimation, e.g. in KITTI depth benchmark \cite{geiger_vision_2013}. The later works tried to propose better models in terms of performance by using novel architectures \cites{laina_deeper_2016, fu_deep_2018, ren_deep_2019, lee_big_2019}, new cost functions \cites{yin_enforcing_2019, diaz_soft_2019, li_monocular_2018, liebel_multidepth_2019}, or multi-task architectures \cites{peng_wang_towards_2015, ji_joint_2016,jiao_look_2018, ochs_sdnet_2019, zama_ramirez_geometry_2019, zhang_pattern-affinitive_2019, lin_depth_2019, chen_towards_2019, kwak_novel_2020}.

\section{Obtaining Pictorial Cues}
Although various works addressed the problem of estimating depth from pictorial cues for \gls{SIDE}, not enough attention has been paid to the relation of pictorial cues and depth estimation. In most of the proposed methods, a \gls{CNN} is responsible for extracting the depth cues implicitly and estimating depth from those cues; however, most of the pictorial cues are high-level abstract concepts and logically need a separate \gls{CNN} to extract them with high accuracy. For example, the recognition of the occlusion between two objects already needs the recognition of the two objects, as the determination of the position of one object relative to another object needs the recognition of the two objects. In a similar way, the size determination of an object already needs the recognition of that object, and the determination of density of a texture needs that the texture be detected first.

As mentioned above a common prerequisite for obtaining pictorial cues is to detect and locate different objects and textures in an image; a task that is somewhat similar to semantic segmentation. For the same reason the most of the proposed multi-task \gls{CNN}s (or \gls{CRF}) were designed to estimate depth and simultaneously segment the input image semantically \cites{peng_wang_towards_2015, ji_joint_2016,jiao_look_2018, ochs_sdnet_2019, zama_ramirez_geometry_2019, zhang_pattern-affinitive_2019, lin_depth_2019, chen_towards_2019, kwak_novel_2020}. Simultaneous learning enables these models to better extract pictorial cues and estimate depth. However, it is plausible that a large aspect of the information, that semantic segments could have delivered to the model, is not used. As stated earlier, semantic segmentation is a prerequisite for obtaining pictorial depth cues; therefore, high-level abstract concepts can not be obtained from the semantic segments in these multi-task models, because the semantic segments are an output of the model and can not be processed anymore.

In some other works, semantic segments were used in other various methods. \citeauthor*{schneider_semantically_2016} used the edges of semantic segments as a guide for up-sampling of sparse depth data of \gls{LIDAR} \cite{schneider_semantically_2016}. \citeauthor*{guizilini_semantically-guided_2020} used a pre-designed and pre-trained segmentation model in parallel with a pre-designed depth estimation model instead of designing a multi-task \gls{CNN}. During training and test of the depth model, the segmentation model injects low-level and high-level features from different layers of its network to the corresponding layers of the depth model \cite{guizilini_semantically-guided_2020}. \citeauthor*{liu_single_2010} proposed the first depth estimation model that uses predicted semantic segments as its input. They used an \gls{MRF} for semantic segmentation and an \gls{MRF} for predicting depth from both images and semantic segments \cite{liu_single_2010}. Furthermore, \citeauthor*{jafari_analyzing_2017} analyzed the mutual impact of depth and semantic segments in refining each other. They designed a \gls{CNN} that uses an estimated depth map and a semantic segment map as its inputs and jointly refines them \cite{jafari_analyzing_2017}.

%===Objective of the research. The hypotheses or research questions that the study addressed.
\section{Research Goals, Methods, and Results}
With the assumption that semantic segmentation is a prerequisite for obtaining pictorial depth cues, this thesis investigates how the use of semantic segments as a raw data, like an input image, can potentially improve the performance of a \gls{SIDE} \gls{CNN}, and if there is an improvement, how significant it is. This thesis also investigates the effect of the use of semantic segments on the robustness of a \gls{SIDE} model against domain shift.

%===Method of the investigation. If deemed necessary, the reasons for the choice of a particular method should be briefly stated.
%===end the introduction by stating the principal results of the investigation and the principal conclusions suggested by the results.
\textbf{Experimental Study on Estimation of Depth from Semantic Labels:} For the first goal of the thesis, a \gls{SIDE} \gls{CNN} is chosen to be empowered by semantic segments as its input, to drive an \gls{SSIDE} \gls{CNN}\footnote{From here onwards, models that only use an image as their input for depth estimation are referred to as \gls{SIDE} models, and models that use an image and a semantic label as their inputs for depth estimation are referred to as \gls{SSIDE} models}. After the selection of a \gls{SIDE} \gls{CNN}, several similar networks are developed from it. These networks use semantic segments along with an image as inputs, in contrast to the original network that only uses an image as the input. Furthermore, the structures of the derived networks have some differences from the first network. The differences are formed to facilitate the utilization of extra data in semantic segments.

For the evaluation of performances, the driven \gls{SSIDE} \gls{CNN}s are compared with the original \gls{SIDE} \gls{CNN} on estimating depth on specific datasets. In fact, all of the experiment is carried out on a synthetic dataset to eliminate the altering effect of noisy and sparse depth labels and inaccurate semantic segments of non-synthetic datasets. At the end, some parts of the experiment are repeated on two other datasets, one synthetic and another non-synthetic, as described in \autoref{sec:firts_exp_method}. The results show that the \gls{SSIDE} \gls{CNN}s outperform the \gls{SIDE} \gls{CNN} by 18 percent of the absolute relative error in the examined situations (For complete results see \autoref{sec:firts_exp_results}).

\textbf{Experimental Study on Robustness of SSIDE against Domain Shift:} For the second objective of this thesis, an \gls{SSIDE} model and a \gls{SIDE} model are trained on a synthetic dataset (a combination of multiple synthetic datasets), and two other \gls{SIDE} models are trained on two different non-synthetic datasets. Note that all \gls{SIDE} models have the same architecture. All models are tested on the test-set of the two non-synthetic datasets, as described in \autoref{sec:third_exp_method}. Results show that the \gls{SSIDE} \gls{CNN} performs like the two \gls{SIDE} models that were trained on non-synthetic datasets, while the \gls{SIDE} model that was trained on the synthetic dataset has a poor performance relative to others. In other words, use of semantic segments improves the robustness of the model against domain shift (For complete results see \autoref{sec:third_exp_results}.).

\textbf{Experimental Study on Use of Synthetic Data instead of Non-Synthetic Data:} An \gls{SSIDE} model needs both an image and semantic segments to estimate depth. The used datasets in the \emph{first experimental study} include all needed data: image, semantic segments, and depth. The non-synthetic datasets in the \emph{final experimental study} do not have semantic labels. To provide semantic segments for these datasets, one way is to train a segmentation model on a synthetic dataset and generate segmentation for these datasets. To ensure that synthetic datasets are capable of training a segmentation model to generate accurate semantic segments for non-synthetic datasets, a supplementary experimental study was carried out before the \emph{final experimental study}. In this experimental study, multiple segmentation models with the same architecture are trained on different synthetic and non-synthetic datasets. Then all models are tested on non-synthetic datasets as described in \autoref{sec:second_exp_method}. The results show that some of synthetic datasets can be more informative than some of non-synthetic datasets, i.e., the model that was trained on a synthetic dataset performed better than a model that was trained on a non-synthetic dataset (For complete results see \autoref{sec:second_exp_results}.).

\section{Datasets}
Due to the abundance and diversity of self-driving cars' datasets, this group of datasets were mainly used in this thesis. These datasets were captured with different sensors (RGB camera, \gls{LIDAR}, \gls{RADAR}, \gls{IMU}, and \gls{GPS}) and contain different data (RGB image, depth data, semantic segments, instance segments, ego velocity, ego position, optical flow) and were captured in different weather conditions and different cities. These datasets are available in non-synthetic types and synthetic types as well. However, in the \emph{first experimental study}, the non-synthetic \emph{NYU Depth Dataset V2} \cite{silberman_indoor_2012} was used too.

For two reason, synthetic datasets were used in this these:
\begin{itemize}
    \item the available non-synthetic datasets does not have enough or doesn't have any semantic labels at all (except \emph{NYU Depth Dataset V2}).
    \item depth maps and semantic labels in non-synthetic datasets are not accurate and may alter the results of the experiments.
\end{itemize}
Note that you may find depth maps and semantic labels in a dataset, while they are not associated with a specific sample\footnote{By the definition of NUSCENES dataset: \cite{caesar_nuscenes_2020} \enquote{A sample is data collected at (approximately) the same timestamp as part of a single LIDAR sweep} \cite{noauthor_nuscenes_nodate}. }. For example \emph{KITTI benchmark} \cite{geiger_vision_2013} has both semantic segmentation subset and depth subset, while they are linked to different samples. Only some of the synthetic datasets have these data synced at the same sample time.

\begin{figure}
    \centering
    \begin{subfigure}{\textwidth}
        \centering
        \includegraphics[width=\linewidth]{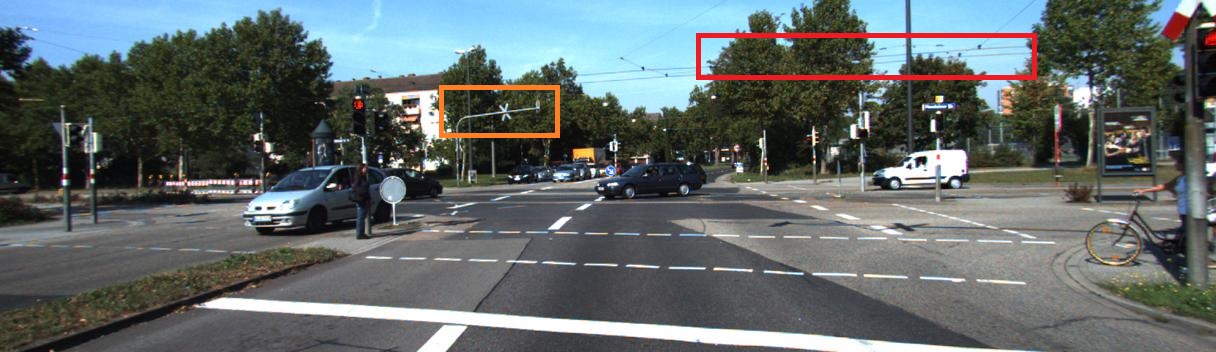}
        \caption{RGB image in frame n}
        \label{fig:kitti_im_14}
    \end{subfigure}%

    \begin{subfigure}{\textwidth}
        \centering
        \includegraphics[width=\linewidth]{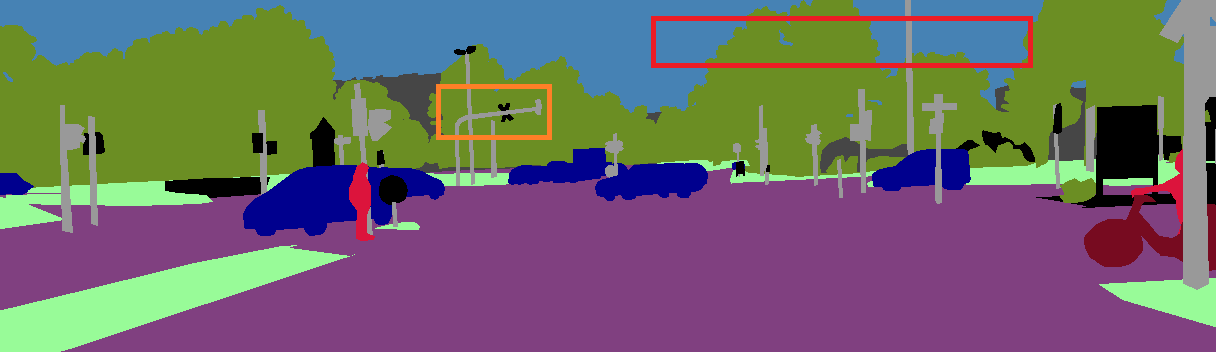}
        \caption{Semantic segments in frame n}
        \label{fig:kitti_ss_14}
    \end{subfigure}%

    \begin{subfigure}{\textwidth}
        \centering
        \includegraphics[width=\linewidth]{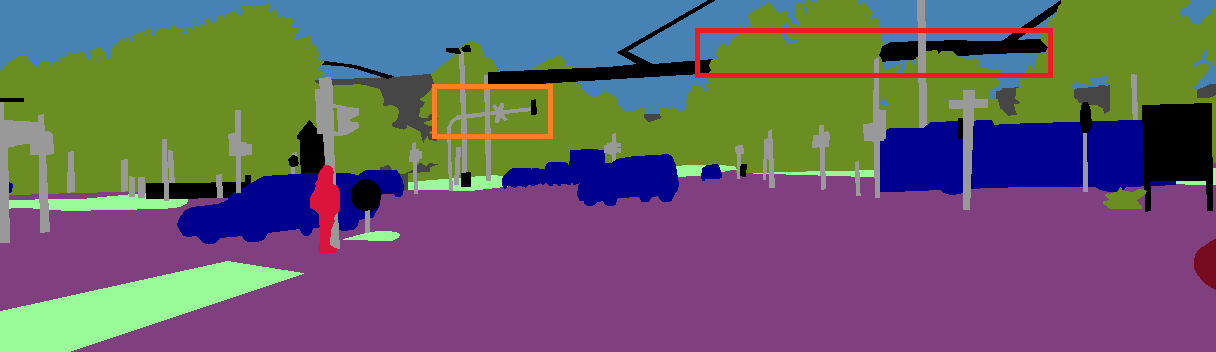}
        \caption{Semantic segments in frame n+1}
        \label{fig:kitti_ss_15}
    \end{subfigure}%

    \begin{subfigure}{\textwidth}
        \centering
        \includegraphics[width=\linewidth]{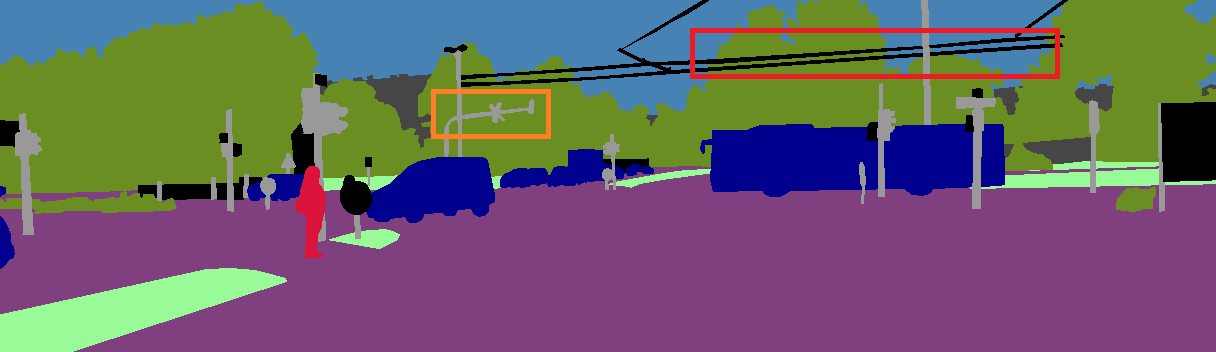}
        \caption{Semantic segments in frame n+2}
        \label{fig:kitti_ss_16}
    \end{subfigure}%
    
    \caption[Inaccurate or false segmentation by a human]{Inaccurate or false segmentation by a human. In these three consecutive frames, different labels are assigned to an object(inside the orange box), and inaccurate labels are assigned to another object such that the label exceeds the covered area by the object(inside the red box). The labels are from KITTI dataset.}
    \label{fig:kitti_ss_error}
    \end{figure}

Semantic labels in non-synthetic datasets are generated manually or automatically by AI-based software under human supervision, mainly because they are not perfect. However, generated semantic segments by a human or supervised by a human are not immune from error and suffer from inaccurate or false segmentation (\autoref{fig:kitti_ss_error}).

\begin{figure}
    \centering
    \includegraphics[width=\linewidth]{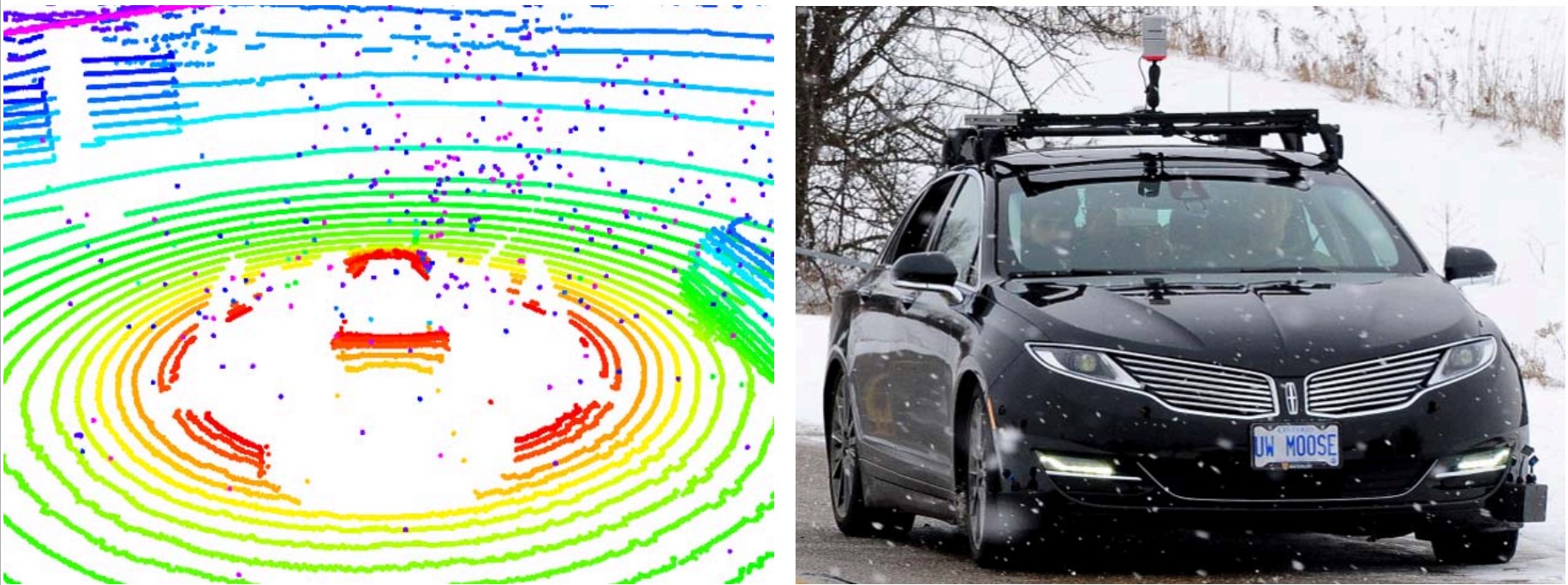}
    \caption[Snowy point cloud collected by LIDAR]{\enquote{Snowy point cloud collected by \gls{LIDAR}} \cite{charron_-noising_2018}}
    \label{fig:LIDAR_in_snow}
    \end{figure}

Depth data in non-synthetic datasets are usually generated by \gls{LIDAR} sensors for outdoor scenes and by RGB-D cameras for indoor scenes. If a \gls{LIDAR} is used for depth generation, \enquote{the sparse point clouds generated can be corrupted by measurement noise due to the internal noise of the sensor and environmental factors such as reflectivity, lighting, and precipitation} \cite{charron_-noising_2018}. The reflectivity of glasses and glossy surfaces also causes many 3D points to be missed \cite{wang_apolloscape_2019}. Downfall of rain and snow blocks laser beams and create noise in the point cloud (\autoref{fig:LIDAR_in_snow}). Another drawback of \gls{LIDAR} is its sparsity of points which increases with distance. Generated depth labels by RGB-D cameras share the same drawbacks in addition to a limited range of view.

The existing issues of depth data and semantic labels in non-synthetic datasets do not exist in synthetic datasets, provided they are designed properly. It can be mentioned that synthetic data are completely correct. But this perfection can be considered superiority only when \gls{CNN}s are capable of predicting high details and impurities in datasets cause them to malfunction.

\subsection{Dataset Adaptation and Domain Adaptation}

Since every dataset has its characteristics and distribution of data , in this thesis multiple datasets from both synthetic and non-synthetic types are examined. In this way the hypothesis of this thesis could be examined more accurately and independent of the impact of datasets' characteristics. Moreover, the difference in the distribution of data in these datasets exposes the used model to the domain shift. In other words, the used models are challenged to adapt to different domains.\footnote{\enquote*{Transfer learning and domain adaptation refer to the situation where what has been learned in one setting (i.e., distribution P1) is exploited to improve generalization in another setting (say distribution P2). In transfer learning, the learner must perform two or more different tasks, but we assume that many of the factors that explain the variations in P1 are relevant to the variations that need to be captured for learning P2. In the related case of domain adaptation, the task (and the optimal input-to-output mapping) remains the same between each setting, but the input distribution is slightly different.} \cite{goodfellow_deep_2016}}. Recently, the problem of domain adaption from synthetic data to real data has attracted lots of attention and has been investigated in several works \cites{sun_virtual_2014, vazquez_virtual_2014, xu_domain_2014, busto_adaptation_2015, ros_synthia_2016, xu_hierarchical_2016, wrenninge_synscapes_2018, luo_taking_2019, nikolenko_synthetic_2019}. Furthermore, this approach introduces a difficulty as well, namely dataset adaptation. Because of some specifics of datasets, they can not be used directly, and some adjustments should be applied to jointly use different datasets.

One necessary adjustment is unifying semantic segmentation policies. Almost every dataset has chosen a different policy in the classification of objects for semantic labels. The only way of unifying datasets is to merge subclasses to superclasses.

\begin{figure}
    \centering
    \includegraphics[width=\linewidth]{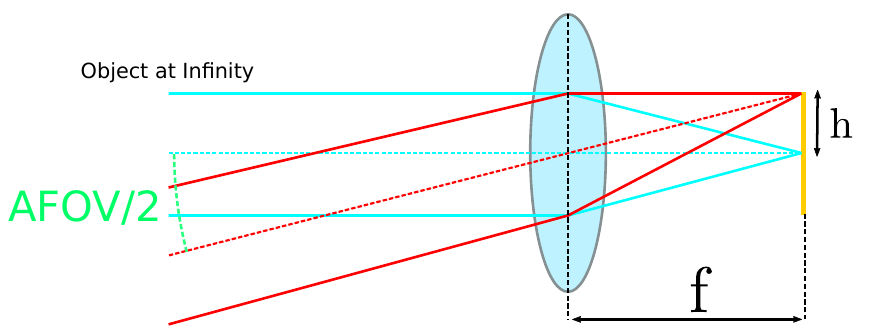}
    \caption[Relation of AFOV with focal length and sensor size]{Relation of \gls{AFOV} with focal length and sensor size. Yellow line is the sensor.}
    \label{fig:AFOV}
    \end{figure}

Another necessary adjustment is the unifying of the \gls{AFOV} of all frames (including RGB images, semantic labels, and depth maps). No two datasets have the same \gls{AFOV}. In photography when you zoom in to magnify the scene, \gls{AFOV} reduces. Therefore, different \gls{AFOV} leads to a different perception of distances. According to optic laws, the value of \gls{AFOV} is a function of the camera's focal length and size of the camera's sensor \autoref{fig:AFOV}.

According to \autoref{eq:AFOV} adjusting the \gls{AFOV} is possible through changing the focal length or sensor size. However, changing the focal length of the camera or the sensor size is not possible after taking a photo. However, since changing the image size is the same as changing the sensor size, one can crop the image to decrease sensor size. Therefore, \gls{AFOV} can only be decreased.

\begin{equation}
    \label{eq:AFOV}
    \text{AFOV} \left[ ^{\circ} \right] = 2 \times \tan ^{-1} { \left( \frac{h}{2f} \right)}
    \end{equation}

    \chapter{Depth from Semantic Labels}

This chapter describes an experimental study that analyzes the performance of several \gls{SIDE} and \gls{SSIDE} \gls{CNN} architectures. This study aims to investigate the probable advantages of \gls{SSIDE} over \gls{SIDE}.

\section{Methods}
\label{sec:firts_exp_method}

For the purpose of the study, one approach is to derive an \gls{SSIDE} \gls{CNN} from an existing \gls{SIDE} \gls{CNN} and compare their performances. The \gls{FCRN} model \cite{laina_deeper_2016} was chosen as the baseline \gls{SIDE} model. The \gls{FCRN} has a straightforward architecture to implement which can be easily manipulated. The details of \gls{FCRN} architecture and its derived architectures are described in the \autoref{sub:DepthModel}, while \autoref{sub:DepthEvaluation} elaborates on the used comparison study. The used datasets and data preparations are described in \autoref{sub:DepthDatsets}.

\subsection{Models} \label{sub:DepthModel}

\begin{figure}[h]
    \centering
    \includegraphics[width=\linewidth]{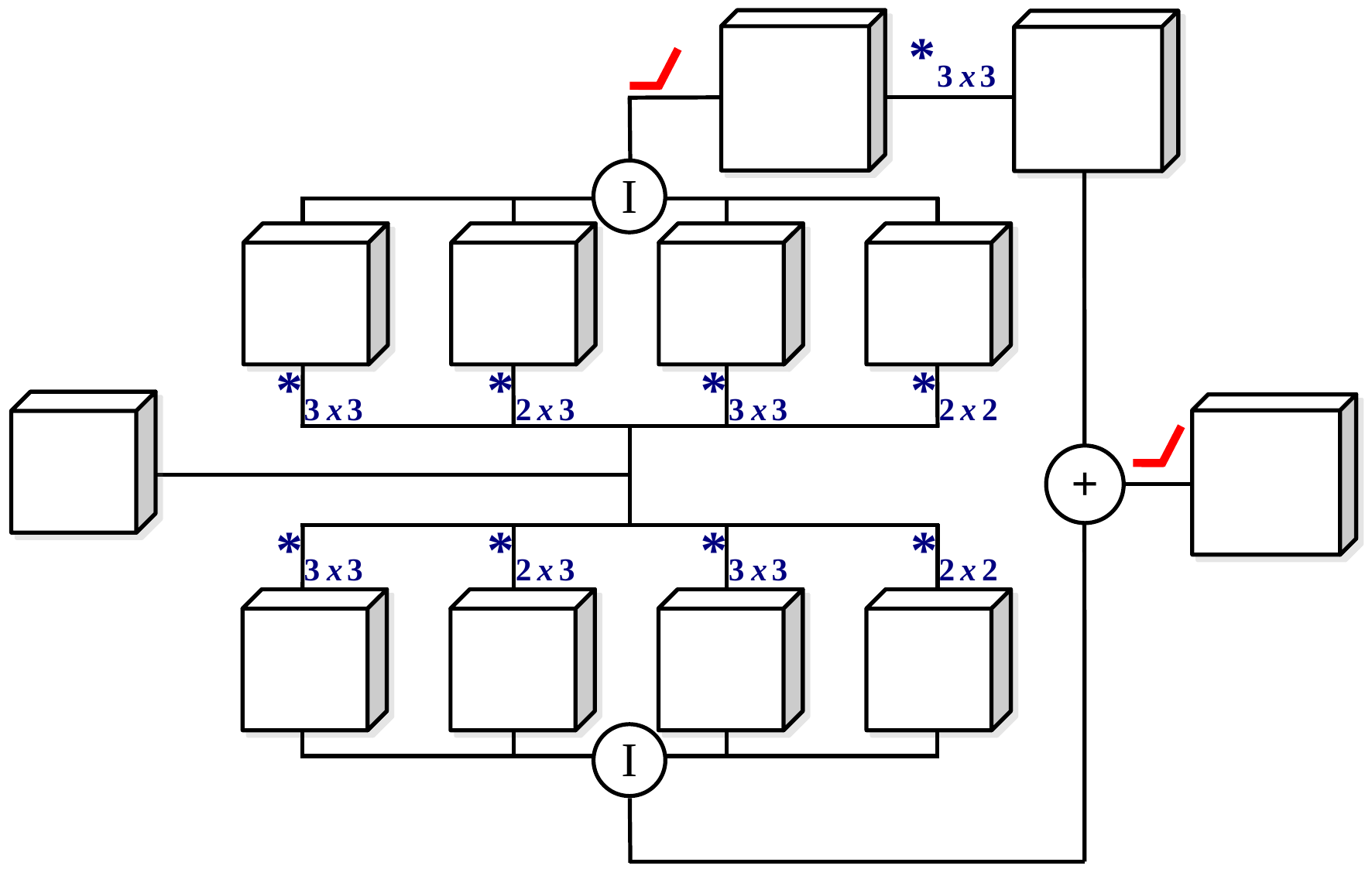}
    \caption[Up-sampling block]{Up-sampling block \cite{laina_deeper_2016}}
    \label{fig:Up-sampling_block}
    \end{figure}

The \gls{FCRN} model has an encoder-decoder architecture. This architecture is depicted in \autoref{fig:FCRN_architecture}. Its encoder is built upon ResNet-50 \cite{he_deep_2016}. The encoder progressively decreases the size of the input data stream, allowing upper layers to extract high-level global features. The encoder consists of the residual units of the ResNet architecture. As it is seen in \autoref{fig:FCRN_architecture}, two kinds of residual units exist (purple units with projection connection and green units with skip connection). Their main difference is that those residual units with convolution in their skip connection reduces the size of the data stream but the other type preserve the size of the data stream. \citeauthor*{laina_deeper_2016} used their proposed up-sampling blocks in the decoder part to yield a depth map using extracted high-level features \cite{laina_deeper_2016}.These blocks have borrowed the idea of residual units in that they have two branches that are added together at the output and one branch plays the role of a skip connection (the lower branch in \autoref{fig:Up-sampling_block}). Each branch consists of four convolutions that their outputs are interleaved together. The \emph{I} sign in \autoref{fig:Up-sampling_block} represents the process of interleaving. Overall, the up-sampling blocks increase the size of the data stream and finally generate the depth map.

\afterpage{%
\begin{landscape}
    \begin{figure}
        \centering
        \includegraphics[width=\linewidth]{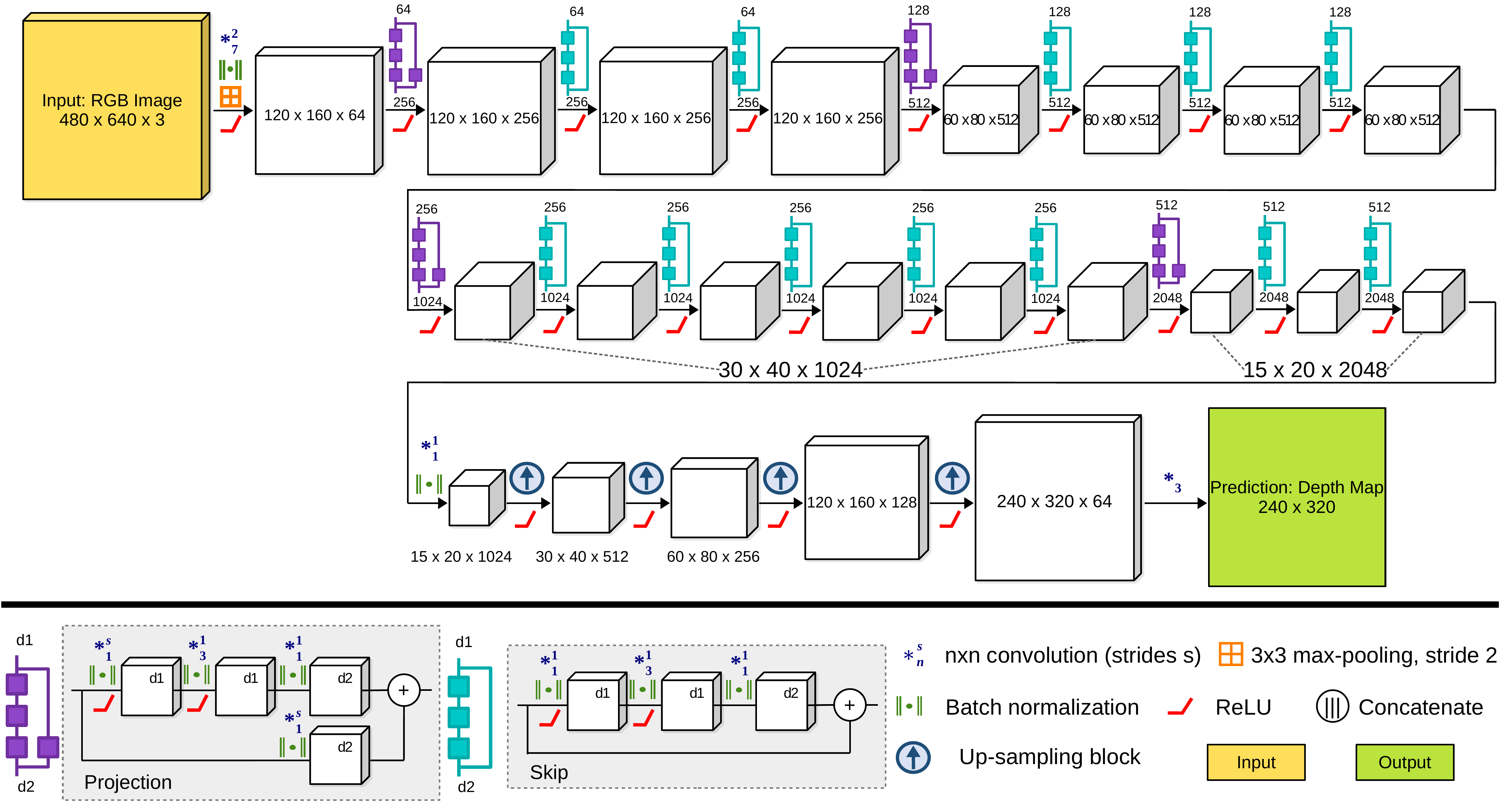}
        \caption[FCRN architecture]{FCRN architecture \cite{laina_deeper_2016}}
        \label{fig:FCRN_architecture}
        \end{figure}
\end{landscape}
}

By selecting the baseline structure, gradually several structures have been driven from it, while derivations continued until the desirable result at the output was achieved. The structures of derived models have a lot in common with the original \gls{FCRN}. Hence, only explanation of differences is necessary. However, full schematics of structures are depicted here for better understanding and reproducibility. Most of the symbols in these schematics are the same as those used in \cite{laina_deeper_2016} for the sake of comparability. Due to being able to distinguish different models, each one is named by a number following \emph{M} character (as an abbreviation of the term \emph{Model}). Also, note that the order of presentation of models is chronological:

\subsubsection{M0 Network}
This network has the same architecture as the \gls{FCRN}, except in its output that has an additional up-sample block. The architecture of this network is depicted in \autoref{fig:M0-2}. This difference enables the M0 to produce a depth map with an equivalent resolution to the resolution of the input image. The performance results of the M0 are used as a representation for the performance of \gls{SIDE} \gls{CNN}s.

\subsubsection{M1 Network}
The difference between the M1 and the M0 lies on the inputs of the networks. In addition to RGB image, semantic label is present in the input of the M1. RGB image and semantic label are fed to the network after being concatenated. The architecture of this network is depicted in \autoref{fig:M0-2}. This structure is probably the simplest derivative of the \gls{FCRN} architecture that integrates RGB image and semantic label to estimate depth map.

\subsubsection{M2 Network}
\begin{figure}
    \centering
    \includegraphics[width=\linewidth]{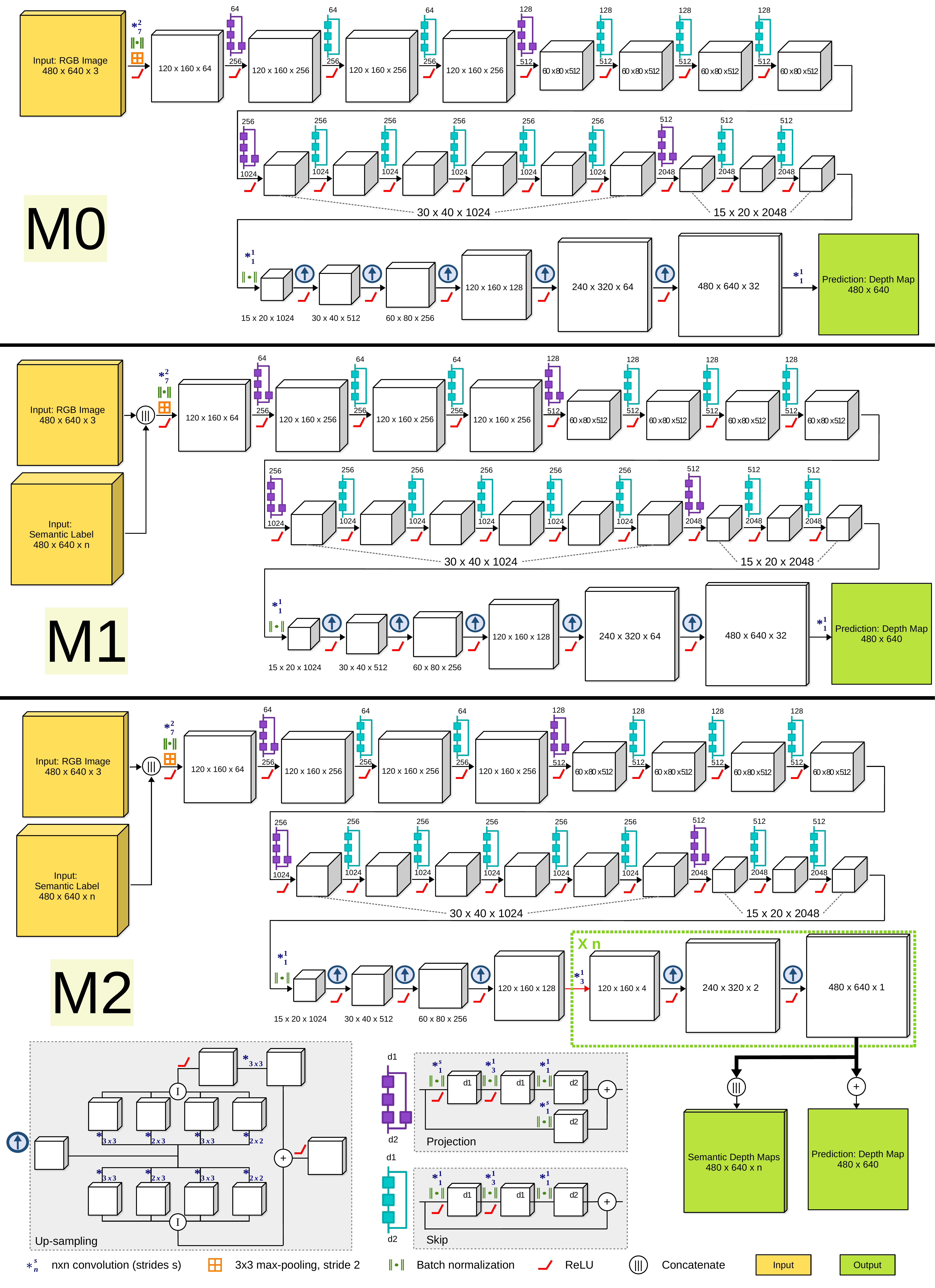}
    \caption{M0-M2 Network Architectures}
    \label{fig:M0-2}
    \end{figure}

In the M2 network, the decoder is divided into two parts: a common part and a semantic part that has $n$ similar branches, while $n$ denotes the number of all semantic segment classes existing in the dataset. These branches are named \emph{semantic decoder}s and the common part is named \emph{common decoder}. The output of the common decoder is fed to all semantic decoders. Then outputs of semantic decoders are concatenated, yielding a semantic depth output and are added up, yielding a complete depth map. Hence, this architecture has two outputs. This design was intended to permit the network separately and semantically learn the depth of each different semantic class. The architecture of this network is depicted in \autoref{fig:M0-2}.

\subsubsection{M3 Network}
The M3 architecture was designed to address a probable drawback of the M2. Despite the designed decoder of M2 with multiple branches for a semantical depth prediction, there are only two outputs, and there are only two gradients used in the process of training, and the gradients are common to all branches (of semantic decoders). It seemed that semantic decoders of the M2 would not learn depth prediction exclusively. According to this issue, the matrix of semantic output with a depth of $n$ was split to $n$ matrices. The architecture of this network is depicted in \autoref{fig:M3-4}.

\subsubsection{M4 Network}
\begin{figure}[t!]
    \centering

    \begin{subfigure}[b]{.32\textwidth}
        \centering
        \includegraphics[width=\textwidth]{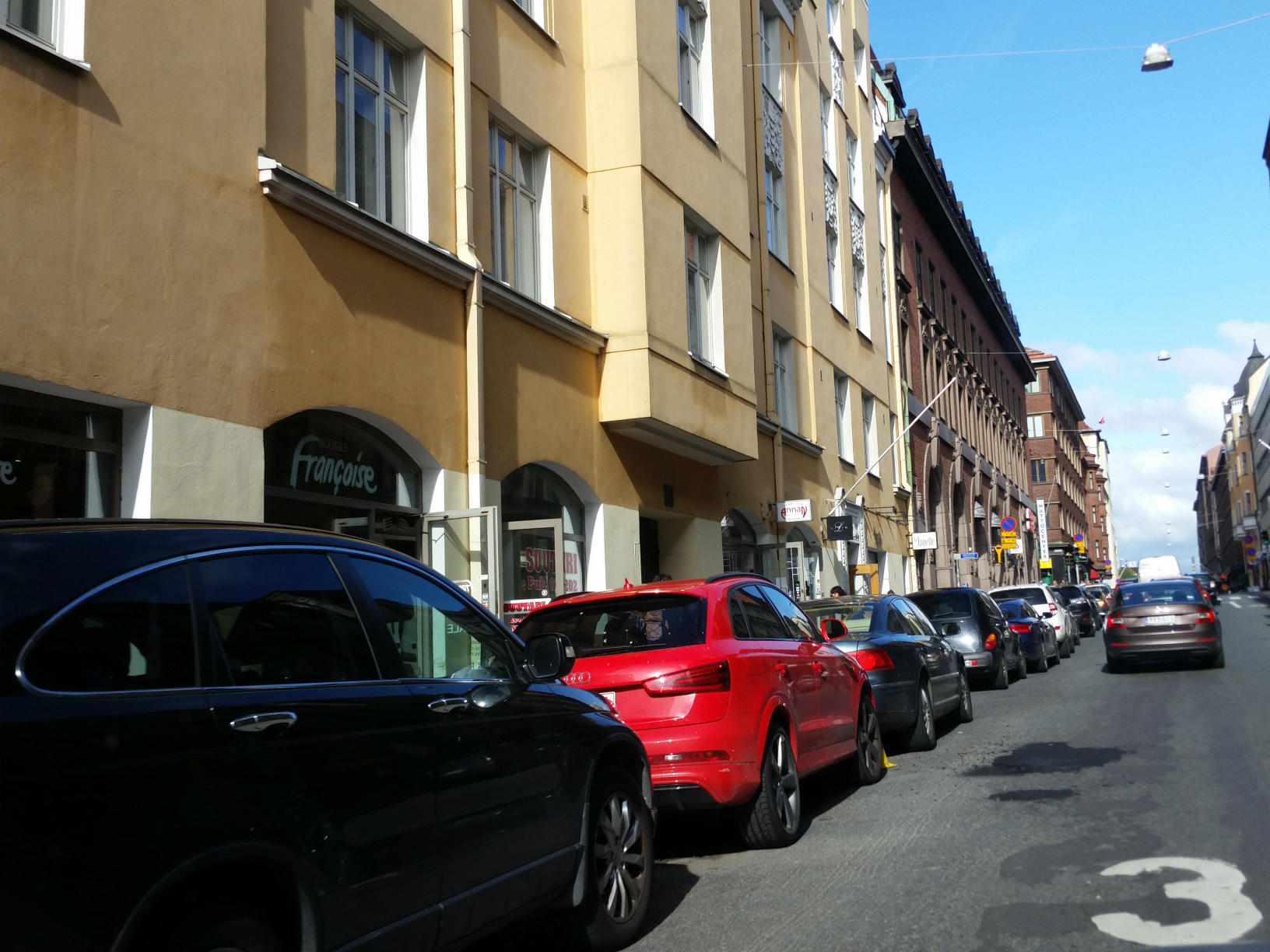}
        \caption{RGB image}
        \label{fig:rgb_image_maskedRGB}
    \end{subfigure}
    \hfill
    \begin{subfigure}[b]{.32\textwidth}
        \centering
        \includegraphics[width=\textwidth]{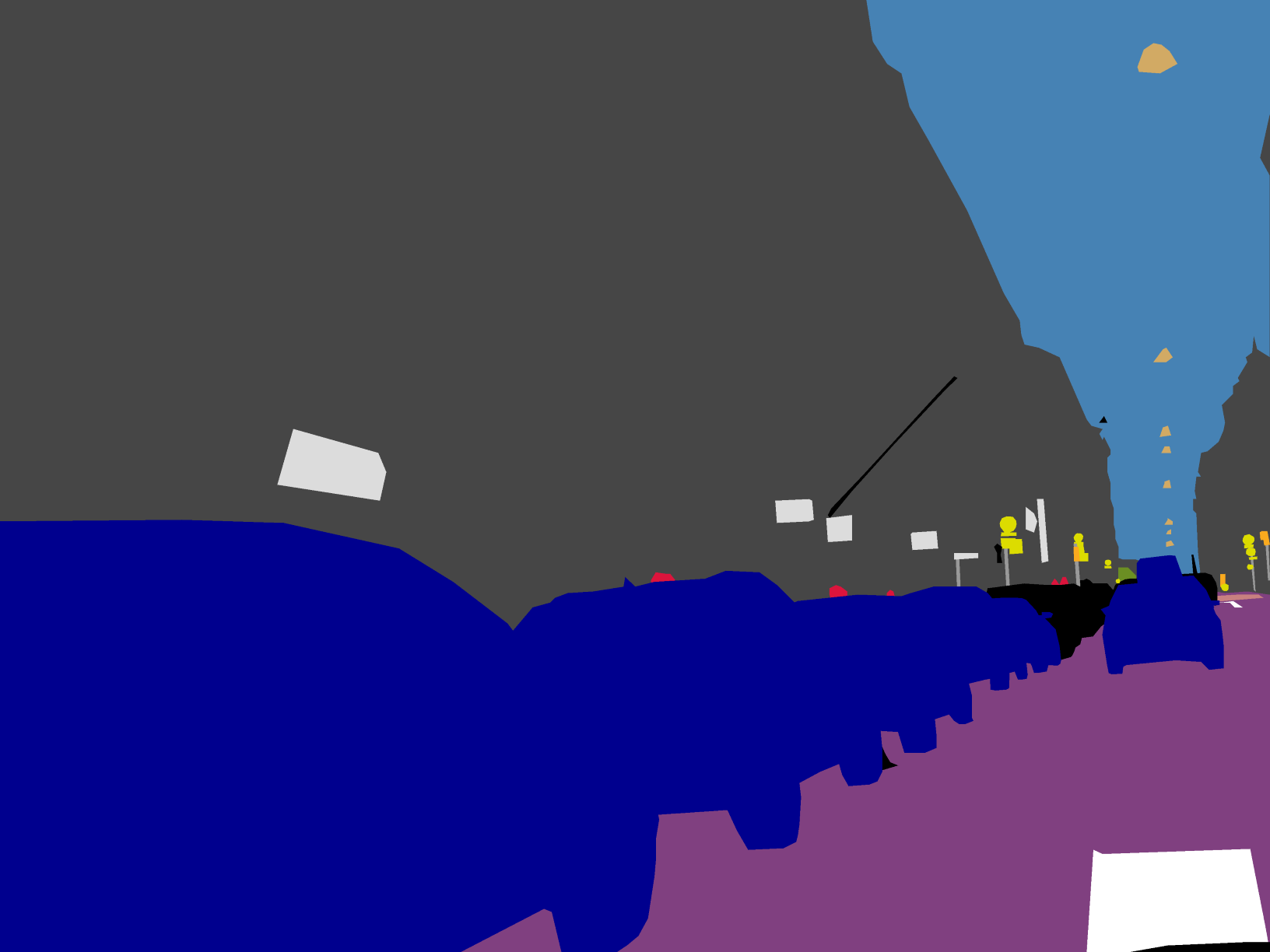}
        \caption{Semantic segments}
        \label{fig:semantic_segmentation_maskedRGB}
    \end{subfigure}
    \hfill
    \begin{subfigure}[b]{.32\textwidth}
        \centering
        \includegraphics[width=\textwidth]{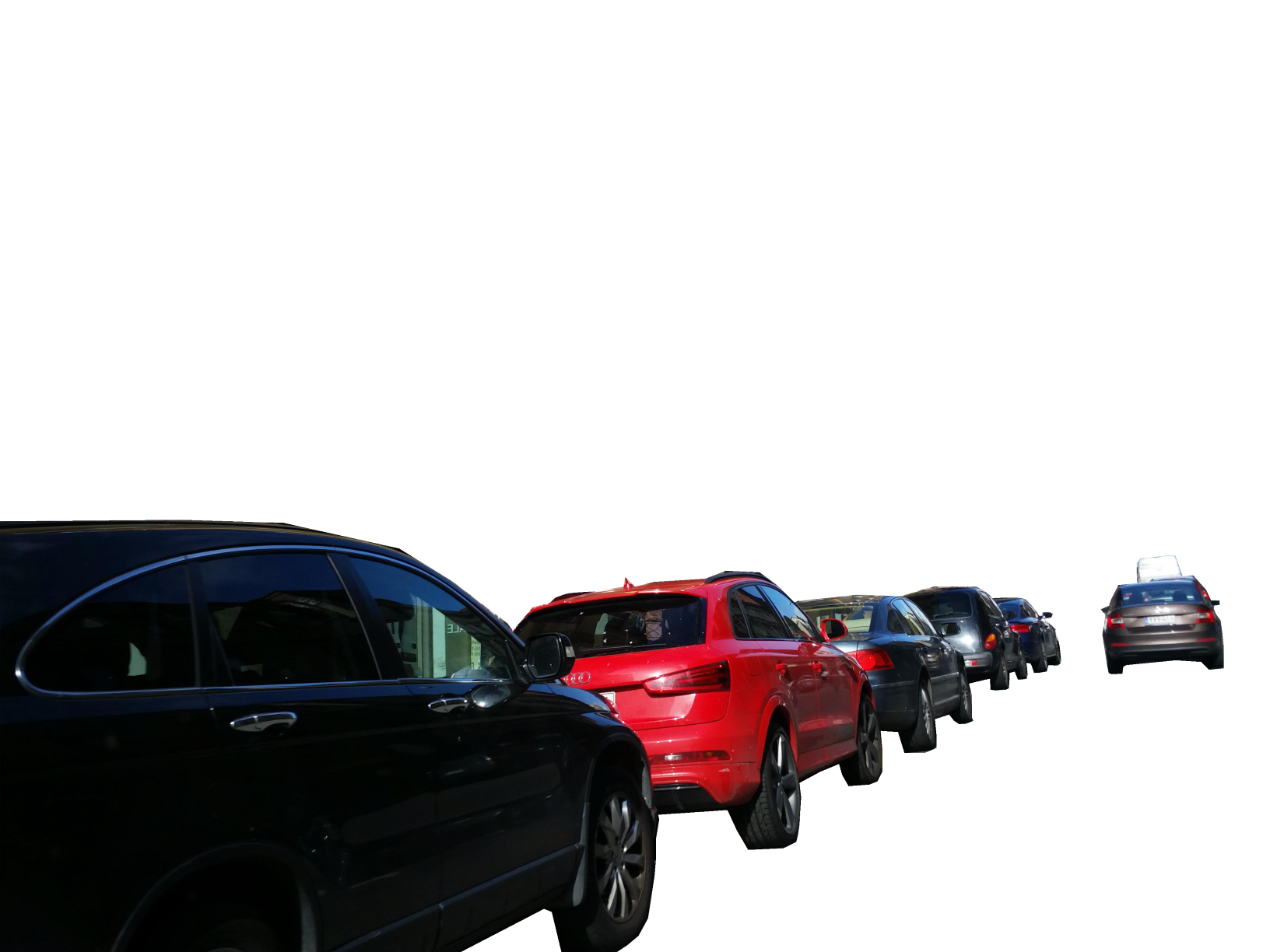}
        \caption{{\tiny Semantically masked image}}
        \label{fig:masked_image_maskedRGB}
    \end{subfigure}

    \caption[Semantically masked image]{Semantically masked image (data from \cite{neuhold_mapillary_2017})}
    \label{fig:maskedRGB}
    \end{figure}

\begin{figure}
    \centering
    \includegraphics[width=\linewidth]{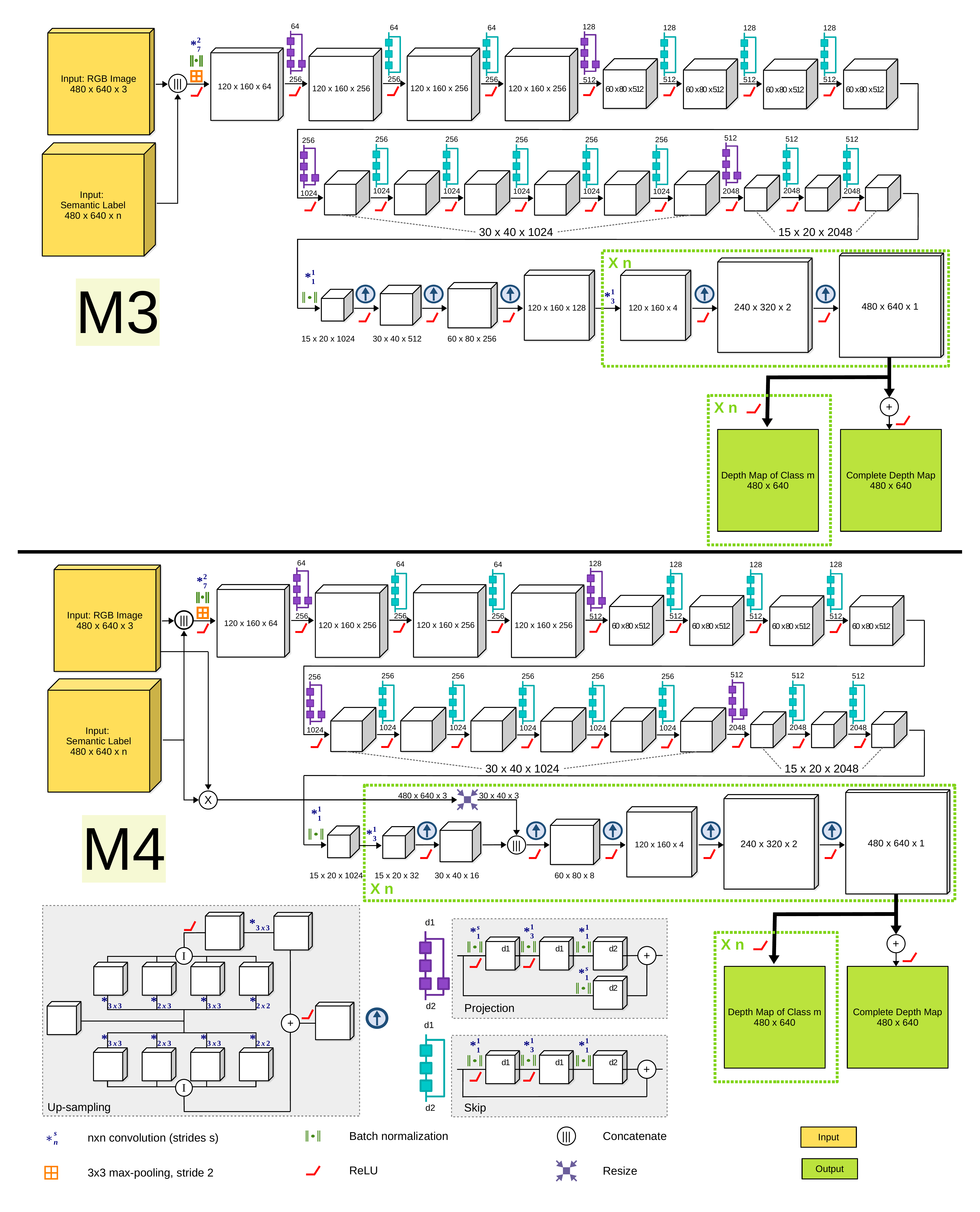}
    \caption{M3-M4 Network Architectures}
    \label{fig:M3-4}
    \end{figure}

The goal of designing the M4 network was to improve the focus of semantic decoders on the probable bindings between RGB images and semantic segments. To achieve this goal, the semantic decoders were expanded and their shares of up-sample blocks were increased The architecture of this network is depicted in \autoref{fig:M3-4}. Furthermore, a new concept was injected into the decoders by masking RGB images with semantic labels and concatenating the results with the mainstream of data at the early stages of the semantic decoders. By this means, a semantic decoder that is responsible for depth prediction of the semantic class of for example \emph{car}, receives a semantically masked image like that shown in \autoref{fig:masked_image_maskedRGB}. Thus, semantic decoders were expected both to help the network to estimate depth maps more accurately and to help the network retrieve the depth map dimensions more precisely.

\subsubsection{M5 Network}
The M5 architecture follows the same goal as the M4 network, but in a bit different way. In the design of the M5 network, the share of the semantic decoders was decreased and the share of the common part was increased to improve the speed of the network, both in learning and inference. In addition to feeding semantic segmented images to the semantic decoders, their outputs are masked semantically to ensure each semantic decoder exclusively estimates the depth of the assigned class to the semantic decoder. The architecture of this network is depicted in \autoref{fig:M5-7}.

\subsubsection{M6 Network}
The M6 network has a similar architecture to that of the M0 (\autoref{fig:M5-7}), while the difference is that the M6 uses edges of the input image instead of the raw image. At the input of the network, a Sobel filter extracts the edges of each channel of an RGB image. The network uses the summation of three edge channels. The purpose of building this network is to evaluate the impact of image's edges on the perception of semantic segments. The edges of semantic segments are (almost, unless it is not visually recognizable) a subset of edges of the image. Hence, edges of semantic labels are helpful in depth estimation if image's edges are helpful in depth estimation. Furthermore, if the performance of M6 is similar to the performance of the best network that uses semantic labels, it indicates that the quality of the depth estimation is increased because of the information in the edges of the image, not because of the information in the semantic segments.

\subsubsection{M7 Network}
\begin{figure}[!h]
    \centering
    \includegraphics[width=\linewidth]{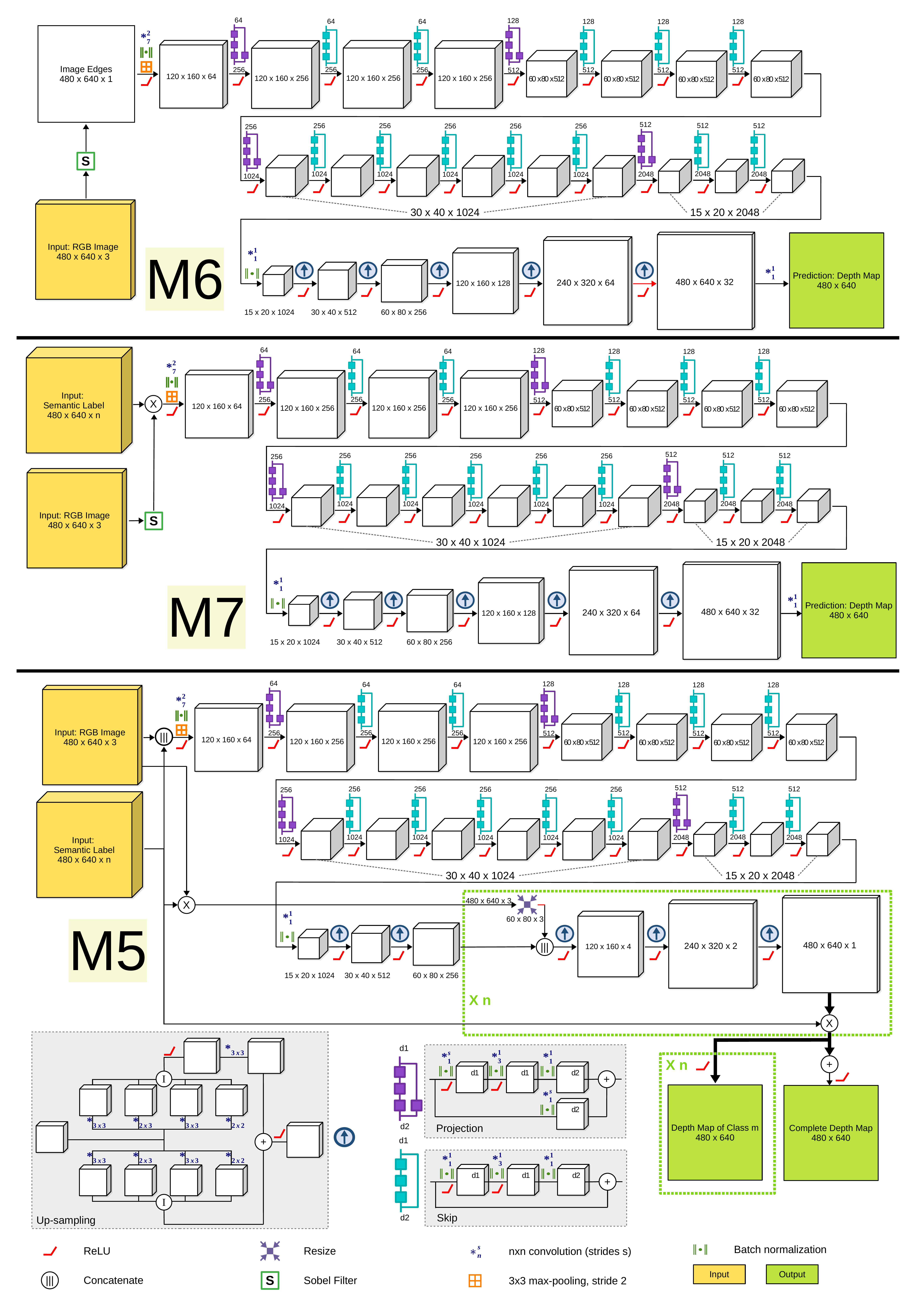}
    \caption{M5-M7 Network Architecture}
    \label{fig:M5-7}
    \end{figure}

The M7 is a combination of the M1 and the M6. Semantic segments layers are multiplied by the image edges form the input of the network, and the result can be named semantic edges. The architecture of this network is depicted in \autoref{fig:M5-7}. If M6 is not capable of improving depth estimation because of the absence of semantic segments, then the M7 may be immune to this defect. If M1 is capable of improving depth estimation, then M7 in addition to M1 capability may have robustness against variations of color in different semantic categories and domain shift, because it uses edges of images.

\subsubsection{M8-M17 Networks}
From M8 to M17 networks a series of models that only use an image or semantic segments as their inputs are developed. From initial models to final models, progressively the size of structures was reduced such that the minimal size of the \gls{FCRN} architecture that is capable of depth estimation be found. Evolution of these models leads to the design of M18, therefore, examination of their structures and their performance in the results section (\autoref{sec:firts_exp_results}) are not necessary.

\subsubsection{M18 Network}
The M18 network is a slim version of M7. During the process of reducing network size, from the M8 to the 17, the capability of slim versions of the \gls{FCRN} architecture in estimating depth was improved. The architecture of this network is depicted in \autoref{fig:M18-21}.

\subsubsection{M19 Network}
The M19 is a slim version of the M6 and a counterpart of M18 that only uses image edges as its input. The architecture of this network is depicted in \autoref{fig:M18-21}. It was examined only for comparison purpose. A comparison of M18 with M6 may not be fair because of the significant difference in their structures.

\subsubsection{M20 Network}
Same as the M19, the M20 is a slim version of M0 and a counterpart of M18 that only uses image as its input. The architecture of this network is depicted in \autoref{fig:M18-21}. It was examined only for comparison purpose. A comparison of M18 with M0 may not be fair because of the significant difference in their structures.

\subsubsection{M21 Network}
\begin{figure}
    \centering
    \includegraphics[width=\linewidth]{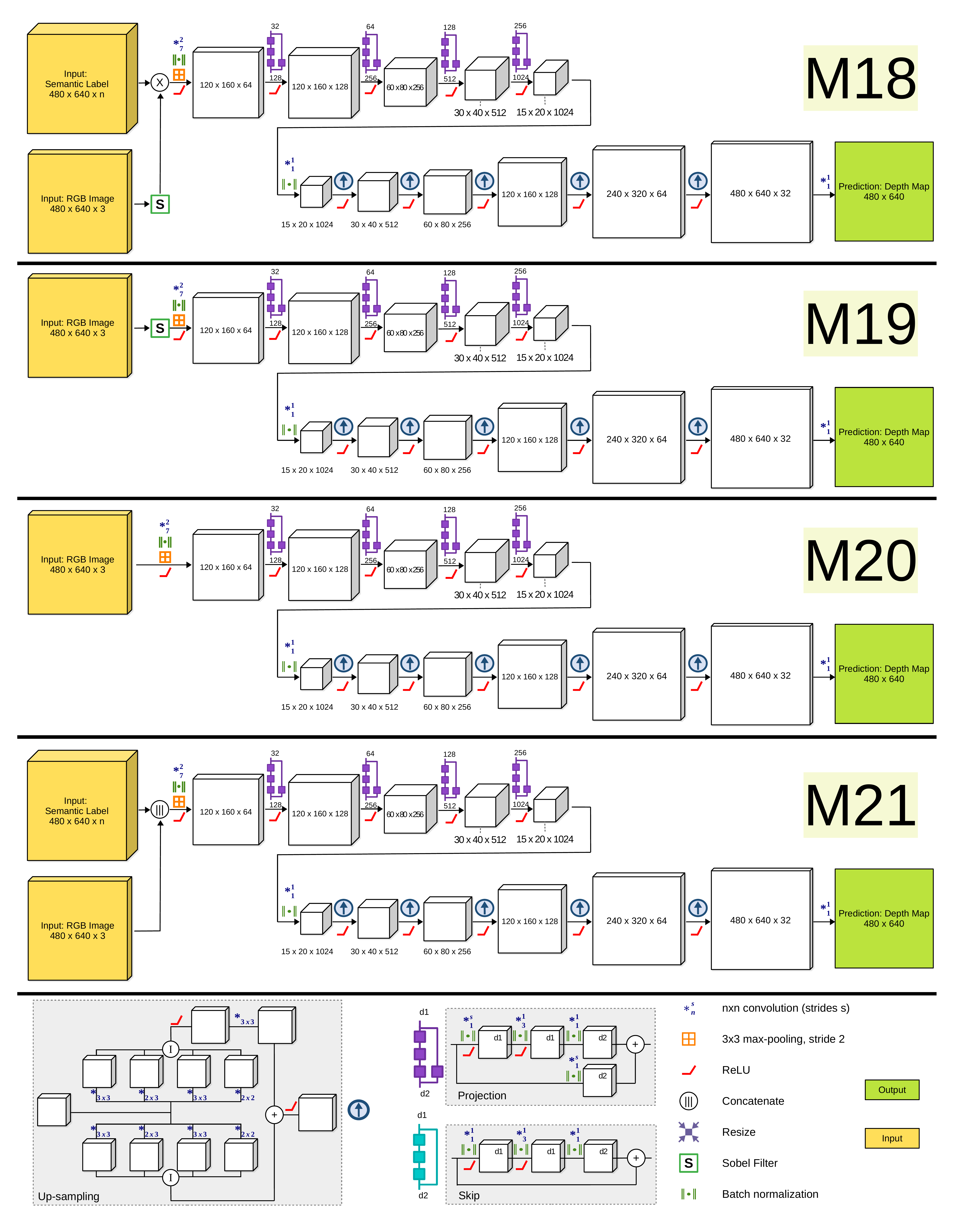}
    \caption{M18-M21 Network Architecture}
    \label{fig:M18-21}
    \end{figure}

Same as the M19 and the M20, the M21 is a slim version of M1 and a counterpart of M18. The architecture of this network is depicted in \autoref{fig:M18-21}. The difference between M21 and M18 is that M18 uses image's edges and semantic segments while the M21 uses raw image and semantic segments.

\subsection{Performance Evaluation and Training Details} \label{sub:DepthEvaluation}
To compare the designed networks, all of them were trained on 75 percent of the SYNTHIA-SF dataset \cite{hernandez-juarez_slanted_2017} and tested on the remaining of the dataset. The number of all semantic segment classes ($n$) existing in the SYNTHIA-SF dataset is 19. The batch size for training of all networks was three \footnote{The maximum possible value for the available hardware}. Since the networks have different structures and different inputs, using same initial weight values is meaningless and impossible, hence no identical seed\footnote{A random seed is a number (or vector) used to initialize a pseudorandom number generator.\cite{noauthor_random_2019}} was used.

After each epoch of training, networks were tested on the test subset of the SYNTHIA-SF. At least all networks were trained for 50 epochs, so as at least 50 times each network was tested, and at least 50 times metrics were calculated for each network. The performances of networks were compared both by visualizing the trend of metrics during 50 epochs and by comparing mean value for each metric from epoch 20 to epoch 50.

\subsubsection{Metrics and Loss Function}
Various metrics and loss functions have been used in works reported on \gls{SIDE}. The following list contains a number of these metrics; however, it is not meant to be an exhaustive list.
\begin{itemize}
    \item \gls{MAPE}
    \begin{equation}
        \frac{100}{n}\sum\left(\frac{|y-y^*|}{y^*}\right)
    \end{equation}
    \item \gls{MSPE}
    \begin{equation}
        \frac{100}{n}\sum\left(\frac{(y-y^*)^2}{y^*}\right)
    \end{equation}
    \item \gls{RMSE}
    \begin{equation}
        \sqrt{\frac{1}{n}\sum\left(y-y^*\right)^2}
    \end{equation}
    \item \gls{RMSElog}
    \begin{equation}
        \sqrt{\frac{1}{n}\sum\left(\log y - \log y^*\right)^2}
    \end{equation}
    \item log10
    \begin{equation}
        \frac{1}{n}\sum|log_{10} y - log_{10} y^*|
    \end{equation}
    \item Delta Thresholds ($\delta_i$)
    \begin{equation}
        \delta_i = \frac{|\{y|\max(\frac{y}{y^*}, \frac{y^*}{y})<1.25^i\}|}{|\{y\}|}
    \end{equation}
    \item \gls{SILog}
    \begin{align}
        D(y,y^*)
        %& =\frac{1}{n}\sum^n_{i=1}(\log y_i - \log y^*_i + \alpha(y,y^*))^2 \\
        & =\frac{1}{n^2}\sum_{i,j}((\log y_i - \log y_j)-(\log y^*_i - \log y^*_j))^2
        \label{eq:Scale_Invariant_Error_1}\\
        & = \frac{1}{n}\sum_i{d^2_i}-\frac{1}{n^2}\sum_{i,j}d_i d_j =
        \frac{1}{n}\sum_i{d^2_i}-\frac{1}{n^2}\left(\sum_i d_i\right)^2
        \label{eq:Scale_Invariant_Error_2}
    \end{align}
\end{itemize}

These metrics may be used directly or indirectly as a loss function. Since examining the impact of different loss functions on depth estimation was not the purpose of this thesis, \gls{MAPE} as a typical metric, was chosen as the loss function. This metric is simple but calculates the error meaningfully. Because of its relative nature, errors in near distances are more important than errors in far distances. Therefore, the network is trained to be more accurate in near distances, as required for practical implementations. Furthermore, other metrics were calculated for detailed evaluation during the tests.

\subsection{Datasets} \label{sub:DepthDatsets}
First, this experimental study was performed on the SYNTHIA-SF dataset. Later, to make sure that the results were general, some parts of the experimental study were repeated on the Virtual KITTI dataset \cite{cabon_virtual_2020} and NYU depth dataset (This dataset contains 894 semantic classes, but only 21 classes were used). For all used datasets, the following data preparations were performed:
\begin{itemize}
    \item Images values were normalized to [0, 1] range.
    \item Semantic segments were converted from RGB mappings
    to one-hot vectors.
    \item Images and Semantic segments were resized to $640 \times 480$.
    \item Depth data dimension were converted to meters.
\end{itemize}

\section{Results}
\label{sec:firts_exp_results}

The comparison of performances is based on the \gls{MAPE} metric on the test set of SYNTHIA-SF dataset. First, performances of \gls{SIDE} and \gls{SSIDE} models are compared separately in two graphs of \autoref{fig:mape_nonsem} and \autoref{fig:mape_sem}, respectively. Then performances of two selected models from \gls{SIDE} group (the best model and the original model) are compared with two selected models from \gls{SIDE} group (two of the best models) in \autoref{fig:mape_sem_vs_nonsem}.

% \subsection{Comparison of SIDE Models}
\autoref{fig:mape_nonsem} shows the performances of the \gls{SIDE} models during test and train. As it can be seen M0's graph and the M6's graph are twisted to each other. The same is true for the M19 and the M20. The M6 and the M19 are counterparts for the M0 and the M20, respectively. The M6 and the M19 instead of using a raw input image, pass it through the Sobel filter and use its edges. These observation imply that the usage of the image edges or the usage of the raw image, results in a fairly similar performance. \autoref{fig:mape_nonsem} also shows that the M20 and the M19 performed better than the M0 and the M6, in general.

\afterpage{%
\begin{landscape}
    \begin{figure}
        \centering
        \includegraphics[height=0.85\textwidth, width=0.8\paperheight]{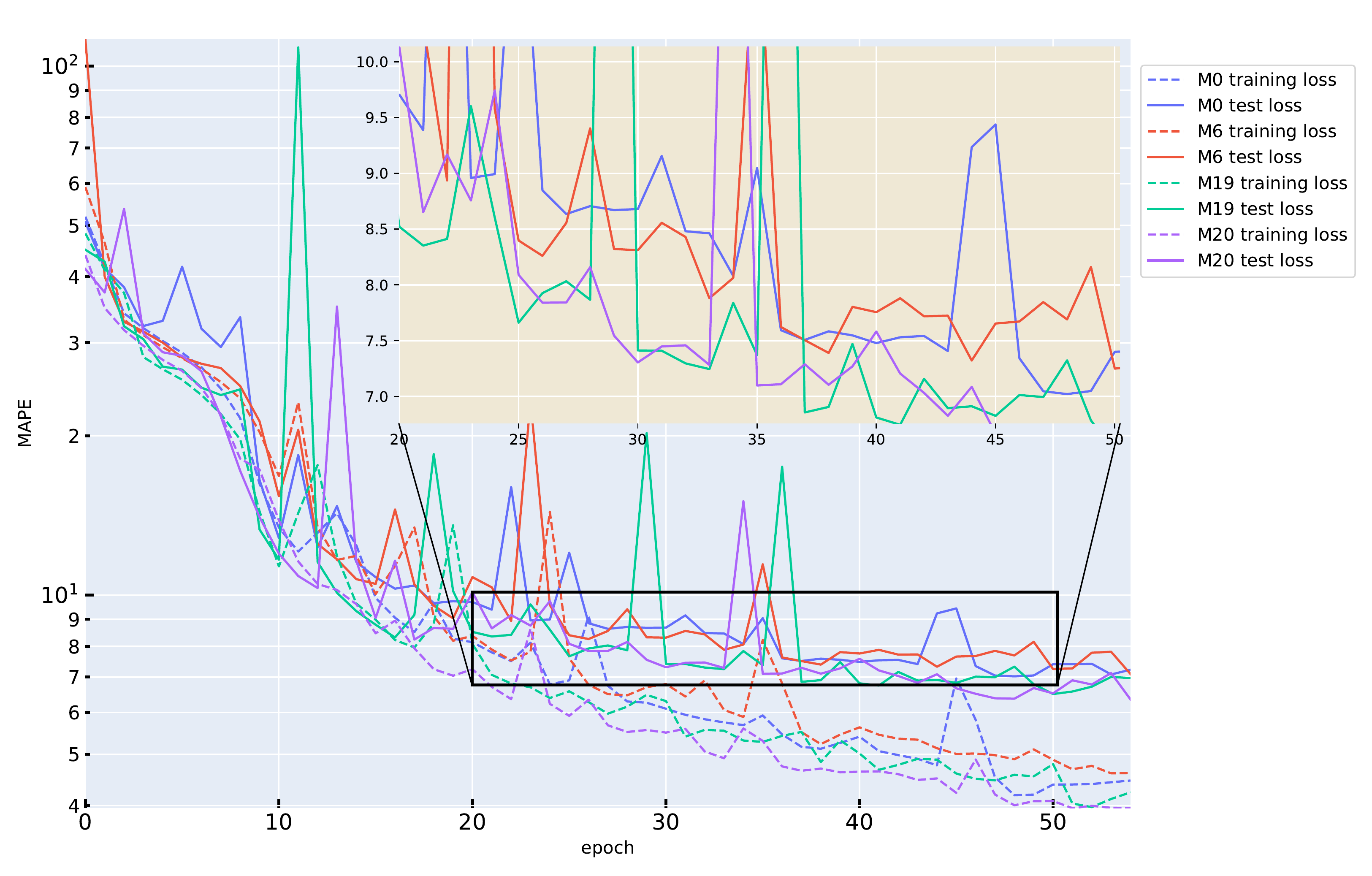}
        \caption[Performance comparison of SIDE DNNs]{Performance comparison of \gls{SIDE} \gls{CNN}s}
        \label{fig:mape_nonsem}
        \end{figure}
\end{landscape}
\begin{landscape}
    \begin{figure}
        \centering
        \includegraphics[height=0.85\textwidth, width=0.8\paperheight]{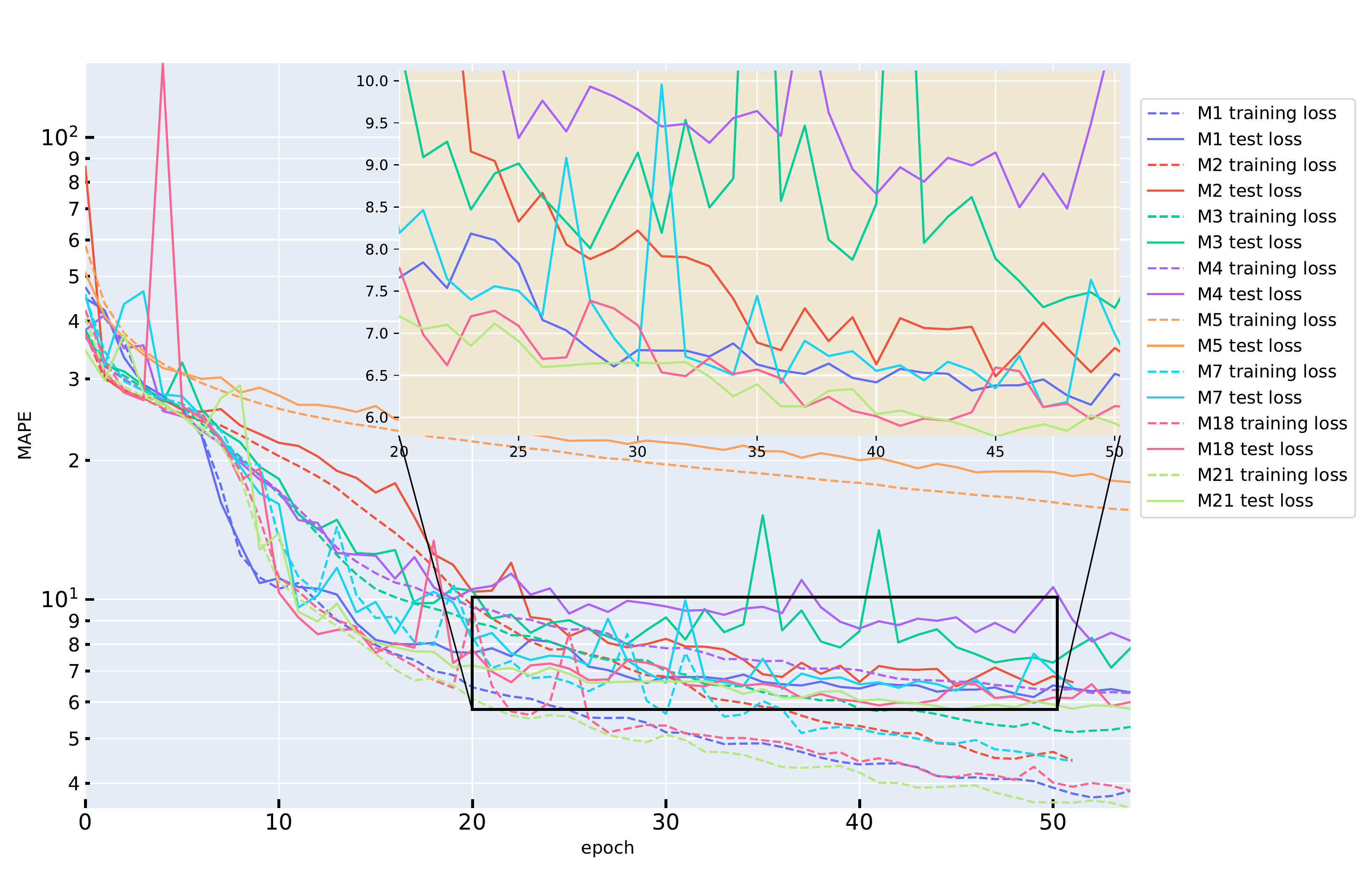}
        \caption[Performance comparison of SSIDE DNNs]{Performance comparison of \gls{SSIDE} \gls{CNN}s. Performance ranks of the networks are not the same during training and test. This shows some network are more capable in generalization.}
        \label{fig:mape_sem}
        \end{figure}
\end{landscape}
\begin{landscape}
    \begin{figure}
        \centering
        \includegraphics[height=0.85\textwidth, width=0.8\paperheight]{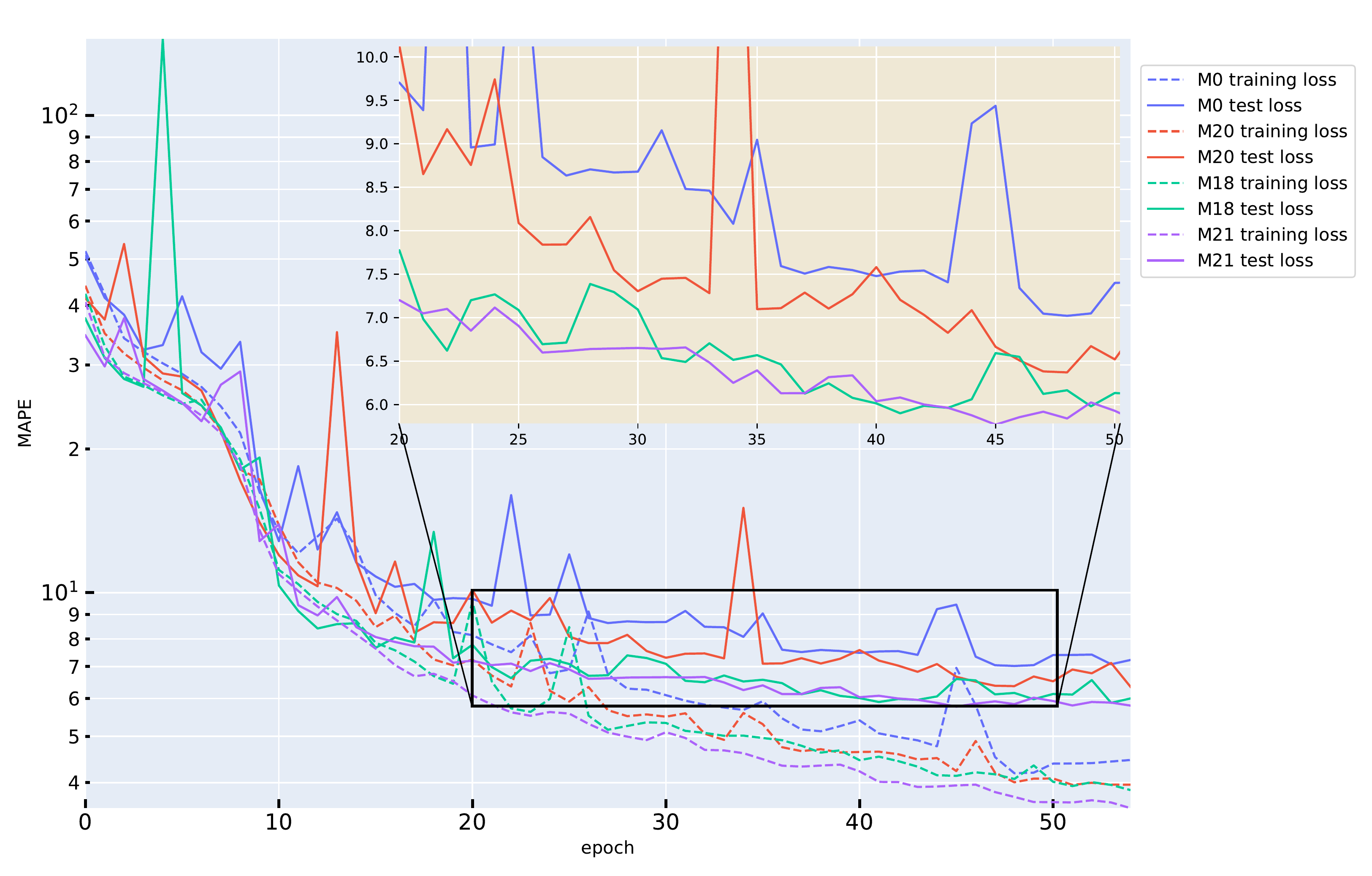}
        \caption{Performance comparison of M0, M20, M18, and M21}
        \label{fig:mape_sem_vs_nonsem}
        \end{figure}
\end{landscape}

\begin{landscape}
    \begin{table}
        \centering
        \caption[Performance comparison of SSIDE and SIDE models]{Performance comparison of \gls{SSIDE} and \gls{SIDE} models. This table contains the average of metrics (calculated on the test subset of SYNTHIA-SF dataset during training) between epoch 20 and 50. The last two columns are the number of trainable parameters of each network and runtime for a single prediction. Smaller values are better except for the delta thresholds, trainable parameters, and runtime.}
        \begin{tabular}{lccccccccccc}
            \hline
            Model & MAPE & MSPE & RMSE & RMSELog & Log10 & $\delta_1$ & $\delta_2$ & $\delta_3$ & SILog & Trainable params & Runtime[s] \\\hline
            M0 & 8.639 & 1607.413 & 102.129 & 0.407 & 0.114 & 0.924 & 0.962 & 0.973 & 0.181 & 62819073 & 0.120 \\
            M1 & 6.852 & 933.107 & 84.719 & 0.331 & 0.088 & 0.939 & 0.974 & 0.983 & 0.113 & 62836481 & 0.172 \\
            M2 & 7.836 & 1110.061 & 96.477 & 0.336 & 0.099 & 0.935 & 0.971 & 0.981 & 0.117 & 62392442 & 0.363 \\
            M3 & 8.921 & 1185.751 & 97.803 & 0.351 & 0.109 & 0.925 & 0.967 & 0.979 & 0.127 & 62392442 & 0.374 \\
            M4 & 9.556 & 1270.288 & 104.149 & 0.382 & 0.121 & 0.916 & 0.963 & 0.975 & 0.148 & \cellcolor[HTML]{9AFF99}\textbf{31776122} & 0.638 \\
            M5 & 20.994 & 11255.954 & 331.330 & 1.600 & 0.492 & 0.749 & 0.786 & 0.795 & 2.589 & 61599898 & 0.501 \\
            M6 & 8.877 & 1592.307 & 104.419 & 0.416 & 0.120 & 0.916 & 0.960 & 0.972 & 0.184 & 62812801 & 0.120 \\
            M7 & 7.131 & 1107.448 & 93.785 & 0.371 & 0.098 & 0.938 & 0.971 & 0.979 & 0.143 & 62833345 & 0.176 \\
            M18 & 6.573 & 991.400 & 87.162 & 0.342 & 0.086 & \cellcolor[HTML]{9AFF99}\textbf{0.951} & 0.976 & 0.982 & 0.121 & 41206145 & 0.154 \\
            M19 & 8.266 & 1649.762 & 93.452 & 0.373 & 0.099 & 0.937 & 0.967 & 0.976 & 0.144 & 41157313 & 0.106 \\
            M20 & 7.823 & 1525.541 & 95.763 & 0.383 & 0.101 & 0.939 & 0.968 & 0.976 & 0.157 & 41149697 & \cellcolor[HTML]{9AFF99}\textbf{0.100} \\
            M21 & \cellcolor[HTML]{9AFF99}\textbf{6.403} & \cellcolor[HTML]{9AFF99}\textbf{925.714} & \cellcolor[HTML]{9AFF99}\textbf{83.090} & \cellcolor[HTML]{9AFF99}\textbf{0.326} & \cellcolor[HTML]{9AFF99}\textbf{0.083} & 0.946 & \cellcolor[HTML]{9AFF99}\textbf{0.976} & \cellcolor[HTML]{9AFF99}\textbf{0.984} & \cellcolor[HTML]{9AFF99}\textbf{0.111} & 41215553 & 0.160\\\hline
            \end{tabular}
        \label{table:average_metrics}
        \end{table}

    \begin{table}
        \centering
        \caption[Relative superiority of the M21 and the M18 over the M0 and the M20]{Relative superiority of the M21 and the M18 over the M0 and the M20 for different metrics in percent. The two last columns are the ratio of the networks' sizes and the ratio of the networks' speeds. The minus sign means retrogression.}
        \begin{tabular}{ l c c c c c c c c c c c }
        \hline
        & MAPE & MSPE & RMSE & RMSELog & Log10 & $\delta_1$ & $\delta_1$ & $\delta_3$ & SILog &   Trainable params & Runtime \\\hline
        m21 relative to m0 & 25.9 & 42.4 & 18.6 & 19.9 & 27.5 & 2.4 & 1.5 & 1.1 & 38.8 & 34.4 & -32.9 \\
        m21 relative to m20 & 18.1 & 39.3 & 13.2 & 14.7 & 17.9 & 0.7 & 0.8 & 0.8 & 29.8 & 0.2 & -59.0 \\
        m18 relative to m0 & 23.92 & 38.32 & 14.66 & 16.14 & 24.21 & 2.91 & 1.45 & 0.95 & 32.81 & 34.41 & -27.99 \\
        m18 relative to m20 & 15.98 & 35.01 & 8.98 & 10.73 & 14.21 & 1.27 & 0.77 & 0.57 & 22.93 & 0.14 & -53.12 \\
        \hline
        \end{tabular}
        \label{table:superiority_in_metrics}
        \end{table}
\end{landscape}
}

% \subsection{Comparison of SSIDE Models}
\autoref{fig:mape_sem} shows the performances of the \gls{SSIDE} models during test and train. As you can see, the performances of the M2, the M3, the M4, and the M5 are weaker than the performances of the M1, the M7, the M18, and the M21. The first conclusion from this observation is that it is harder to improve models' performance by injecting raw information (semantic segments) in higher layers of an \gls{SSIDE} network (as the M4 and the M5). The second conclusion is that creating different branches for every semantic class at the final layers does not necessarily help to improve the performance of an \gls{SSIDE} network (as in the M2, the M3, the M4, and the M5). The final conclusion is that the M18 and the M21 (minimized counterparts of the M7 and the M1 respectively) have superior performances.

% \subsection{Bests of SSIDE vs Bests of SIDE}
In \autoref{fig:mape_sem_vs_nonsem} the original \gls{SIDE} network (M0), the superior network of the first group (M20), and two superiors of the second group (M18 and M21) are compared. The performance superiority of the M21 and the M18 is clearly seen in this graph.

% \subsection{Numerical comparison}
For a quantitative comparison, averages of all used metrics from epoch 20 to epoch 50 were calculated for all the networks and reported in here. The lower bound of this span was selected because the smooth trend of graphs starts approximately from the 20'th epoch. The higher bound was also selected since some of the networks were only trained up to 50 epochs. These averages are collected and represented in \autoref{table:average_metrics}. According to the table data, the M21 had the best performance in eight metrics among the nine metrics considered in this study.

Besides metric values, the runtime of a single prediction and the number of trainable parameters of each network were mentioned in the \autoref{table:average_metrics}. The M21 has two million parameters less than the original network (M0); nevertheless, its runtime is 40 ms longer than the runtime of the original network (M0). The reason for this is that the M21 has 19 more layers in its input matrix. This fact significantly increases the calculation time of the first convolutional layer.

\autoref{table:superiority_in_metrics} contains the percentage of the relative outperforming of the M21 and the M18 over the M0 and the M20 in each metric. Outperforming vary from 0.7 to 42 percent. Apart from delta thresholds, other metrics improved at least by 8.9 percent. However, delta thresholds improved by less than two percent. These small improvements are the result of the discrete nature of the delta thresholds. A delta threshold only counts the desirable results that lie within a specific threshold and it does not take into account the difference of the results to the threshold. This results in the fact that varying the threshold may increase or decrease the improvement percentage.

\begin{figure}[t]
    \centering
    \includegraphics[width=\textwidth]{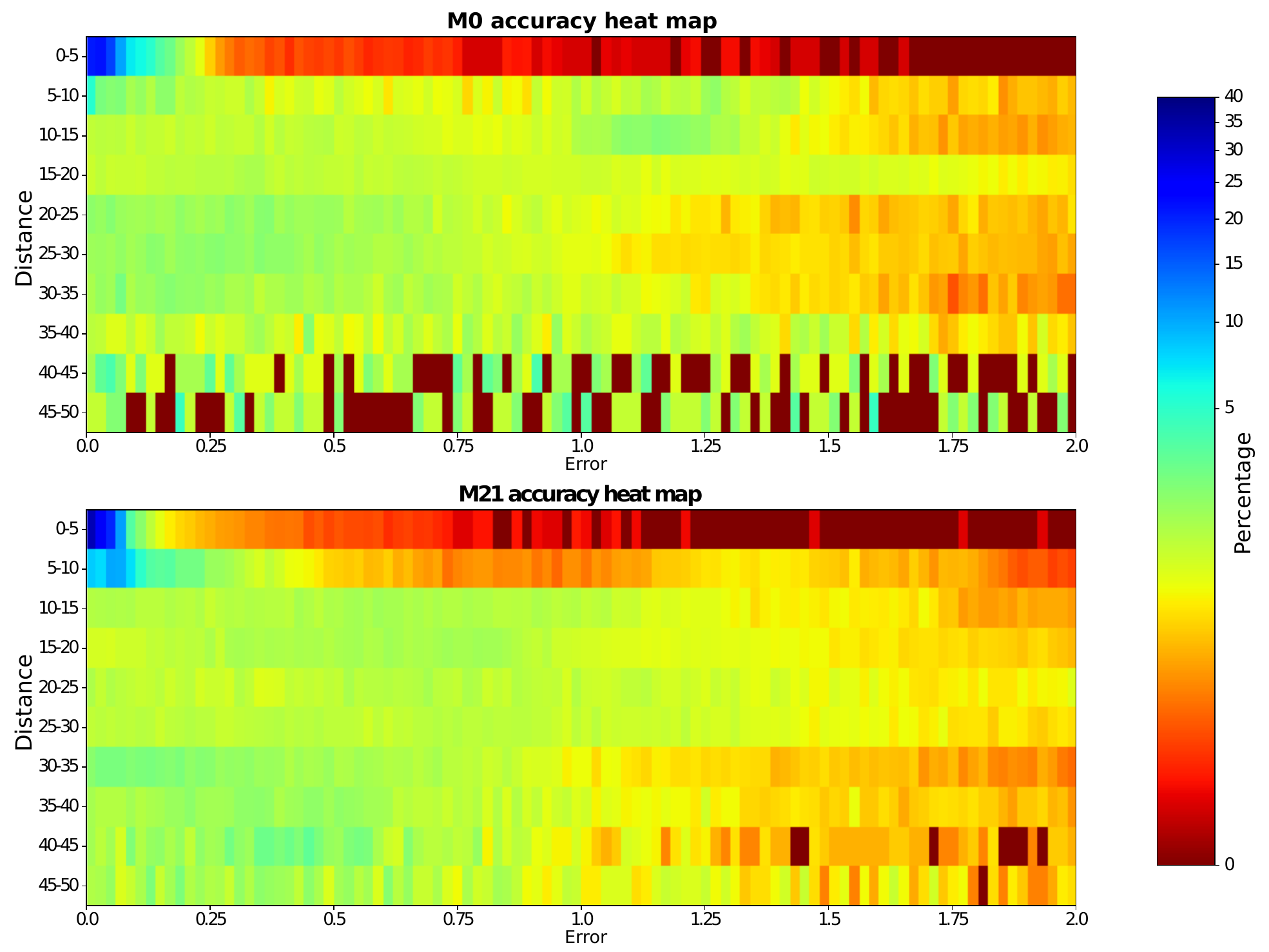}
    \caption[Accuracy heat map of M0 and M21]{Accuracy heat map of M0 and M21. [0, 20] and  [30, 40]: error distribution for the M21 has a closer median to zero. Also, you can see a peak in the error distribution for M0 around 1.2m in [10, 15] m range and it's not valid for M21. [20, 30]: the M0 has a better error distribution. [40, 50]: distribution of the the error for the M0 seems like noise, however for the M21 this distribution is smooth and has a growing trend toward zero. Note that colors are mapped to percentages in a power norm.}
    \label{fig:accuracy_heat_map_m0_m21}
    \end{figure}

Delta thresholds are not detailed and they are neither usable in practical implementations. In practice, one may want to know the accuracy of the depth estimation in different distances separately. Furthermore, a few thresholds for measuring accuracy (common thresholds for delta threshold metric are $\delta_1$: 1.250m, $\delta_2$: 1.563m, $\delta_3$: 1.953m) are not enough to determine the distribution of error. A histogram for the error distribution may express more information about the accuracy of depth estimation. Combining these two ideas to a heat map diagram \autoref{fig:accuracy_heat_map_m0_m21} is the result of calculating error histogram for different ranges. This graph shows how the M21 outperformed the M0 in different rages. For example, the error distribution in range [0, 5] $m$ for the M21 is more intensive near zero relative to that of the M0.

To study the visual performance comparison for M18 and M20 networks, one sample of depth estimation of M20 and M18 (in SYNTHIA-SF dataset) is depicted in \autoref{ch:Experiments_Results}. Similar comparisons between M18 and M20 were performed in Virtual KITTI and NYU depth datasets. The M18 outperformed the M20 by 52 percent of \gls{MAPE} metric in the Virtual KITTI datasets and by 17 percent of \gls{MAPE} metric in the NYU depth dataset.

    \chapter{Synthetic Data instead of Non-Synthetic Data}

To ensure that synthetic datasets are capable of training a segmentation model to generate accurate semantic segments for non-synthetic images, an experimental study was carried out. This chapter describes this experimental study.

\section{Methods}
\label{sec:second_exp_method}

In this experimental study, first, multiple segmentation models with the same architecture are trained on different synthetic and non-synthetic datasets. Second, each model is tested on each dataset to determine how general is the training dataset of the model; if the performance of the model is uniform over all datasets, the training dataset of the model is comprehensive relative to other datasets. Third, performances of all models are compared on each non-synthetic datasets to determine how models that were trained on synthetic datasets perform relative to models that were trained on non-synthetic datasets. Note that this comparison excluded the performance of the model that is not affected by domain shift, but this performance was mentioned as a reference in the comparison. If a model that was trained on a synthetic dataset performs better than a model that was trained on a non-synthetic dataset, the training dataset of the earlier model is assigned as appropriate for the task mentioned in the goal of this experimental study, i.e., to train a segmentation model to generate semantic segments for non-synthetic images.

\subsection{Semantic Segmentation Model}
Lots of semantic segmentation models are available, but for simplicity and ease of implementation, a basic model was used. This model is an improved variant \cite{noauthor_google_nodate} of the U-Net \cite{ronneberger_u-net:_2015} network.

\begin{figure}[ht]
    \centering
    \includegraphics[width=\linewidth]{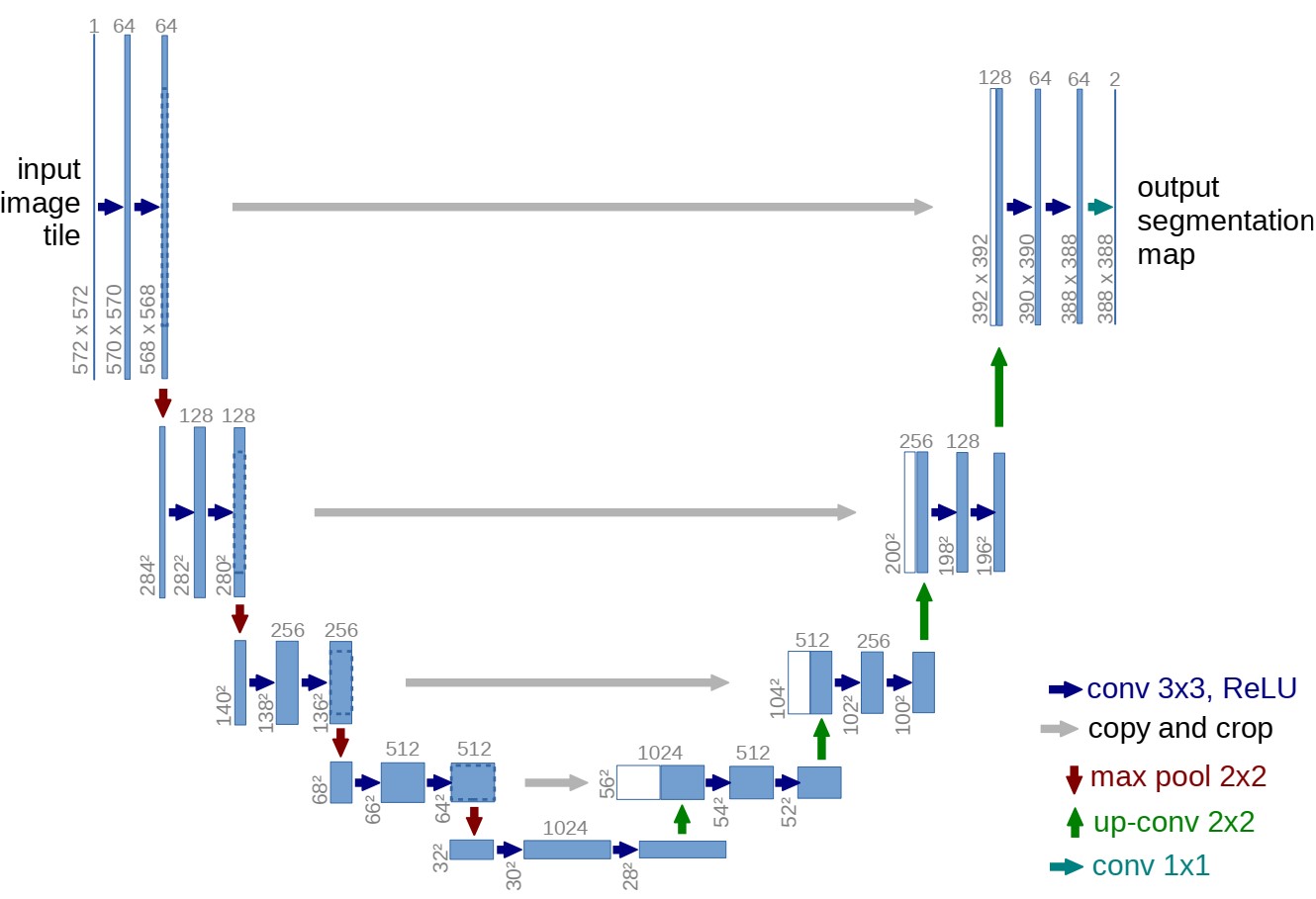}
    \caption[U-Net architecture]{\enquote{U-Net architecture (example for 32x32 pixels in the lowest resolution). Each blue box corresponds to a multi-channel feature map. The number of channels is denoted on top of the box . The x-y-size is provided at the lower left edge of the box. White boxes represent copied feature maps. The arrows denote the different operations}\cite{ronneberger_u-net:_2015}.}
    \label{fig:U-Net_Architecture}
    \end{figure}

The architecture of the original U-Net network is depicted in \autoref{fig:U-Net_Architecture}. It has an encoder-decoder architecture. Its encoder consists of repeated blocks of convolutional layers with \gls{ReLU} activation function, followed by a max-pooling operation. The feature map generated by each block of the encoder is fed to the next block of the encoder as well as the corresponding block in the decoder\cite{ronneberger_u-net:_2015}.

The improved variant of the U-Net network has supplementary layers of batch-normalization after each convolution (before activation functions) and each concatenation layer (in decoder blocks). Furthermore, it has zero-padded convolution that makes the network output to have the same resolution as the input of the network. Another difference is the presence of an additional step in the encoder and the decoder.

\subsection{Metric and Loss Function}
\begin{figure}[ht]
    \centering
    \includegraphics[width=0.4\linewidth]{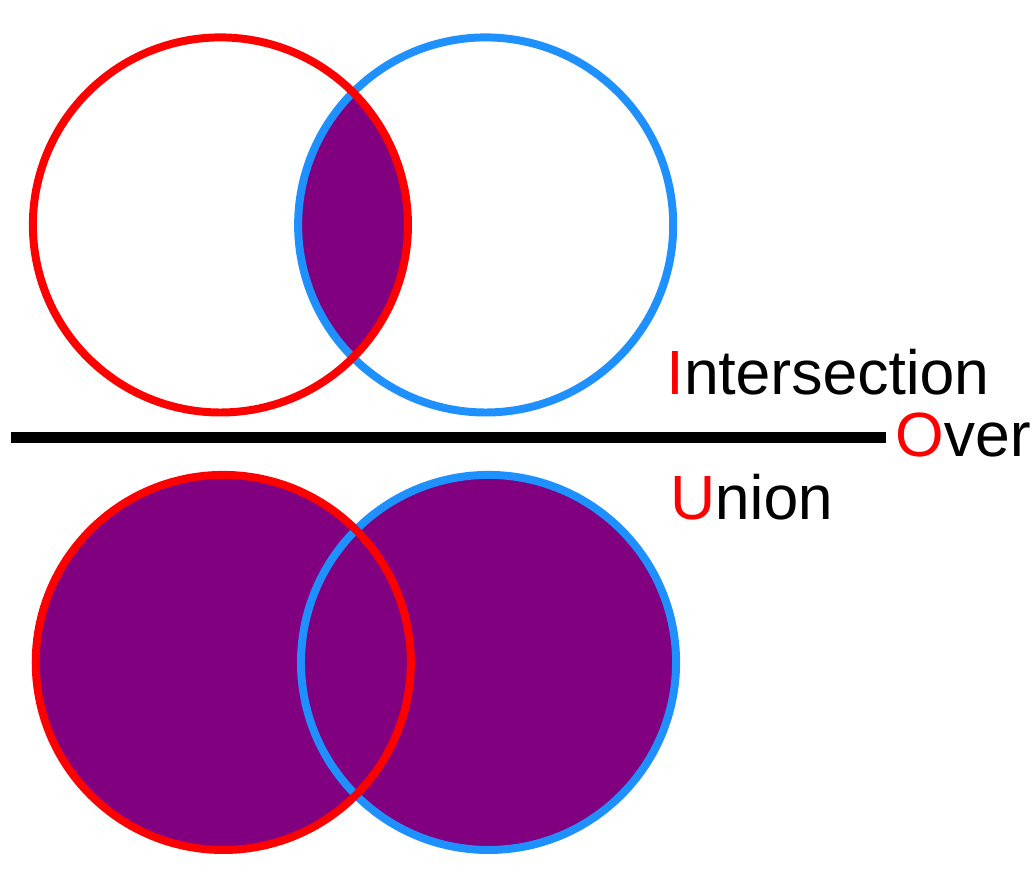}
    \caption{Intersection over Union}
    \label{fig:IOU}
    \end{figure}

\gls{IoU} is used as the metric for the evaluation of semantic segmentation \gls{CNN}s. The \gls{IoU} is the area of intersection between the predicted segmentation and its ground truth divided by the area of their union (\autoref{fig:IOU}). Also, \gls{IoU} is used in the loss function for training semantic segmentation models. The output of a semantic segmentation model consists of a matrix with depth $d$. The $d$ is the number of all classes available in the used dataset. The \gls{IoU} is calculated for every layer of the output matrix, then an average of all \gls{IoU} values is calculated that is between zero and one because the \gls{IoU} metric is between zero and one. The \gls{IoU} reaches one as predictions improve. Therefore, $1 - IoU_{average}$ is used as the loss function to have a decreasing loss function as predictions improve.

\subsection{Datasets}
Two non-synthetic datasets (Cityscapes \cite{cordts_cityscapes_2016} and Mapilary Vistas \cite{neuhold_mapillary_2017}) and two synthetic datasets (VIPER \cite{richter_playing_2017} and SYNTHIA-SF \cite{hernandez-juarez_slanted_2017}) are used in this experimental study. The following data preparations were performed:
\begin{itemize}
    \item Images values were normalized to [0, 1] range.
    \item Semantic segment labels were converted from
    RGB mappings to one-hot labels
    (see \autoref{ch:SemanticSegmentation} for a definition
    of one-hot encoding).
    \item Images and semantic segment labels were resized to
    $640 \times 480$.
    \item To make datasets compatible in terms of semantic
    segment classes, only common classes were considered in
    the experimental study.
\end{itemize}

\section{Results}
\label{sec:second_exp_results}

In each of the following four subsections, the performance of a U-Net network on different datasets is examined. If the performance of the network is uniform over all datasets, the training dataset of the network is comprehensive relative to other datasets. In each subsection, the training dataset is fixed. In the last subsection (fifth), performances of differently trained U-Net networks (i.e., they were trained on different datasets) are compared on the non-synthesized datasets to determine to what extent and conditions synthesized datasets may be replaced non-synthesized ones.

\subsection{Cityscapes as the Training Dataset}
First, a U-Net \gls{CNN} is trained on the Cityscapes dataset information, and then it is tested on all other datasets considered in this study. Results showed that the trained network has some ability to discover semantic segments in the test subset of the Cityscapes dataset. However, the network performance is not satisfying when tested on other datasets. The visual performance results of the U-Net for the test sets of Cityscapes, Mapillary, VIPER, and SYNTHIA-SF datasets are depicted in \autoref{fig:visual_Cityscapes_Cityscapes} to \autoref{fig:visual_Cityscapes_SYNTHIA-SF} respectively (in \autoref{ch:Experiments_Results}).

\begin{figure}
    \centering
    \includegraphics[width=\linewidth]{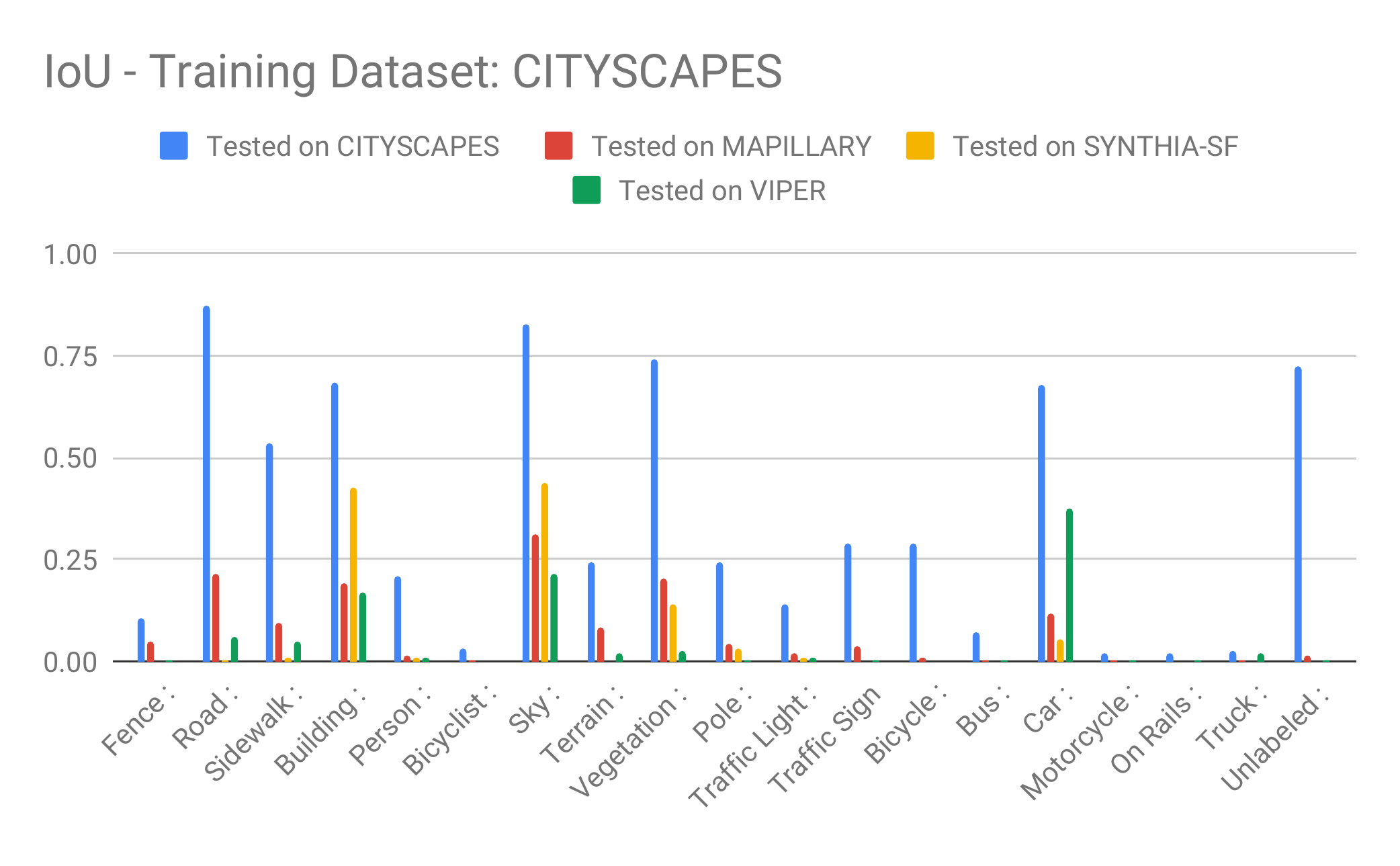}
    \caption[Performance of U-Net trained on Cityscapes]{Performance of U-Net trained on \emph{Cityscapes}. The network is weak in generalization of concepts to other datasets.}
    \label{fig:IoUTrainingDatasetCITYSCAPES}
    \end{figure}

For quantitative scrutiny, a comparison of \gls{IoU} values for each semantic class is performed. The \gls{IoU} values for different datasets are depicted in \autoref{fig:IoUTrainingDatasetCITYSCAPES}. Each bar represents the accuracy of the network in segmenting a specific category. For example, when the network is tested on Cityscapes dataset, the accuracy is more than $75\%$ for two classes of \emph{Road} and \emph{Sky}, and between $75\%$ and $50\%$ for \emph{Sidewalk}, \emph{Building}, \emph{Vegetation}, \emph{Car}, and \emph{Unlabeled} categories. The accuracy for other categories is lower than $50\%$ because of their complexity for segmentation or because they appeared less often in training set relative to other categories. However, the important point is that the accuracy of the network in the first database, whatever it is, is half or less in other datasets. Even this observation is true for two classes of \emph{Road} and \emph{Sky}, which have the highest accuracy in the Cityscapes dataset. We expected that the network to perform well on other non-synthetic datasets due to their common origin. However, the results showed that the Cityscapes dataset is not suitable enough to prepare a semantic segmentation \gls{CNN} to operate in different environments.

\subsection{MAPILLARY as the Training Dataset}
\begin{figure}
    \centering
    \includegraphics[width=\linewidth]{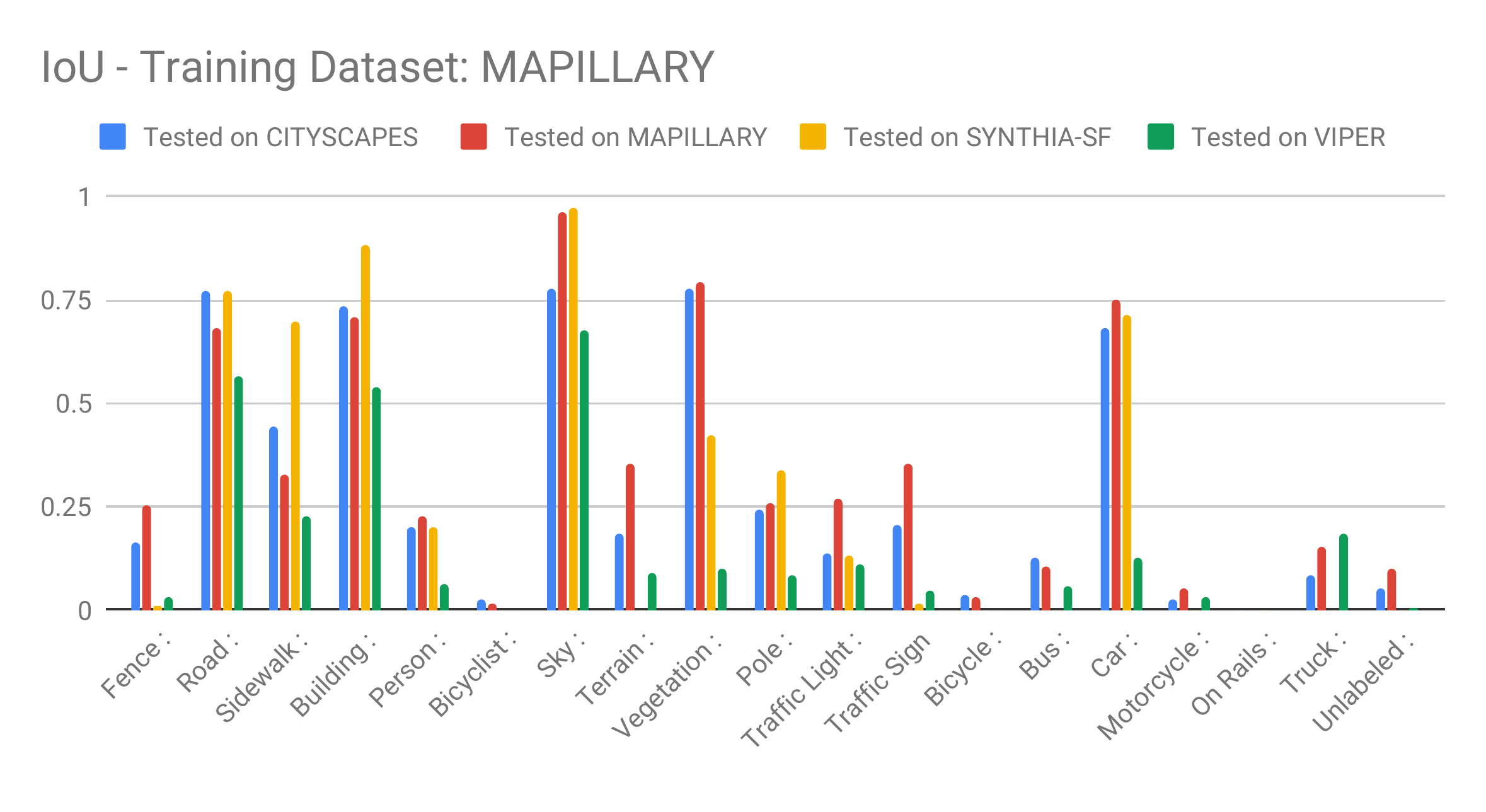}
    \caption[Performance of U-Net trained on MAPILLARY]{Performance of U-Net trained on \emph{MAPILLARY}. The network is fairly able to generalize learned concepts to other datasets.}
    \label{fig:IoUTrainingDatasetMAPILLARY}
    \end{figure}

Now a U-Net \gls{CNN} is trained on the MAPILLARY dataset information, and then it is tested on all other datasets in hand. Unlike the previous network, this network has a more uniform performance on different datasets as seen in \autoref{fig:IoUTrainingDatasetMAPILLARY}. For example, the accuracy of segmentation in \emph{Road}, \emph{Building}, \emph{Sky} categories is uniform over all datasets and is partially uniform over some datasets in \emph{Sidewalk}, \emph{Pole}, \emph{Car}, \emph{Person} categories. Even in some categories, the accuracy of segmentation is higher on datasets other than MAPILLARY, that show how comprehensive the MAPILLARY dataset is relative to other (\emph{Road}, \emph{Sidewalk}, and \emph{Building}). The results also show that this network has some difficulties when predicting on synthesized VIPER datasets. Overall, these results showed that the Mapillary dataset permits the U-Net \gls{CNN} to generalize its estimation in different environments in a more uniform result.  The visual performance results of the U-Net for the test sets of Cityscapes, Mapillary, VIPER, and SYNTHIA-SF datasets are depicted in  \autoref{ch:Experiments_Results} and in \autoref{fig:visual_Mapillary_Cityscapes} to \autoref{fig:visual_Mapillary_SYNTHIA-SF}, respectively.

\subsection{SYNTHIA-SF as the Training Dataset}
\begin{figure}
    \centering
    \includegraphics[width=\linewidth]{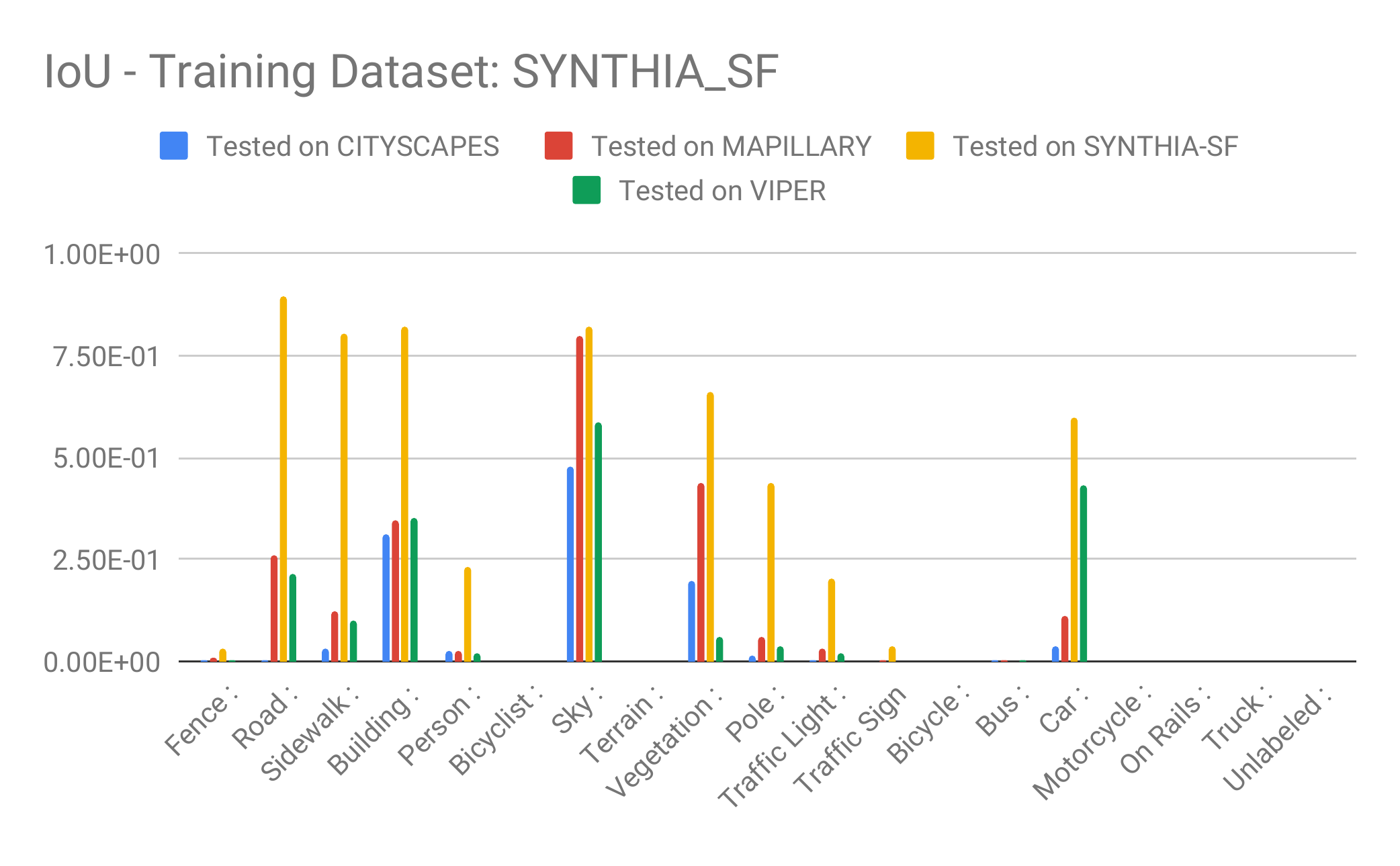}
    \caption[Performance of U-Net trained on SYNTHIA-SF]{Performance of U-Net trained on \emph{SYNTHIA-SF}. The network failed in generalizing learned concepts on other datasets. }
    \label{fig:IoUTrainingDatasetSYNTHIA-SF}
    \end{figure}

Next a U-Net \gls{CNN} is trained on the SYNTHIA-SF dataset, and then it is tested on all other datasets. The accuracy of the network in several categories is zero or near zero which is a sign of absence of these categories in the training set of SYNTHIA-SF dataset. Like the previous result, when the training set was from Cityscapes dataset, we observed that this network is poor to extend its prediction over different other datasets and only performed well on segmentation of \emph{sky}, as it can be seen in \autoref{fig:IoUTrainingDatasetSYNTHIA-SF}. In summary, these results showed the SYNTHIA-SF dataset is not appropriate for training a U-Net \gls{CNN} that is supposed to operate in different environments. The visual performance results of the U-Net for the test sets of Cityscapes, Mapillary, VIPER, and SYNTHIA-SF datasets are depicted in \autoref{ch:Experiments_Results} and in \autoref{fig:visual_SYNTHIA-SF_Cityscapes} to \autoref{fig:visual_SYNTHIA-SF_SYNTHIA-SF}, respectively.

\subsection{VIPER as the Training Dataset}
After a U-Net \gls{CNN} was trained on the VIPER dataset, it was tested on all datasets. Like the results from the network that was trained on MAPILARY dataset, the obtained results for VIPER dataset showed that the network can generalize some of its learned concepts from the training sets to other test sets (\autoref{fig:IoUTrainingDatasetSYNTHIA-SF}). The network extends well its prediction of building, sky, and vegetation classes according to this graph. Furthermore, the accuracy of segmentation is more uniform in other categories relative to the accuracy of the networks that were trained on the Cityscapes or SYNTHIA-SF dataset. A more detailed comparison of the performance of different networks is elaborated in the next section. The visual performance results of the U-Net for the test sets of Cityscapes, Mapillary, VIPER, and SYNTHIA-SF datasets are depicted in \autoref{ch:Experiments_Results} and in \autoref{fig:visual_VIPER_Cityscapes} to \autoref{fig:visual_VIPER_SYNTHIA-SF}, respectively.

\begin{figure}
    \centering
    \includegraphics[width=\linewidth]{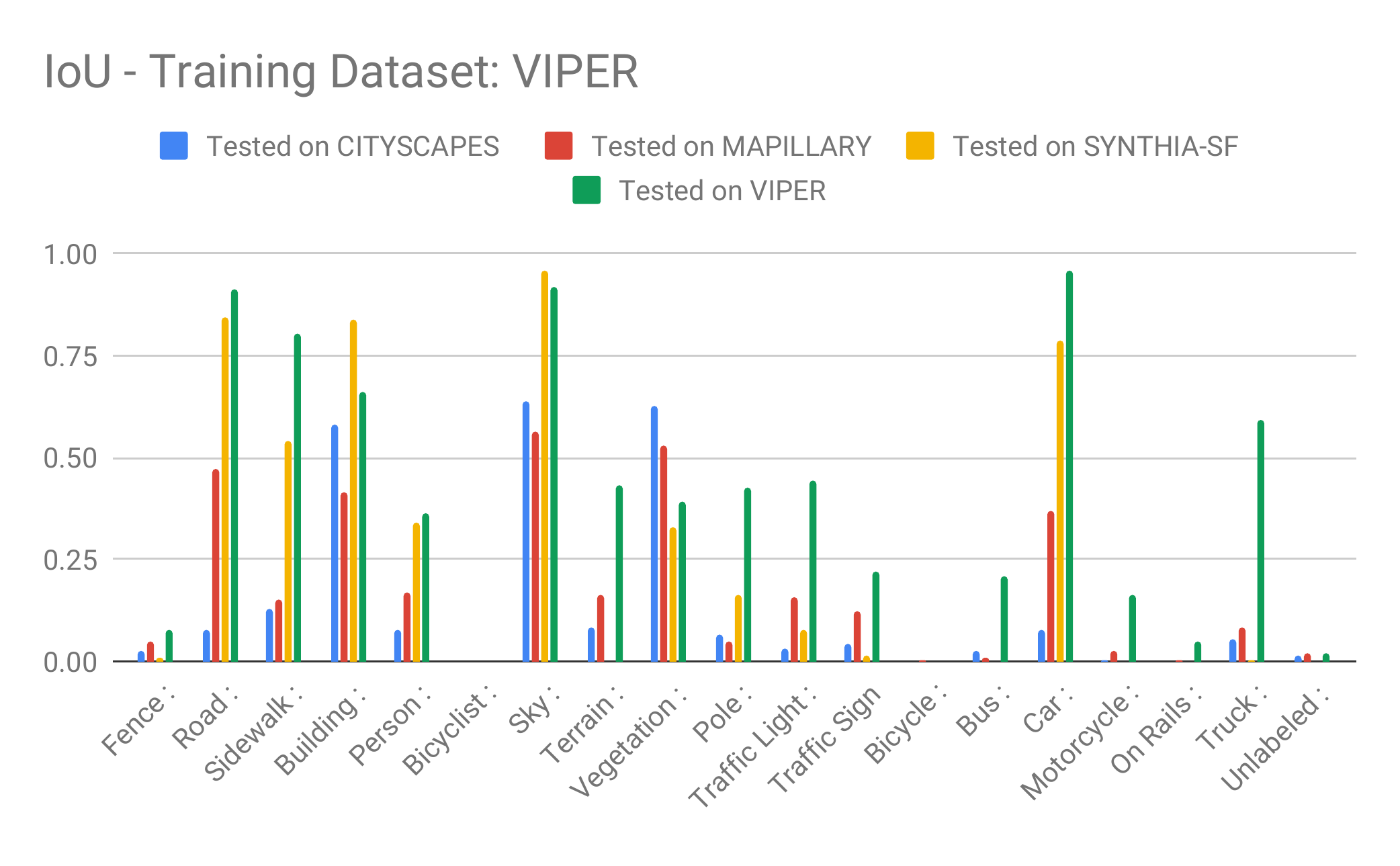}
    \caption[Performance of U-Net trained on VIPER]{Performance of U-Net trained on \emph{VIPER}. The network succeed in generalizing of some classes (building, sky, vegetation classes) and failed in others}
    \label{fig:IoUTrainingDatasetVIPER}
    \end{figure}

\subsection{Comparison on Non-Synthesized Datasets}
\begin{figure}[h]
    \centering
    \includegraphics[width=\linewidth]{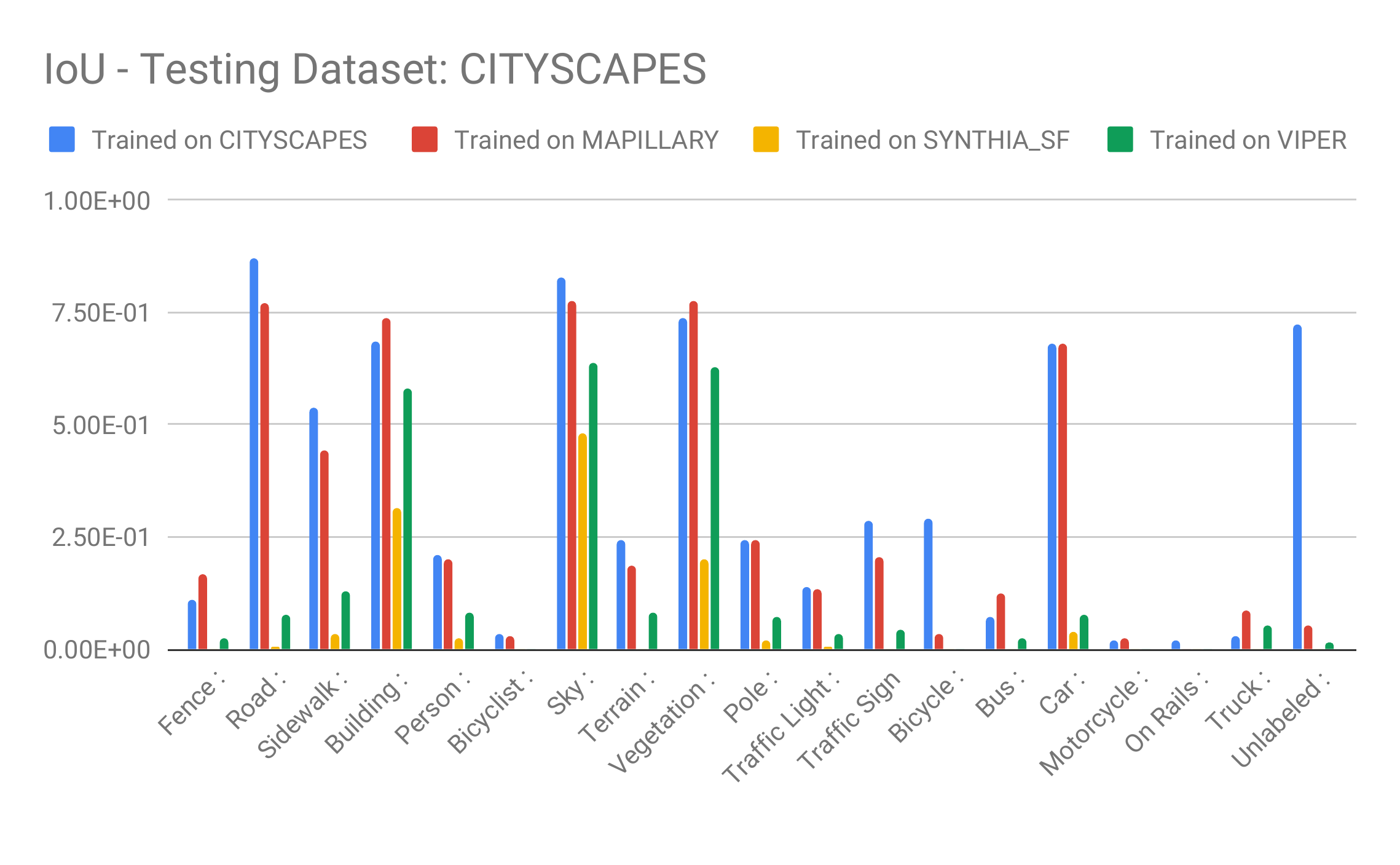}
    \caption[Performance of different U-Net networks on CITYSCAPES]{Performance of different U-Net networks on CITYSCAPES}
    \label{fig:IoUTestingDatasetCITYSCAPES}
    \end{figure}

\begin{figure}[ht]
    \centering
    \includegraphics[width=\linewidth]{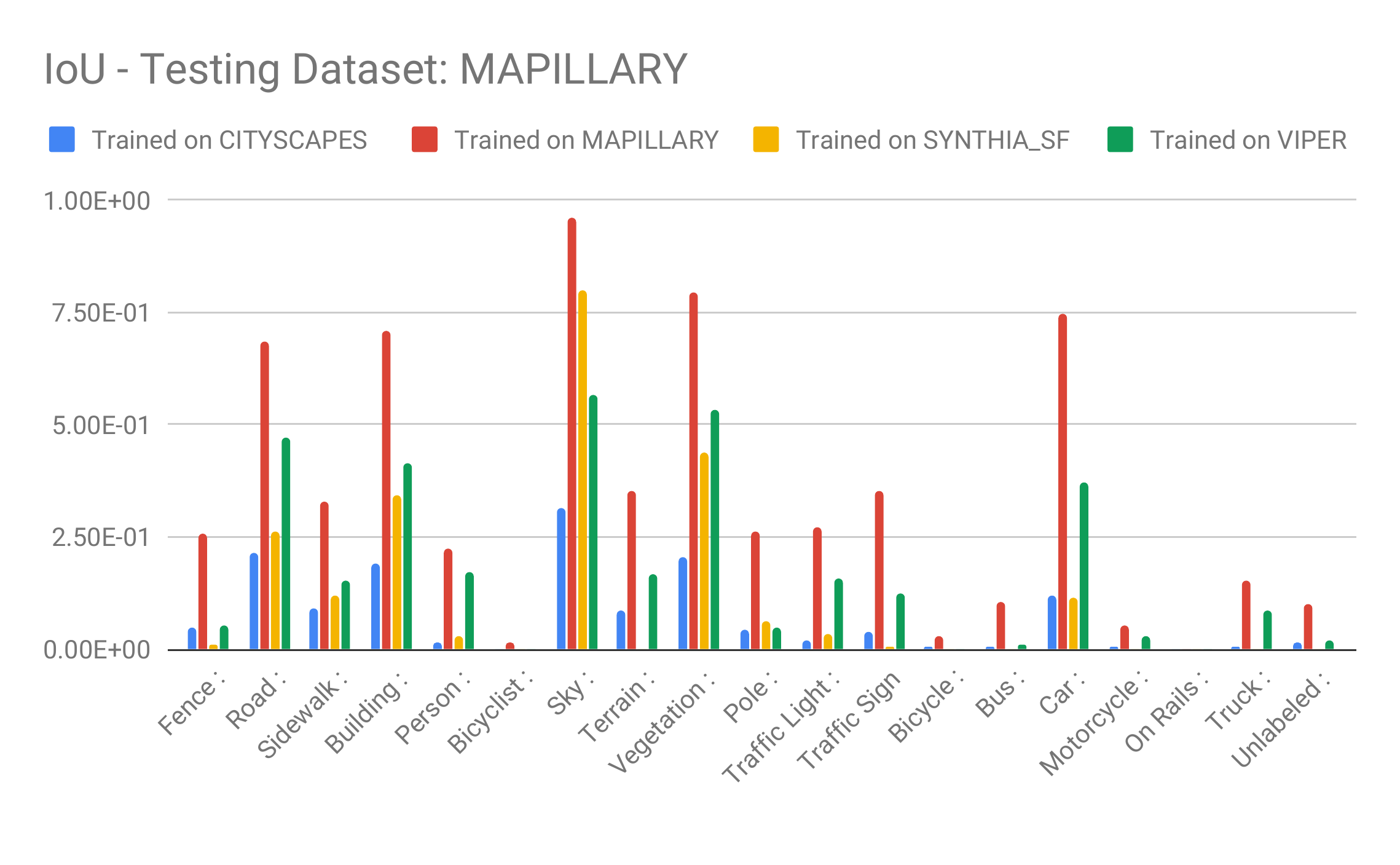}
    \caption[Performance of different U-Net networks on MAPILLARY]{Performance of different U-Net networks on MAPILLARY}
    \label{fig:IoUTestingDatasetMAPILLARY}
    \end{figure}

Two different graphs with duplicate data from previous graphs are considered for this comparison. In this comparison, results are related to a common test dataset instead of a common training dataset. This comparison contains \gls{IoU} scores of all four trained \gls{CNN}s, while evaluated only on two non-synthesized datasets. \autoref{fig:IoUTestingDatasetCITYSCAPES} shows the performance of differently trained networks when tested on Cityscapes dataset. The best performance as expected belongs to the \gls{CNN} that is trained on the Cityscapes, since by this means there is no domain shift. The surprising fact is that second-best \gls{CNN}  which is superior in 8 classes is trained on MAPILLARY dataset. To further explore the  characteristics of these datasets, \autoref{fig:IoUTestingDatasetMAPILLARY} shows the performance of differently trained \gls{CNN}s when tested on the MAPILLARY dataset. In this graph, all superior results belong to the \gls{CNN} that is trained on the MAPILLARY itself as expected, while the second-best performance is on the \gls{CNN} which is trained on a \emph{synthesized} dataset, on VIPER! The worst result belongs to the \gls{CNN} which is trained on a non-synthesized dataset, i.e. the Cityscapes.

\subsection{Discussion on the Observed Results}
In the first and third experiments, we observed that networks are poor in extending their predictions over different datasets. The network in the first experiment was trained on Cityscapes dataset. This dataset was captured in different cities of Germany and contains 2975 samples. Therefore, this dataset contains no information about urban areas of other countries, different types of vegetation, different types of sun lighting in different latitudes, and any other objects that are subject to change due to a change in location. Similar conditions apply to the SYNTHIA-SF dataset, which led to the results of the third experiment. This dataset was captured in a small uniform virtual environment and contains 2224 samples.

On the other hand, in the second experiment, we observed that the network performs well in extending its predictions over different datasets, and somewhat well in the fourth experiment. The MAPILARY datasets, that was used as the reference dataset in the second experiment, was captured by different imaging devices like mobile phones, tablets, action cameras, and professional capturing rigs and in cities of different countries and continents, and contains 25000 samples. The VIPER dataset, that was used as the reference dataset in the fourth experiment, sits between uniform datasets and non-uniform datasets. It was captured in a large virtual environment (about 8.12 square kilometers) that has different cities, jungles, seas, deserts, and mountains, and contains 250000 samples.

Finally, we concluded that the diversity in the content of datasets' samples is very influential in the performance of segmentation networks and the observed performance differences mainly stem from differences in the diversity of contents in used datasets. Although other factors like the realness of synthesized images or quality of images in terms of the resolution also affect the performance of the networks, the obtained results showed that these factors are minor here. Therefore, in cases where appropriate non-synthetic datasets (like Mapillary) is not available, one may use a synthetic dataset with diverse contests and large samples (like VIPER).

    \chapter{Robustness of SSIDE against Domain Shift}

This chapter describes the final experimental study which is based on the results of the two previous experimental studies: first, it is assumed that an \gls{SSIDE} \gls{CNN} exists that outperforms its \gls{SIDE} counterpart. Second, it is assumed that semantic segments generated for non-synthetic images by a \gls{CNN} trained on synthetic datasets are accurate enough. The goal of this experimental study is to investigate the effect of the use of semantic segments on the robustness of a \gls{SIDE} model against domain shift. In other words, the goal is to determine how much an \gls{SSIDE} \gls{CNN} outperforms its \gls{SIDE} counterpart while both generalize concepts from synthetic data to non-synthetic data.

\section{Methods}
\label{sec:third_exp_method}

For the goal of this study, multiple models of the M20 architecture are trained on two non-synthetic datasets (\emph{KITTI} \cite{uhrig_sparsity_2017} and \emph{Lyft Level 5} \cite{noauthor_lyft_nodate}) and three synthetic datasets (\emph{Virtual KITTI 2} \cite{cabon_virtual_2020}, \emph{PreSIL} \cite{hurl_precise_2019}, and \emph{SYNTHIA-SF} \cite{hernandez-juarez_slanted_2017}). Several other M20 models are also trained on different combinations of synthetic datasets. Furthermore, multiple models of the M18 and the M21 architecture are trained on three synthetic datasets (\emph{Virtual KITTI 2}, \emph{PreSIL} , and \emph{SYNTHIA-SF}) and on different combinations of synthetic datasets.

All networks are tested on non-synthetic datasets (KITTI and Lyft Level 5) after every 5000 batches during training for 30 times. Then networks' performances are compared separately on each test dataset (KITTI and Lyft Level 5). Each comparison excluded the performance of the network that the training set of that network and test set, which comparison is based on it, belonged to one dataset.

Since the used non-synthetic datasets have no semantic segments along with the depth data, a U-Net \gls{CNN} is used to generate semantic segments for each sample. Various combinations of synthetic datasets are tested to train multiple U-Net \gls{CNN}s. Among them, a combination is selected that outperformed others on the KITTI semantic segmentation dataset (training set of the dataset).

Finally, a statistical analysis is performed on depth data of all used datasets. This supplementary analysis is carried out to determine the differences and similarities of the depth data distribution between all used datasets. In this analysis, the histograms of the depth distribution for each row of depth maps are calculated (on all samples of each dataset). The range size and the bin size of histograms are 100 $m$ and 0.2 $m$, respectively. The size of depth maps is $1216 \times 352$, and therefore, the number of rows is 352. For each dataset, instead of drawing 352 histograms, a heat map is generated and reported. The histograms constituted the rows of the heat maps, such that the heat maps have 352 rows and 500 columns ($500 = size_{range} / size_{bin} = 100 / 0.2$).

\subsubsection{Trainings Details}
The loss function, batch size, optimizer, and learning rate for the training of all depth estimation models were \gls{MAPE}, two, Adam optimizer \cite{kingma_adam_2015}, and $0.001$, respectively.

\subsection{Datasets}
To make datasets consistent in terms of depth data, the \gls{AFOV}s of all samples are reduced to a specific \gls{AFOV}. It is carried out by cropping margins of samples. After the cropping, all samples are resized to $1216 \times 352$. Initial and final \gls{AFOV} of each datasets is mentioned in \autoref{table:AFOV_unification}.

\begin{table}
    \caption[Unification of AFOV between datasets]{Unification of AFOV between datasets. \emph{Focal Length} column shows the focal length of the camera used in each dataset. \emph{Original size} column shows the original size of samples in each dataset (images, semantic segments, and depth maps). The \gls{AFOV}s of all samples were reduced to the \gls{AFOV} of samples of Lyft level 5 dataset.}
    \label{table:AFOV_unification}
    \begin{tabular}{lccccccc}
    \multirow{2}{*}{Dataset Name} & Focal & \multicolumn{2}{c}{Original size} & \multicolumn{2}{c}{Cropped size} & \multicolumn{2}{c}{Final AFOV} \\ \cline{3-8}
                    & Length & width & height & width & height & horizontal & vertical \\ \hline
    Lyft Level 5    & 880.0  & 1224  & 1024   & 1216  & 352    & 69.28      & 22.62    \\ \hline
    PreSIL          & 960.0  & 1920  & 1080   & 1328  & 384    & 69.34      & 22.62    \\ \hline
    SYNSCAPES       & 1590.0 & 1440  & 416    & 1440  & 352    & 48.72      & 14.91    \\ \hline
    SYNTHIA-SF      & 847.6  & 1920  & 1080   & 1172  & 340    & 69.32      & 22.68    \\ \hline
    Virtual KITTI   & 725.0  & 1242  & 375    & 1002  & 290    & 69.29      & 22.62    \\ \hline
    VIPER           & 1158.0 & 1920  & 1080   & 1602  & 464    & 69.34      & 22.66    \\ \hline
    KITTI           & 721.5  & 1242  & 375    & 998   & 289    & 69.3       & 22.6
    \end{tabular}
    \end{table}

Five datasets are used for the training of the U-Net. These five datasets are KITTI (for the test), Virtual KITTI 2, VIPER, Synscapes \cite{wrenninge_synscapes_2018}, and SYNTHIA-SF. To have unified semantic categories between these datasets, those subcategories which are not available in all datasets are merged to their higher categories. Among these subcategories, some did not have a higher category, and therefore, they are removed (see \autoref{ch:CommonSemanticLabelsSet} for details).

\section{Results}
\label{sec:third_exp_results}

\subsection{Statistical Analysis of Depth Data}
\begin{figure}
    \includegraphics[trim={2.7cm 2.4cm 3cm 5.5cm}, clip, width=\linewidth]{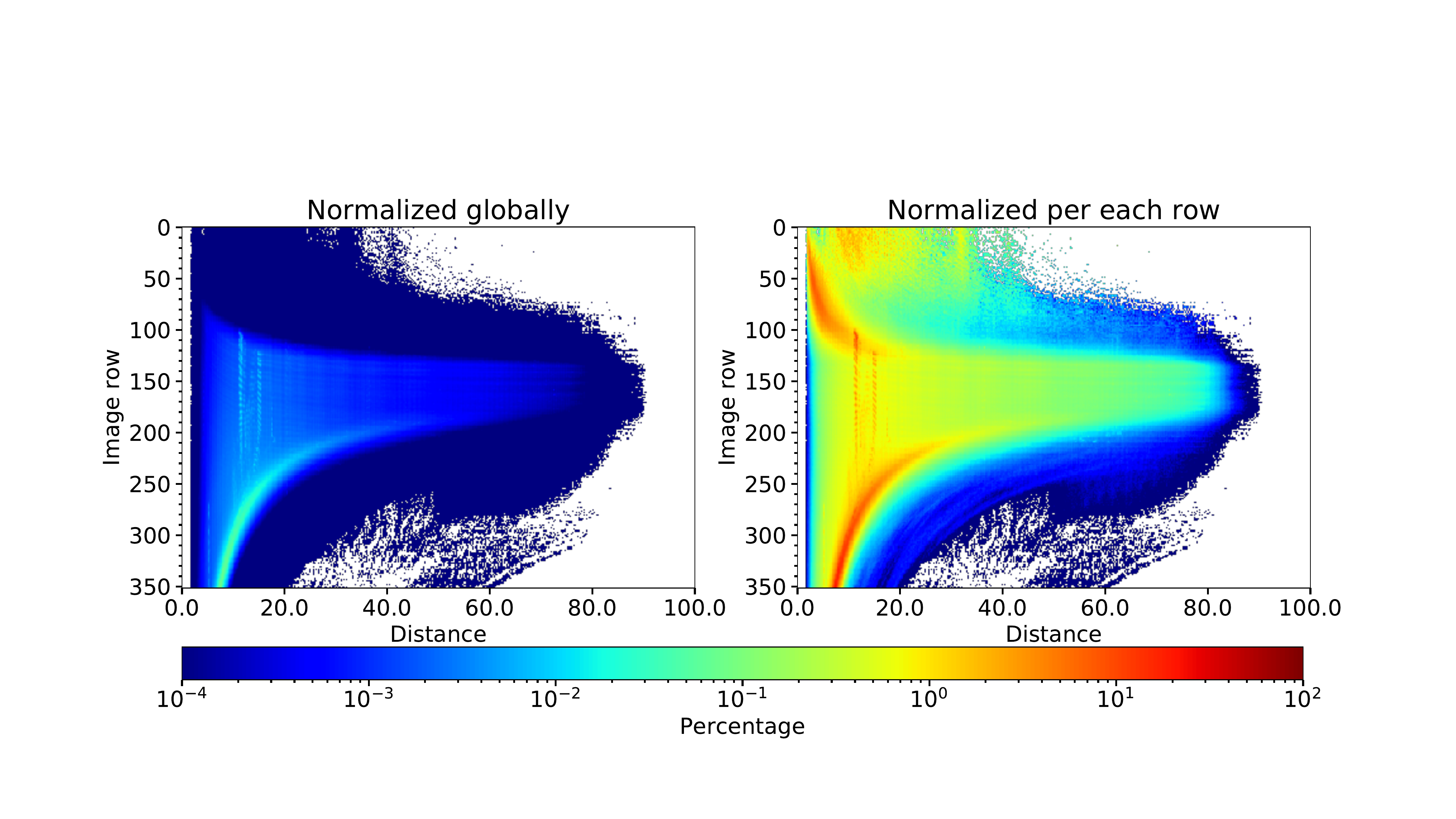}
    \caption[Depth heat map of KITTI dataset]{Depth heat map of KITTI dataset. }
    \label{fig:depth_heat_map_kitti_NewSetLyft}
\end{figure}

\begin{figure}
    \includegraphics[trim={2.7cm 2.2cm 3cm 5.5cm}, clip, width=\linewidth]{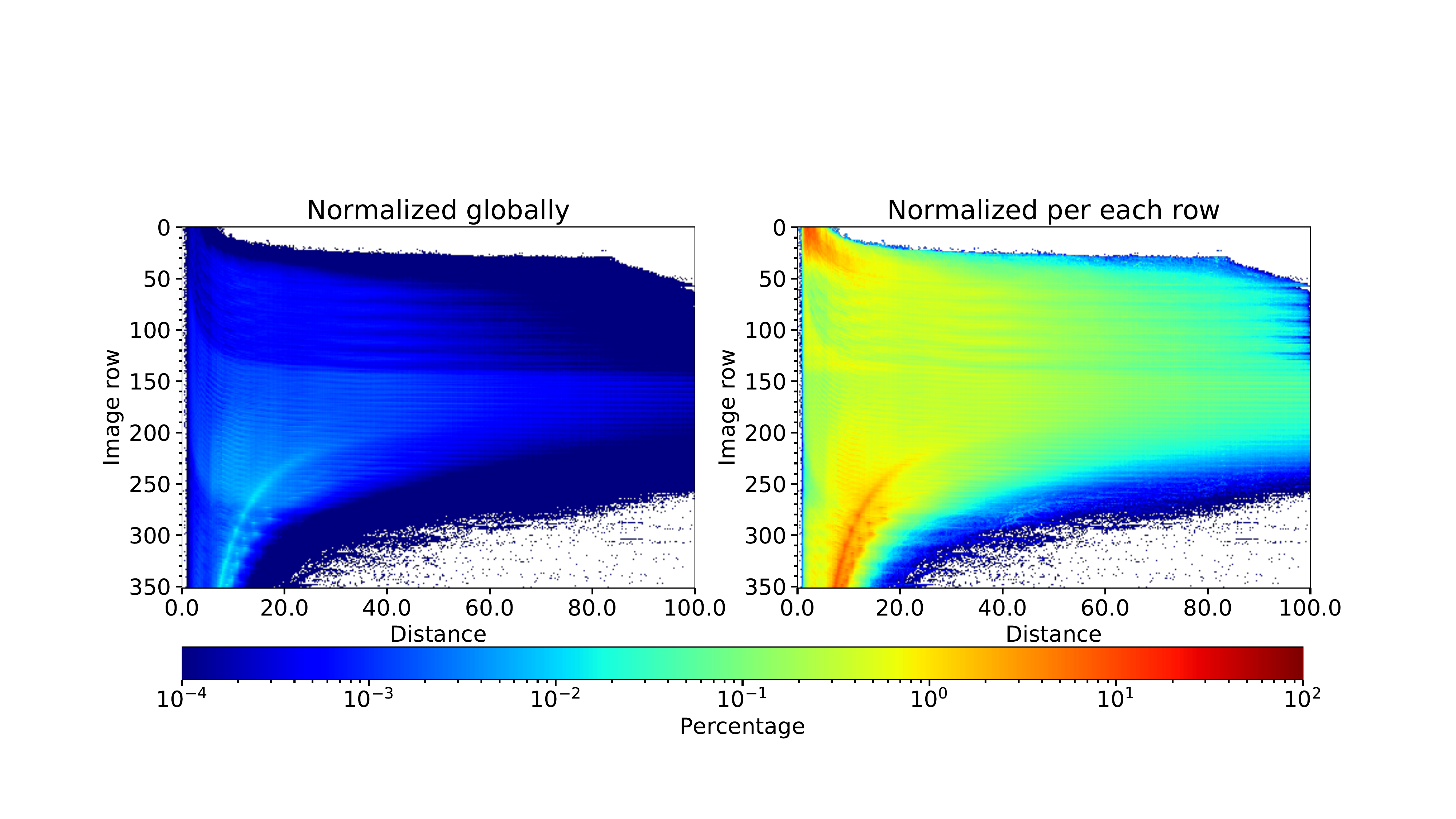}
    \caption[Depth heat map of Lyft Level 5 dataset]{Depth heat map of Lyft Level 5 dataset}
    \label{fig:depth_heat_map_lyft_NewSetLyft}
\end{figure}

\begin{figure}
    \includegraphics[trim={2.7cm 2.2cm 3cm 5.5cm}, clip, width=\linewidth]{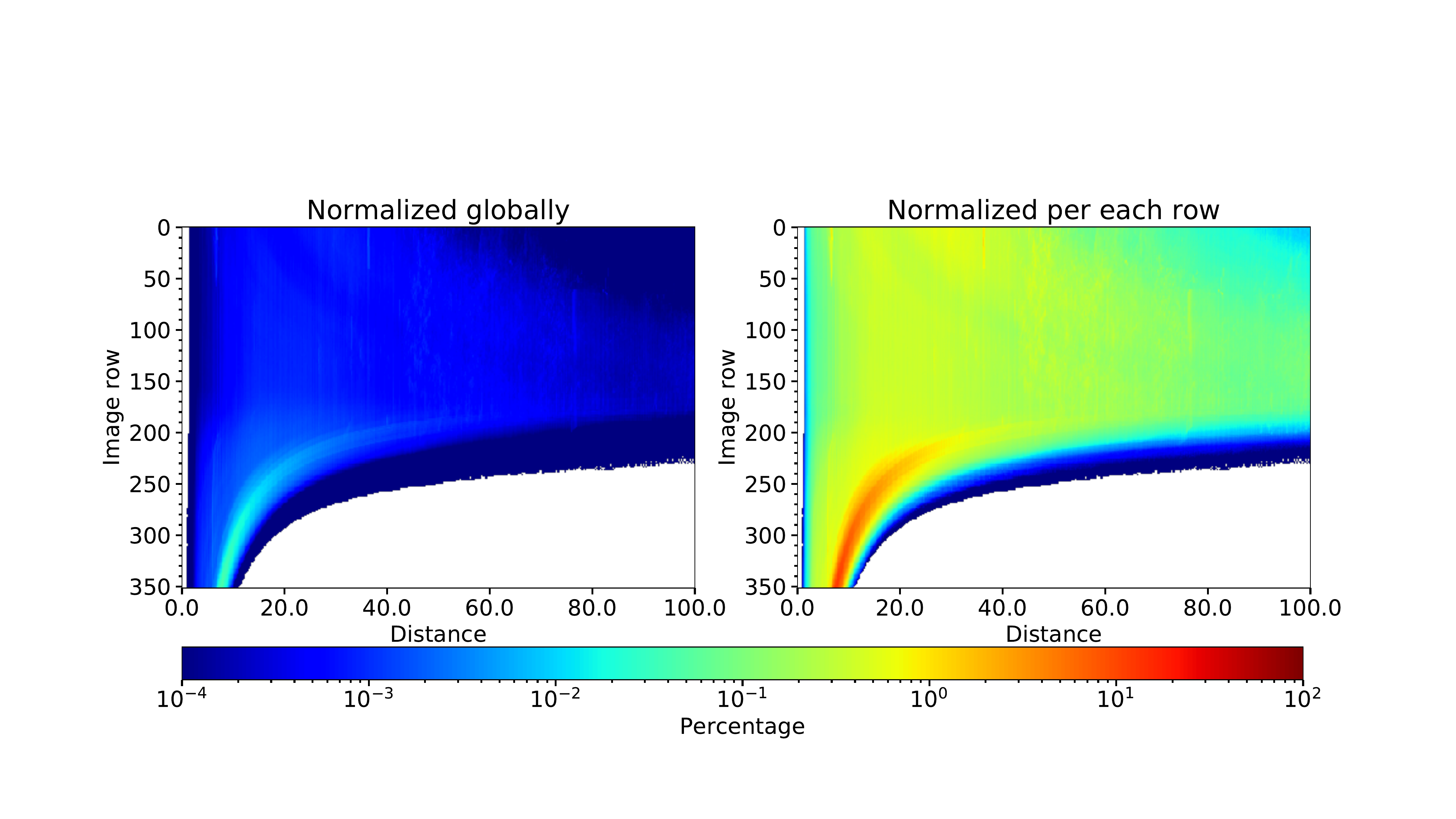}
    \caption[Depth heat map of Virtual KITTI 2 dataset]{Depth heat map of Virtual KITTI 2 dataset}
    \label{fig:depth_heat_map_vkitti_NewSetLyft}
\end{figure}

\begin{figure}
    \includegraphics[trim={2.7cm 2.2cm 3cm 5.5cm}, clip, width=\linewidth]{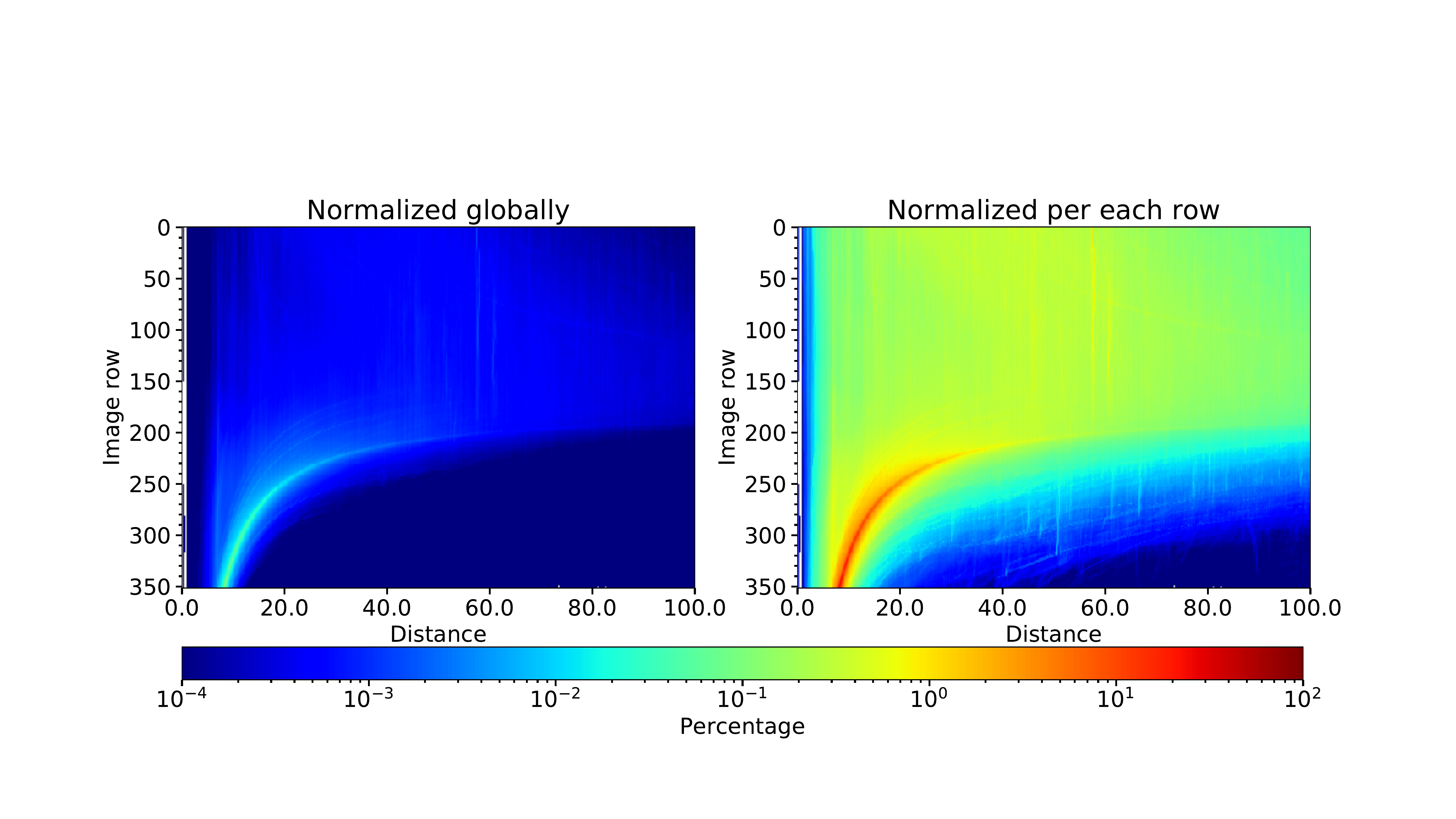}
    \caption[Depth heat map of PreSIL dataset]{Depth heat map of PreSIL dataset}
    \label{fig:depth_heat_map_presil_NewSetLyft}
\end{figure}

\begin{figure}
    \includegraphics[trim={2.7cm 2.2cm 3cm 5.5cm}, clip, width=\linewidth]{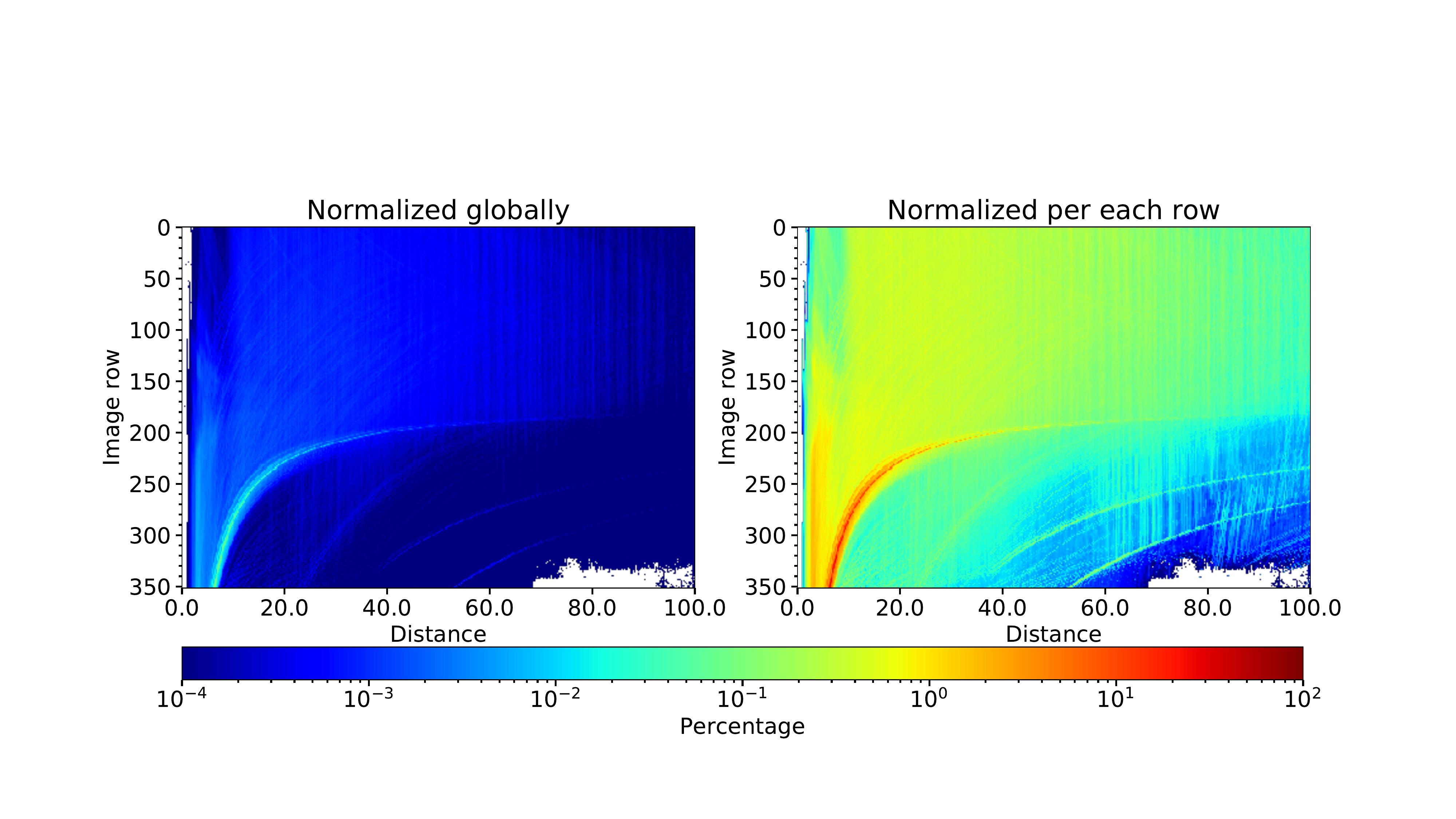}
    \caption[Depth heat map of SYNTHIA-SF dataset]{Depth heat map of SYNTHIA-SF dataset}
    \label{fig:depth_heat_map_synthiasf_NewSetLyft}
\end{figure}

Among the models that were trained on synthetic datasets, the best performances belonged to the models that were trained on the Virtual KITTI dataset. The statistical analysis of the depth distribution of the datasets could be used to explain the underlying facts behind this observation. The depth heat maps shows that the distribution of depth in the synthetic Virtual KITTI dataset is more similar to the distribution of depth in the non-synthetic datasets relative to other synthetic datasets. These heat maps are depicted in \autoref{fig:depth_heat_map_kitti_NewSetLyft} to \autoref{fig:depth_heat_map_synthiasf_NewSetLyft}. In the left heat map of each figure, the sum of percentages of all points is 100 (globally normalized), while in the right heat maps, the sum of percentages of all points in each row is 100 (normalized per each row).

These heat maps contain significant amount of information about depth distributions. As an example, consider the heat maps of the KITTI dataset in \autoref{fig:depth_heat_map_kitti_NewSetLyft}, which shows that the used \gls{LIDAR} had a range limit of 80 m, while this range in Lyft Level 5 dataset is more than 100 m, as it can be seen in \autoref{fig:depth_heat_map_lyft_NewSetLyft}. For this reason, depth labels of used synthetic datasets were limited to 100 $m$. The existence of large distances in the upper part of the heat maps is a sign of wide streets and existence of large objects like tall buildings and trees in the sampling scene, like in \autoref{fig:depth_heat_map_vkitti_NewSetLyft}, \autoref{fig:depth_heat_map_presil_NewSetLyft}, and \autoref{fig:depth_heat_map_synthiasf_NewSetLyft}. As a similarity to the non-synthetic datasets, the Virtual KITTI dataset has the lowest density of depth distribution in the upper right corner of its heat map among the heat maps of synthetic datasets (it is better visualized in the graphs in the right). The existence of large distances in the lower part of the heat maps shows that some samples of the datasets were captured on uphill streets or on roads along valleys, like in \autoref{fig:depth_heat_map_presil_NewSetLyft}, and \autoref{fig:depth_heat_map_synthiasf_NewSetLyft}. Again, another similarity of Virtual KITTI dataset to non-synthetic datasets is seen here; the density of depth data in lower part of heat maps of Virtual KITTI dataset is concentrated in short distances.

In addition to visual similarity, calculation of the Euclidean distance of these heat maps shows that the distribution of depth in the Virtual KITTI dataset is very similar to the depth distribution in the non-synthetic version of this dataset. Considering these arguments and observations, in the following analysis, the results obtained from Virtual KITTI, KITTI, Lyft level 5, PreSIL, and SYNTHIA-SF datasets are reported.

\subsection{Performance Analysis}
For the performance comparison between \gls{SIDE} architectures and \gls{SSIDE} architectures, the M20 architecture was selected as the representative of \gls{SIDE} networks and M18 and M21 architectures were selected as the representatives of \gls{SSIDE} networks. We selected two architectures of M18 and M21 as the representatives of \gls{SSIDE} networks because their performances were very close to each other when they were trained and tested on the same dataset. However, we considered the possibility that one architecture may perform better when a domain shift occurs. As mentioned before, the M21 architecture uses RGB images as its inputs and the M18 uses semantic edges as its inputs.

\begin{figure}[t]
    \centering
    \begin{subfigure}[b]{.49\textwidth}
        \centering
        \includegraphics[width=\textwidth]{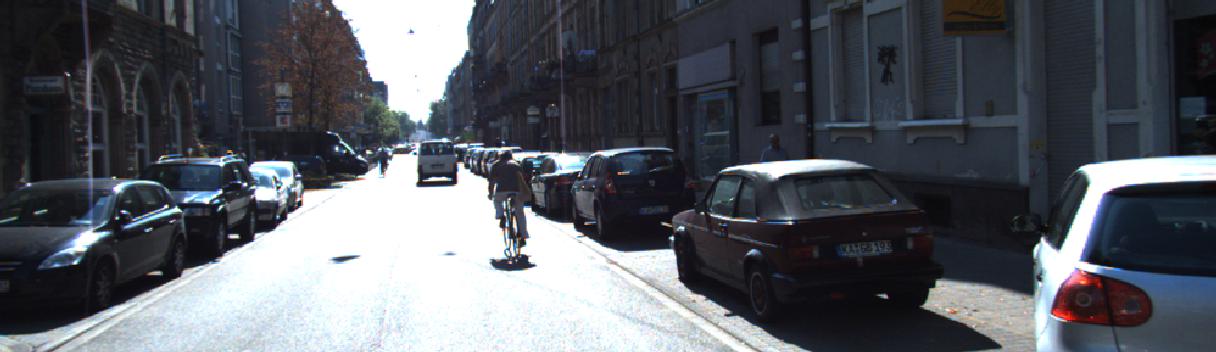}
        \caption{RGB image}
    \end{subfigure}
    \hfill
    \begin{subfigure}[b]{.49\textwidth}
        \centering
        \includegraphics[width=\textwidth]{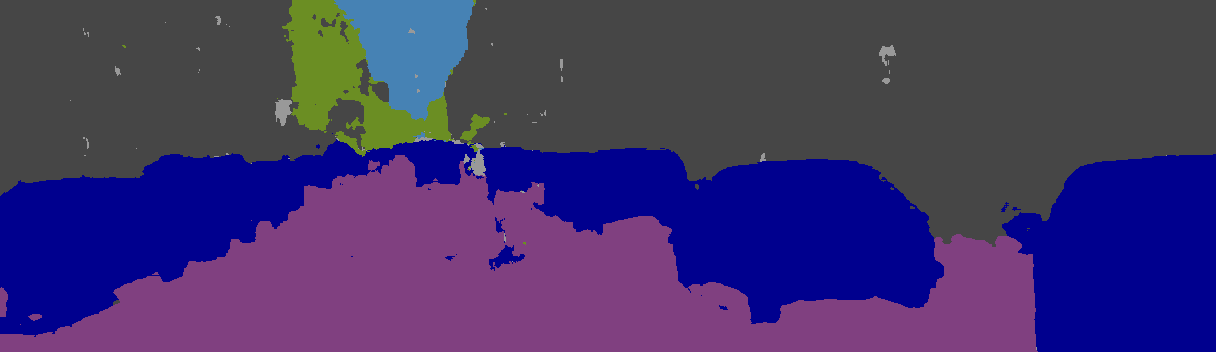}
        \caption{An acceptable semantic label}
        \label{fig:ss_for_KITTI_acceptable}
    \end{subfigure}
    \hfill
    \begin{subfigure}[b]{.49\textwidth}
        \centering
        \includegraphics[width=\textwidth]{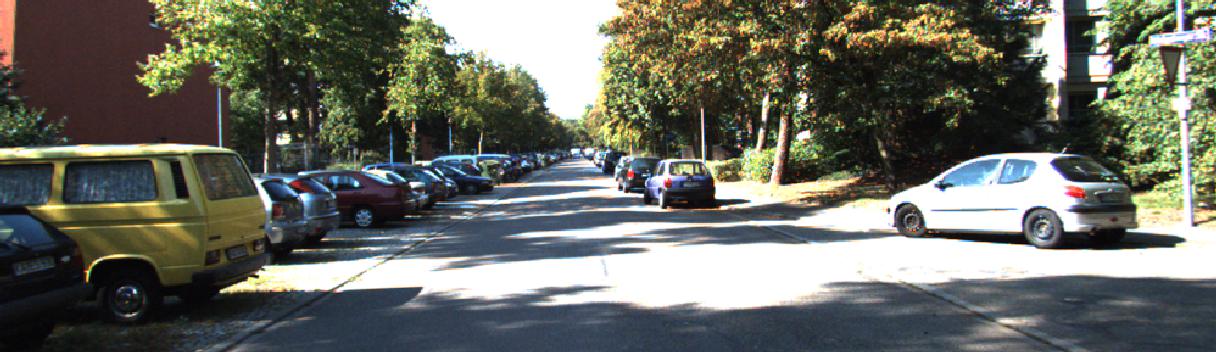}
        \caption{RGB image}
    \end{subfigure}
    \hfill
    \begin{subfigure}[b]{.49\textwidth}
        \centering
        \includegraphics[width=\textwidth]{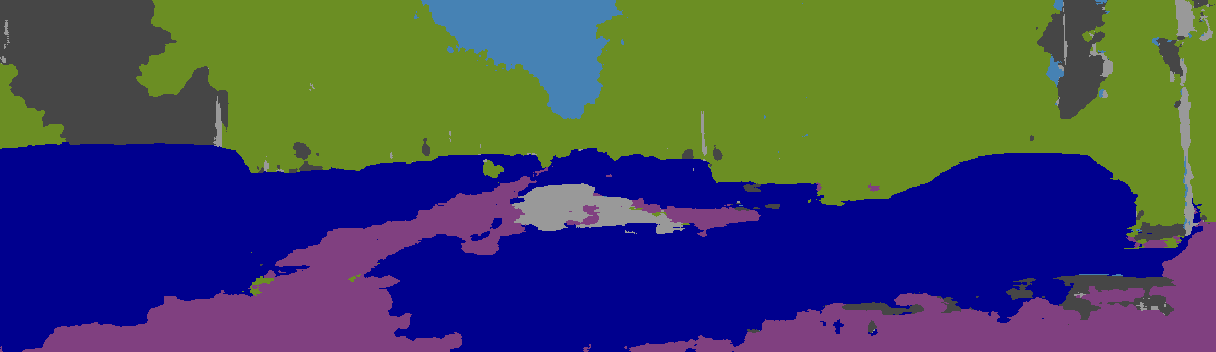}
        \caption{A weak semantic label}
        \label{fig:ss_for_KITTI_weak}
    \end{subfigure}

    \caption[Examples of generated semantic labels for KITTI dataset]{Examples of generated semantic labels for KITTI dataset. b is generated for a and d is generated for c.}
    \label{fig:ss_for_KITTI}
    \end{figure}

For evaluation of an M18 and an M21 network ,which were trained on Virtual KITTI, semantic labels for the test sets of the KITTI and Lyft Level 5 datasets had to be prepared first. After training a U-Net network on different combinations of synthetic datasets, the best performance was achieved when the training set was composed of 10000 samples of VIPER dataset, 10000 samples of SYNSCAPES dataset, 10000 samples of Virtual KITTI dataset, and all samples of SYNTHIA-SF dataset(more the 4000 samples). By using this combination for the training, the U-Net achieved a mean \gls{IoU} of 73\% on the training set of the KITTI dataset (KITTI segmentation dataset consists of 200 samples.). Some samples of the generated semantic labels are depicted in \autoref{fig:ss_for_KITTI}.

In the first sample, minor errors are seen in \autoref{fig:ss_for_KITTI_acceptable}. Among them, the notable error is the semantic label for the bicycle. In the second sample, some gross errors exist as seen in \autoref{fig:ss_for_KITTI_weak}. For example, a large area of the street is labeled as \emph{Car}, which its reason is probably the saturated colors in the image that the network never experienced before in training sets. In the middle of the generated segmentation map, another large area of the street is labeled as \emph{Pole}.

These observation lead us to two important conclusions. First, a comprehensive synthesized dataset along with all kinds of diversities in its samples should contain samples with synthesized artifacts like image distortion, color saturation, and motion blur. This will enable the network to perform better when it faces with artifacts in non-synthetic data. Second, the accuracy of the generated semantic labels may lead the \gls{SSIDE} networks to perform worse than what is expected. Therefore, a minimum accuracy requirement of semantic labels must always be met to have a reliable output from \gls{SSIDE} networks.

\begin{figure}
    \centering
    \includegraphics[trim={2cm 1cm 1cm 2cm}, clip, width=\linewidth]{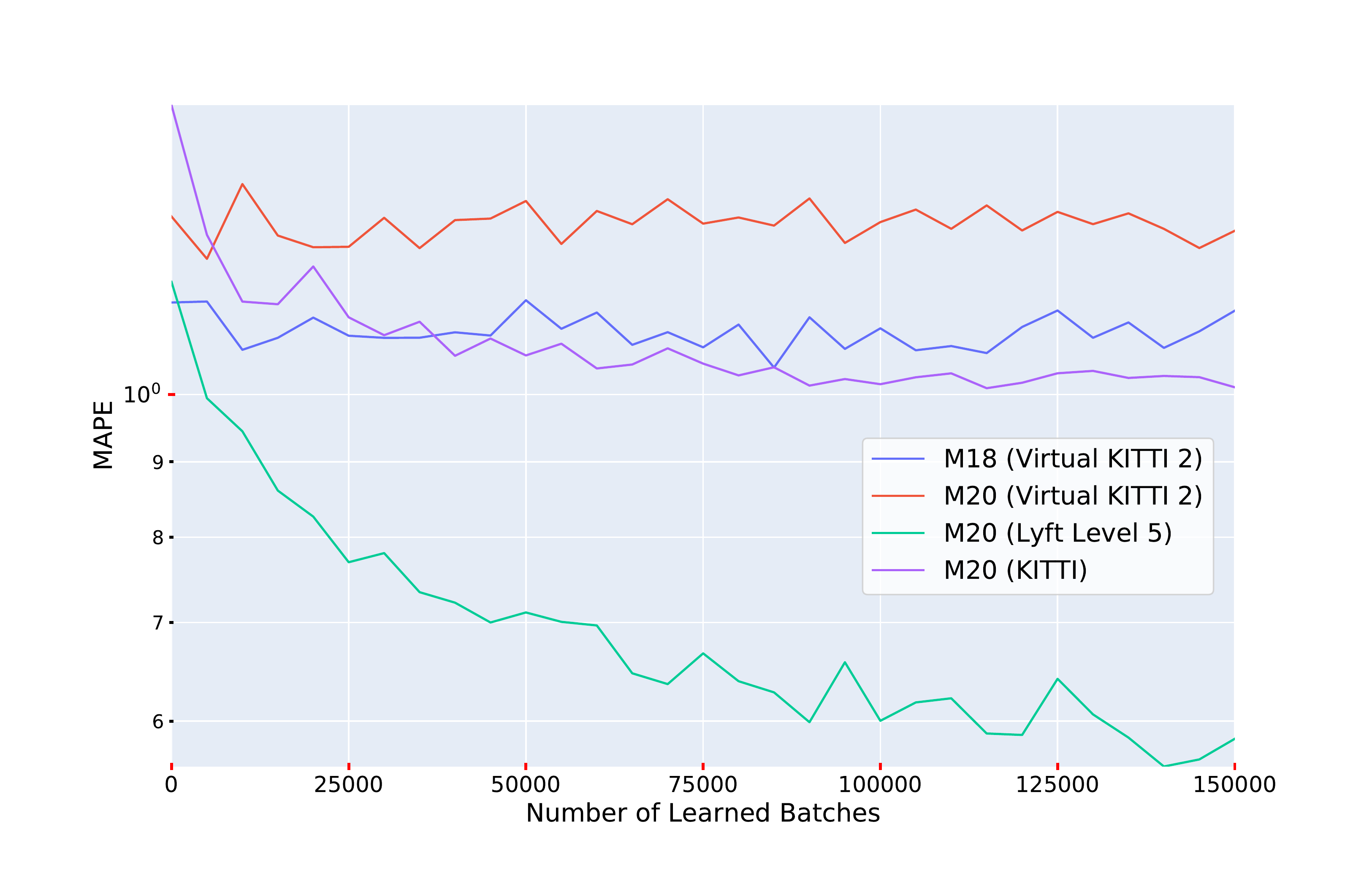}
    \caption[Performance Comparison on the Lyft Level 5 dataset]{Performance Comparison on the Lyft Level 5 dataset}
    \label{fig:SSIDE_domain_shift_lyft}
    \end{figure}
    
To summarize the comparison study of \gls{SSIDE} models performances, we find that unlike the results obtained in the first experimental study (\autoref{sec:firts_exp_results}), the performance of the M18 model is better than the performance of the M21 model by 13 percent with respect to \gls{MAPE}. It indicates that the architecture of M18 is more robust against domain shift relative to the architecture of the M21. In other words, the use of semantic edges\footnote{See the definition of \emph{semantic edge} in \autoref{sub:DepthModel}} is more effective in the robustness of an \gls{SSIDE} model compared to that on the cases when simply concatenation of semantic labels and raw image is used. Therefore, in the upcoming observation reported in this section, we only consider the performance of M18.

\begin{figure}
    \centering
    \includegraphics[trim={2cm 1cm 1cm 2cm}, clip, width=\linewidth]{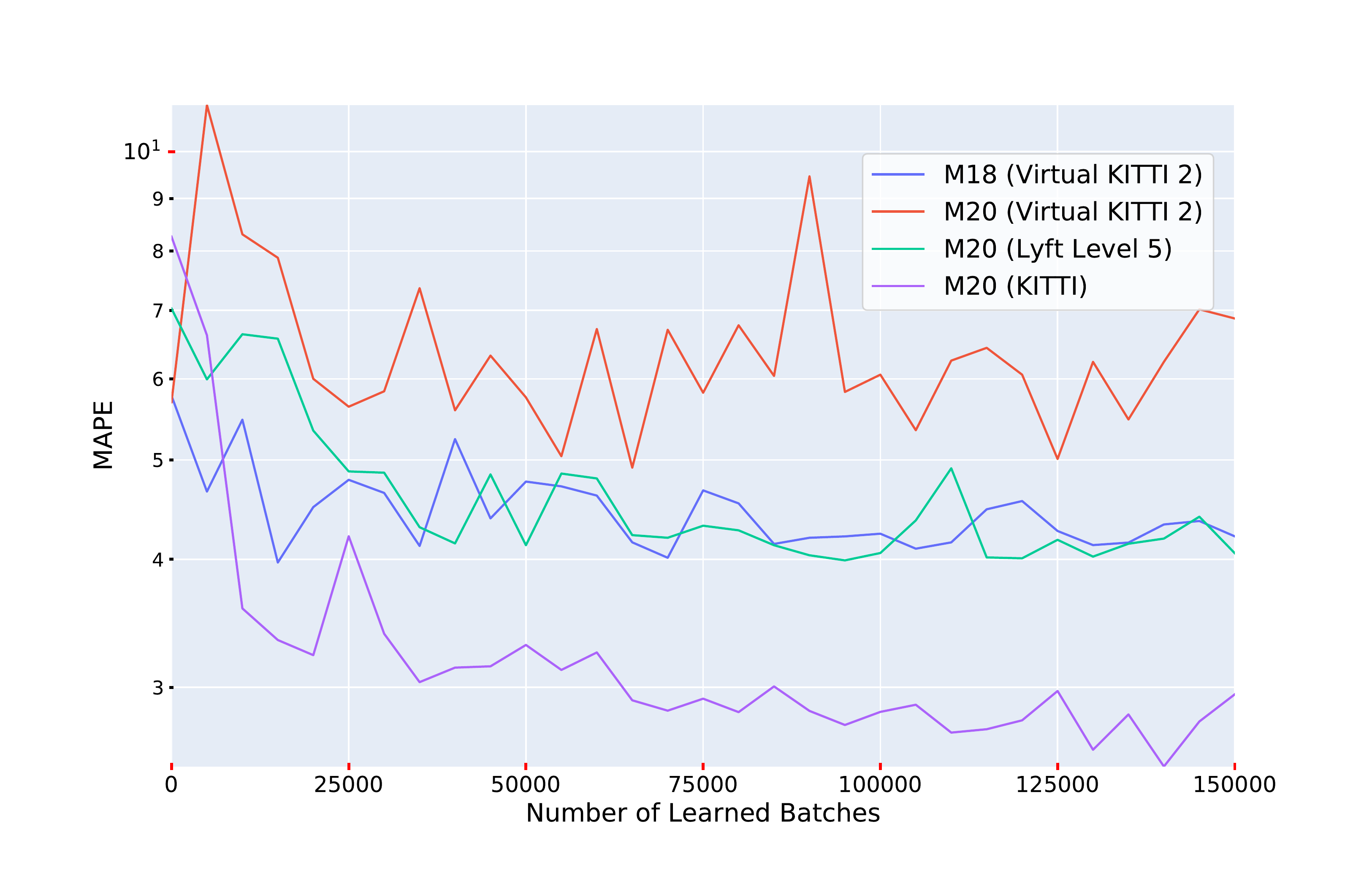}
    \caption[Performance Comparison on the KITTI dataset]{Performance Comparison on the KITTI dataset}
    \label{fig:SSIDE_domain_shift_kitti}
    \end{figure}

In the performance comparison of \gls{SIDE} models that were trained on KITTI, Lyft Level 5, and Virtual KITTI, the M20 network trained on Virtual KITTI has the worst performance (red graphs in \autoref{fig:SSIDE_domain_shift_lyft} and \autoref{fig:SSIDE_domain_shift_kitti}). On the other hand, the best performance is obtained when the training and test dataset were the same, as expected since there is no domain shift in these cases (green graph in \autoref{fig:SSIDE_domain_shift_lyft} and purple graph in \autoref{fig:SSIDE_domain_shift_kitti}).

The most important observation in comparison of \gls{SIDE} models is that the domain shift cause a significant reduction in accuracy when the model is trained on a synthetic dataset compared to when it is trained on a non-synthetic dataset (the red graph compared to purple graph in \autoref{fig:SSIDE_domain_shift_lyft} and the red graph compared to green graph in \autoref{fig:SSIDE_domain_shift_kitti}).

However, the M18 model, as the representative of \gls{SSIDE} networks, performed well compared to the M20 model that was trained on the synthetic Virtual KITTI dataset (blue graph compared to red graph in \autoref{fig:SSIDE_domain_shift_lyft} and \autoref{fig:SSIDE_domain_shift_kitti}). The M18 model outperforms M20 model by 15\% of \gls{MAPE} metric in the KITTI dataset and by 30\% of \gls{MAPE} metric in the Lyft Level 5 dataset. Furthermore, the performance of the M18 network is very close to the performance of the M20 model that is trained on KITTI when the test dataset is from Lyft Level 5 and close to the performance of the M20 model that is trained on the Lyft Level 5 when test dataset is from KITTI. In summary, we conclude that the use of semantic labels in single image depth estimation improves the depth estimation and make the model robust against domain shift while \gls{SIDE} models are more sensitive to domain shift.

    \chapter{Conclusions and Future Work}
\label{ch:discussion}

\section{Conclusions}
In this thesis, we investigated how the use of semantic labels as raw data, like an input image, can potentially improve the performance of a \gls{SIDE} network. We also investigated the effect of the use of semantic labels on the robustness of a \gls{SIDE} model against domain shift.

In the first experimental study (\autoref{sec:firts_exp_results}), we experimented and evaluated various methods to reinforce a pre-designed \gls{SIDE} network with the knowledge of the input image's semantic label. The results showed that by concatenation of the raw image and its semantic label and feeding them both to a well structured network can significantly improve the performance of the depth estimation. From this observations, it is concluded that the use of semantic labels can improve the performance of a \gls{SIDE} network. This result and conclusion are aligned with the assumptions that semantic segments are a prerequisite for the perception of pictorial cues, and furthermore, noting that pictorial cues could be effective in the improvement of a \gls{SIDE} network performance. This conclusion also agrees with the various works that reported on the improvement in the depth estimation by different use of semantic segmentation or semantic segments \cites{peng_wang_towards_2015, ji_joint_2016,jiao_look_2018, ochs_sdnet_2019, zama_ramirez_geometry_2019, zhang_pattern-affinitive_2019, lin_depth_2019, chen_towards_2019, kwak_novel_2020, schneider_semantically_2016, guizilini_semantically-guided_2020, liu_single_2010, jafari_analyzing_2017}.

It is further concluded that the use of a semantic label is most effective when it is fed to a \gls{SIDE} network through the input layer of the network and not through its higher levels. This conclusion theoretically implies that multi-task models, that try to simultaneously estimate the depth and segment the image, are not efficient. This is mainly because this kind of model generates the semantic segments in the last layers.

In the second experimental study, we examined different synthetic datasets and compared them to some non-synthetic datasets in terms of training a segmentation model. The obtained results showed that if a synthetic dataset is more diverse in its content relative to a non-synthetic dataset, then it is superior in training a segmentation network. Therefore, we concluded that regardless of the origin of the dataset (whether it is synthetic or non-synthetic), the more diverse a dataset is in its content, the better performance for the network is obtained. This conclusion lead us to generate semantic labels for non-synthetic images by means of synthetically trained models in the last experimental study.

Finally, in the last experimental study, an \gls{SSIDE} model outperformed its \gls{SIDE} counterpart, while both suffered from domain shift from synthetic data to non-synthetic data. The \gls{SSIDE} model performed like other \gls{SIDE} models that were trained on non-synthetic datasets. From these results, we concluded that the use of semantic segments increases the robustness of the model against domain shifts. Therefore, for the task of depth estimation, if an appropriate synthetic dataset related to the environment of the task would be available, without loosing too much accuracy one can use an \gls{SSIDE} model along with the synthetic datasets to achieve suitable performance.

Unlike other works, in this thesis by isolating the experiments
against altering parameters the following conclusions are obtained.
\begin{itemize}
    \item Use of semantic segments may significantly improve the accuracy of the depth estimation through well structured deep neural networks.
    \item The performance of a depth estimation network improves only when semantic segments are fed to it from the input layer.
    \item The use of semantic segments improves the robustness of a depth estimation model against domain shifts.
    \item The robustness of the model is maximized when it uses semantic edges.
\end{itemize}

These conclusions are derived by the use of ground truths instead of inaccurate predictions for semantic segments. Furthermore the synthetic datasets are analyzed to distinguish the accurate and more reliable synthetic dataset from inaccurate and noisy non-synthetic ones.

Considering all the conclusions reported in this thesis, leads to a practical application; a method that could possibly improve the performance of a \gls{SIDE} model. In this method a \gls{SIDE} network could be converted to an \gls{SSIDE} network only by feeding the semantic edges of the input image.

\section{Future Work}

In this thesis, we used a simple \gls{SIDE} as the base network to develop our models. However, this base network is not very powerful and the developed \gls{SSIDE} networks do not have competitive performance compared to recent state-of-the-art methods. Therefore, a demanded future work could be the examination of the proposed method in this thesis to convert recently proposed \gls{SIDE} networks to \gls{SSIDE} networks.

In previous section we mentioned that multi-task methods that both estimate a depth map and a semantic label are not efficient. Another interesting future research could be to develop a multi-task network based on the M0 architecture, and to compare it with other developed \gls{SSIDE} networks.

In the second experimental study, we observed that a synthetic dataset led to better performance compared to a non-synthetic dataset that had lower diversity in its content. In this experimental study, we only used synthetic datasets that seemed visually with high quality or more realistic, compared to other available synthetic datasets. This leads to the following questions: What would happen if we used low-quality synthetic datasets? Or, if two synthetic and non-synthetic datasets with the same diversity exist, what qualities should the synthetic dataset have to lead to a better performance of a network? These questions could be answered in future work.

In the last experimental study we used generated semantic labels for non-synthetic datasets to evaluate an \gls{SSIDE} network. Definitely, the quality of generated semantic labels affected the performance of the network,. However, its extent is unknown to us. A probable future work related to this issue could be to investigate the relation of the accuracy of semantic labels to the performance of \gls{SSIDE} networks.

    \appendix
    \chapter{Semantic Segmentation}
\label{ch:SemanticSegmentation}
\begin{figure}[h]
    \centering
    \begin{subfigure}{.49\textwidth}
        \centering
        \includegraphics[width=\linewidth]{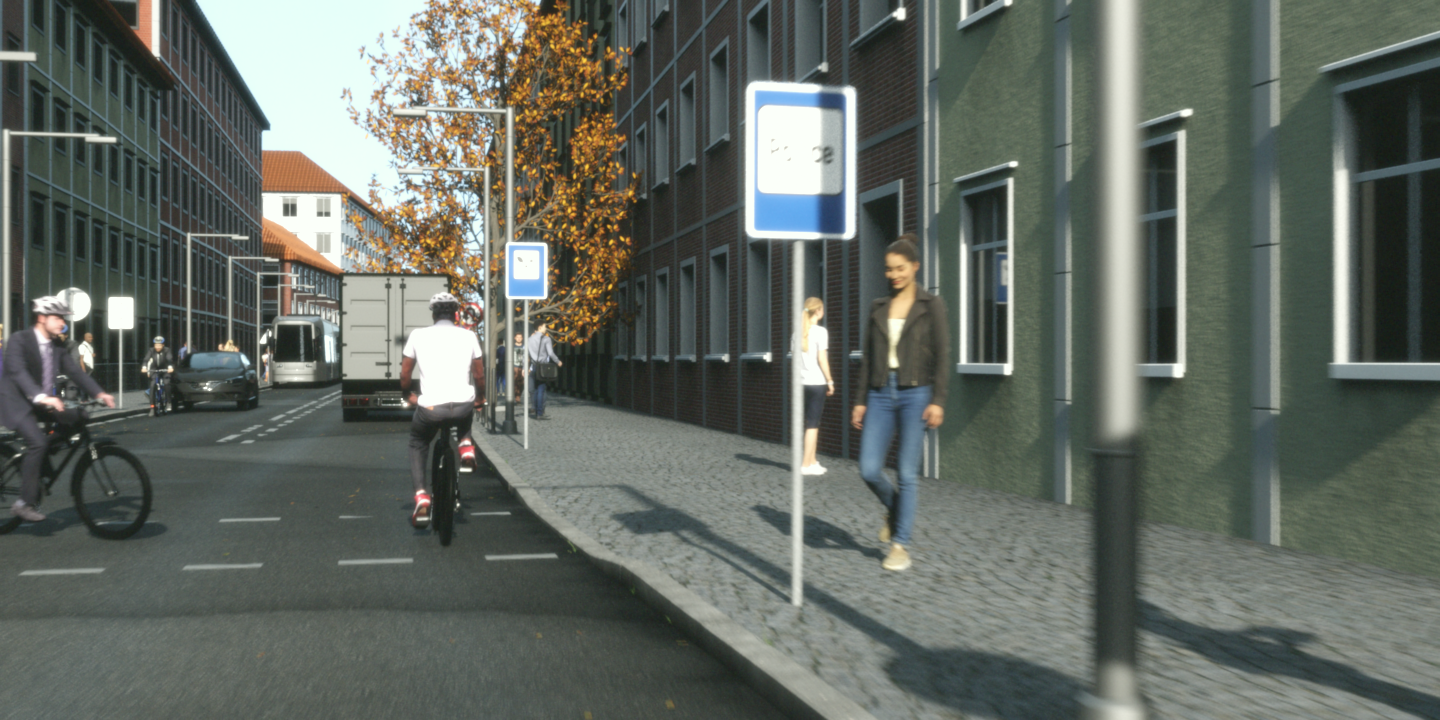} 
        \caption{RGB image}
    \end{subfigure}
    \hfill
    \begin{subfigure}{.49\textwidth}
        \centering
        \includegraphics[width=\linewidth]{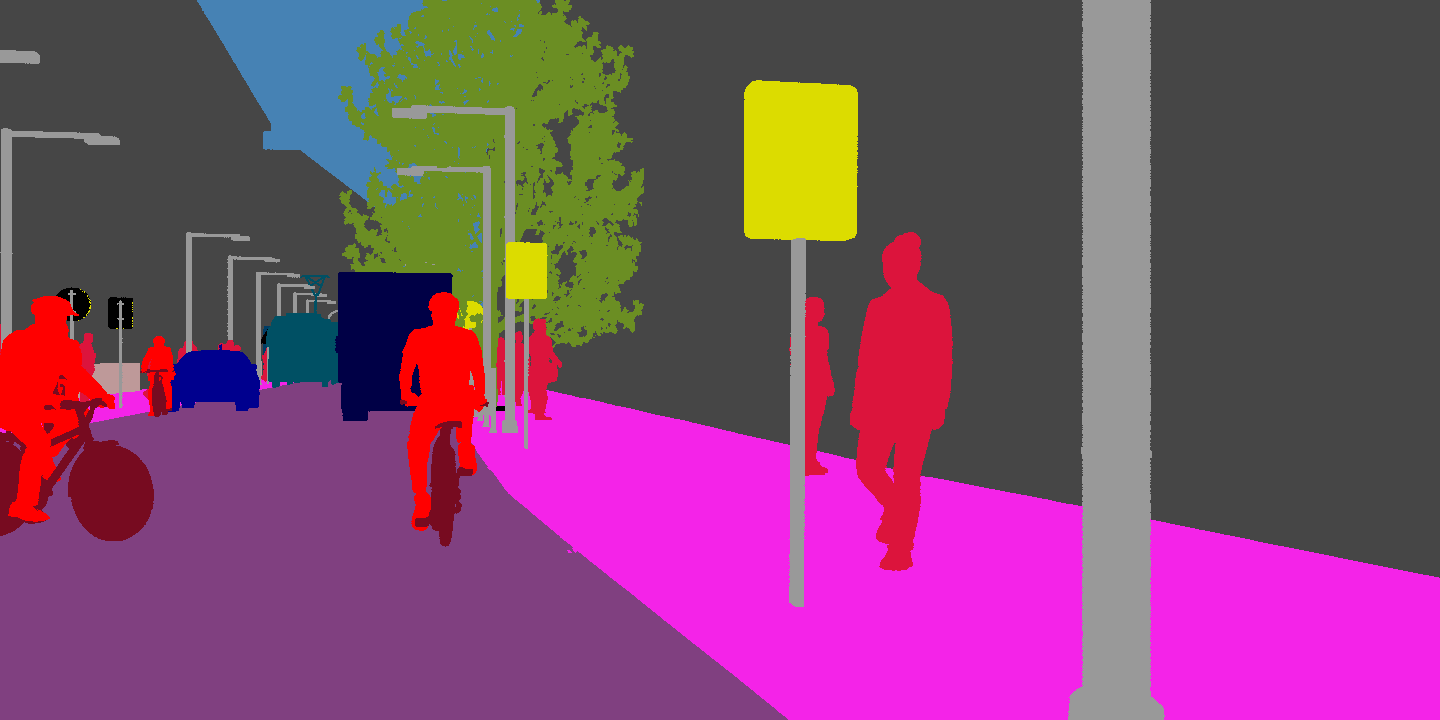} 
        \caption{RGB mapping of semantic segments}
    \end{subfigure}
    \hfill
    \begin{subfigure}{.49\textwidth}
        \centering
        \includegraphics[width=\linewidth]{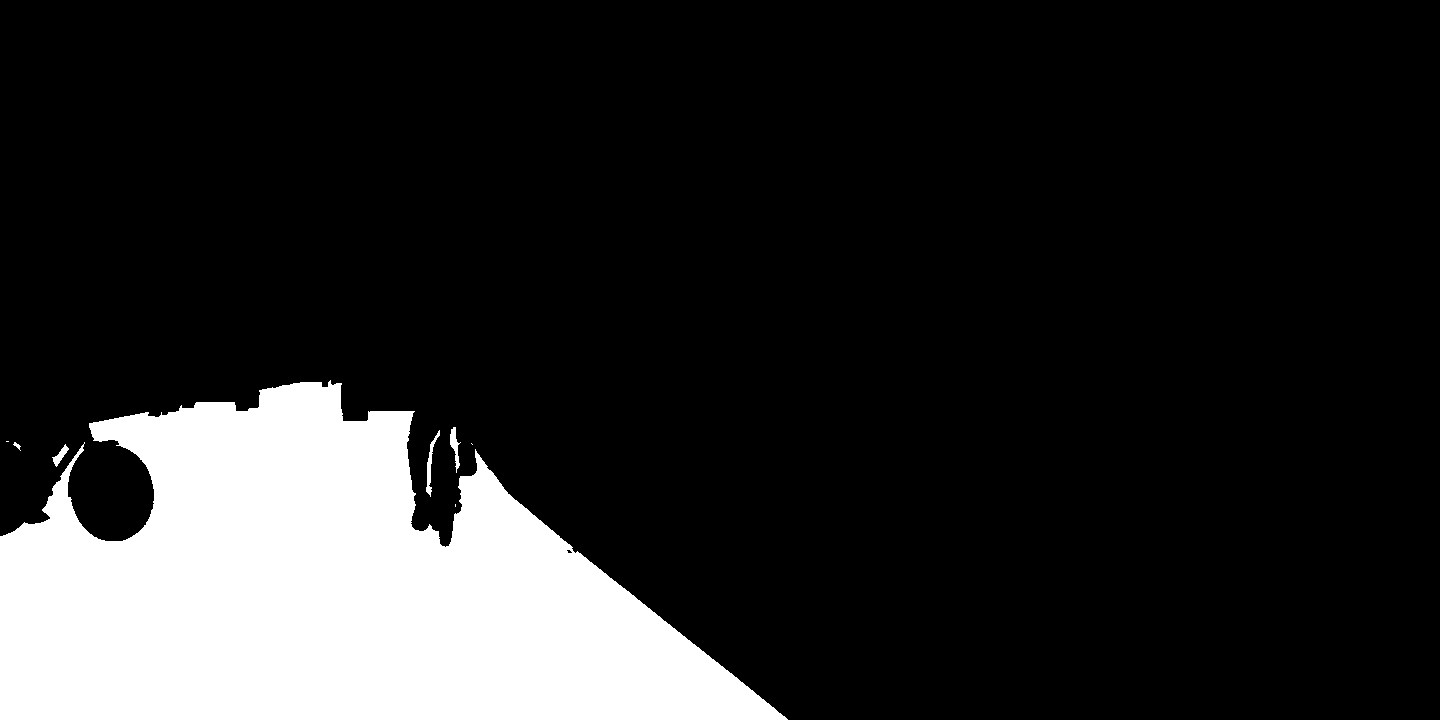} 
        \caption{\small one-hot encoding of \emph{road} class}
    \end{subfigure}
    \hfill
    \begin{subfigure}{.49\textwidth}
        \centering
        \includegraphics[width=\linewidth]{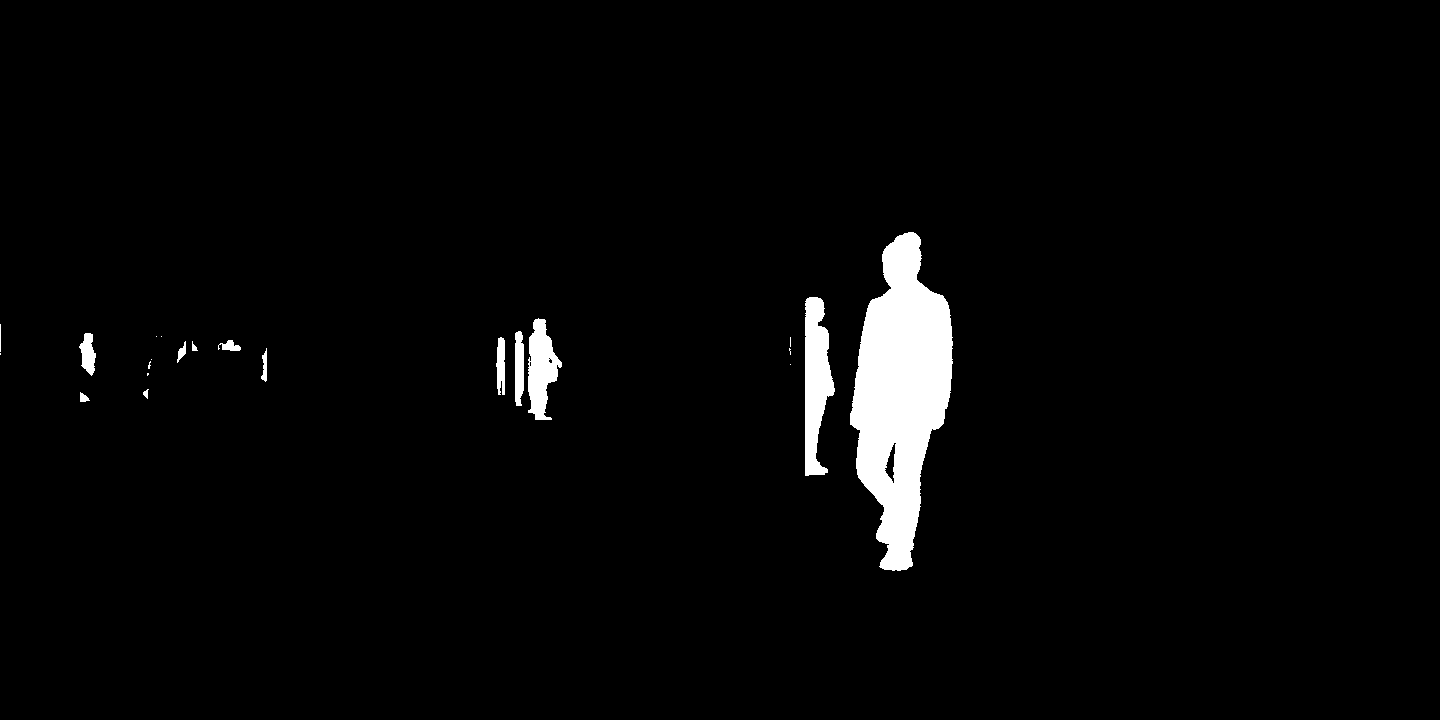} 
        \caption{\small one-hot encoding of \emph{person} class}
    \end{subfigure}
    
    \caption[Different encodings for semantic segments]{Different encodings for semantic segments.An example of an RGB image with its corresponding semantic labels from
        \cite{wrenninge_synscapes_2018}}
    \label{fig:synscapes_example1}
    \end{figure}

Semantic segmentation of an image is to assign a semantic class to every pixel of the image. Usually every semantic class is encoded with an RGB color, so that a semantic segmented image can be visualized in a color image and stored in an image-type format(\autoref{fig:synscapes_example1}). 

\small

    But this type of encoding is not useful for training a \gls{CNN}, because semantic segmentation is a classification task which needs an array of binary numbers as the ground truth for each pixel; 1 shows the existence of a class and 0 shows the absence of a class. Therefore an RGB mapping of semantic segments can be converted to a one-hot labels, i.e.,  a group of binary images (\autoref{fig:synscapes_example1}).

    \chapter{Common Semantic Labels Set}

\label{ch:CommonSemanticLabelsSet}
\begin{table}[h]
    \caption[Common semantic labels set]{Common semantic labels set. Each column shows the available semantic classes in a dataset by mentioning its RGB code. The last column shows a set of unified semantic classes. In tha last column an arrow shows that the class of the row is merged to the class that arrow points to.}
    \resizebox{\textwidth}{!}{%
        \begin{tabular}{clcccccc}
        \hline
         & Class name & VIPER & SYNTHIA-SF & Synscapes & Virtual KITTI & KITTI & Common set \\ \hline
        1 & Fence & [190, 153, 153] & [190, 153, 153] & [190, 153, 153] & - & [190, 153, 153] & $\to$       Unlabeled \\
        2 & Guard Rail & - & - & [180, 165, 180] & [250, 100, 255] & [180, 165, 180] & $\to$ Unlabeled \\
        3 & Wall & - & [102, 102, 156] & [102, 102, 156] & - & [102, 102, 156] & $\to$ building \\
        4 & Parking & - & - & [250, 170, 160] & - & [250, 170, 160] & $\to$ Road \\
        5 & Rail Track & [230, 150, 140] & - & [230, 150, 140] & - & [230, 150, 140] & $\to$ Road \\
        6 & Road & [128, 64, 128] & [128, 64, 128] & [128, 64, 128] & [100, 60, 100] & [128, 64, 128] &         [128, 64, 128] \\
        7 & Sidewalk & [244, 35, 232] & [244, 35, 232] & [244, 35, 232] & - & [244, 35, 232] & [244, 35,         232] \\
        8 & Bridge & - & - & [150, 100, 100] & - & [150, 100, 100] & $\to$ building \\
        9 & Building & [70, 70, 70] & [70, 70, 70] & [70, 70, 70] & [140, 140, 140] & [70, 70, 70] & [70,       70, 70] \\
        10 & Tunnel & - & - & [150, 120, 90] & - & [150, 120, 90] & $\to$ building \\
        11 & Person & [220, 20, 60] & [220, 20, 60] & [220, 20, 60] & - & [220, 20, 60] & [220, 20, 60]      \\
        12 & Bicyclist & [255, 0, 0] & [255, 0, 0] & [255, 0, 0] & - & [255, 0, 0] & $\to$ Person \\
        13 & Lane Marking - General & - & [157, 234, 50] & - & - & - & $\to$ Road \\
        14 & Sky & [70, 130, 180] & [70, 130, 180] & [70, 130, 180] & [90, 200, 255] & [70, 130, 180] &         [70, 130, 180] \\
        15 & Terrain & [152, 251, 152] & [152, 251, 152] & [152, 251, 152] & [210, 0, 200] & [152, 251,         152] & [152, 251, 152] \\
        16 & Vegetation & [35, 142, 35] & [107, 142, 35] & [107, 142, 35] & [90, 240, 0] & [107, 142, 35] &         [107, 142, 35] \\
        17 & Pole & [153, 153, 153] & [153, 153, 153] & [153, 153, 153] & [255, 130, 0] & [153, 153, 153] &         [153, 153, 153] \\
        18 & Traffic Light & [250, 170, 30] & [250, 170, 30] & [250, 170, 30] & [200, 200, 0] & [250, 170,      30] & $\to$ Pole \\
        19 & Traffic Sign (Front) & [220, 220, 0] & [220, 220, 0] & [220, 220, 0] & [255, 255, 0] & [220,       220, 0] & $\to$ Pole \\
        20 & Trash Can & [81, 0, 81] & - & - & - & - & $\to$ Unlabeled \\
        21 & Bicycle & [119, 11, 32] & [119, 11, 32] & [119, 11, 32] & - & [119, 11, 32] & [119, 11, 32]         \\
        22 & Boat & [50, 0, 90] & - & - & - & - & $\to$ Unlabeled \\
        23 & Bus & [0, 60, 100] & [0, 60, 100] & [0, 60, 100] & - & [0, 60, 100] & $\to$ Car \\
        24 & Car & [0, 0, 142] & [0, 0, 142] & [0, 0, 142] & [255, 127, 80] & [0, 0, 142] & [0, 0, 142] \\
        25 & Caravan & [0, 0, 90] & - & [0, 0, 90] & - & [0, 0, 90] & $\to$ Car \\
        26 & Motorcycle & [0, 0, 230] & [0, 0, 230] & [0, 0, 230] & - & [0, 0, 230] & $\to$ Bicycle \\
        27 & On Rails & [0, 80, 100] & [0, 80, 100] & [0, 80, 100] & - & [0, 80, 100] & $\to$ Car \\
        28 & Trailer & - & - & [0, 0, 110] & - & [0, 0, 110] & $\to$ Car \\
        29 & Truck & [0, 0, 70] & [0, 0, 70] & [0, 0, 70] & [160, 60, 60] & [0, 0, 70] & $\to$ Car \\
        30 & Unlabeled & [0, 0, 0] & [0, 0, 0] & [0, 0, 0] & [0, 0, 0] & [0, 0, 0] & [0, 0, 0] \\
        31 & ground & - & - & [81, 0, 81] & - & [81, 0, 81] & $\to$ Unlabeled \\
        32 & dynamic & [111, 74, 0] & - & [111, 74, 0] & - & [111, 74, 0] & $\to$ Unlabeled \\
        33 & plane & [0, 100, 100] & - & - & - & - & $\to$ Unlabeled \\
        34 & trash & [81, 0, 21] & - & - & - & - & $\to$ Unlabeled \\
        35 & chair & [168, 153, 153] & - & - & - & - & $\to$ Unlabeled \\
        36 & firehydrant & [173, 153, 153] & - & - & - & - & $\to$ Unlabeled \\
        37 & mobilebarrier & [180, 180, 100] & - & - & - & - & $\to$ Unlabeled \\
        38 & billboard & [150, 20, 20] & - & - & - & - & $\to$ Unlabeled \\
        39 & tree & [87, 182, 35] & - & - & [0, 199, 0] & - & $\to$ Unlabeled \\
        40 & Misc & - & - & - & [80, 80, 80] & - & $\to$ Unlabeled \\
        % 41 & Van & - & - & - & [0, 139, 139] & - & $\to$ Unlabeled \\ 
        \hline
        \end{tabular}
    }
\end{table}

    \chapter{Experiments Results} 
\label{ch:Experiments_Results}

\section{Supplementary Data for First Experiment}

\begin{figure}[h!]
    \centering
    \begin{subfigure}[b]{.49\textwidth}
        \centering
        \includegraphics[trim={3cm 1cm 1cm 1.5cm}, clip, width=\textwidth]{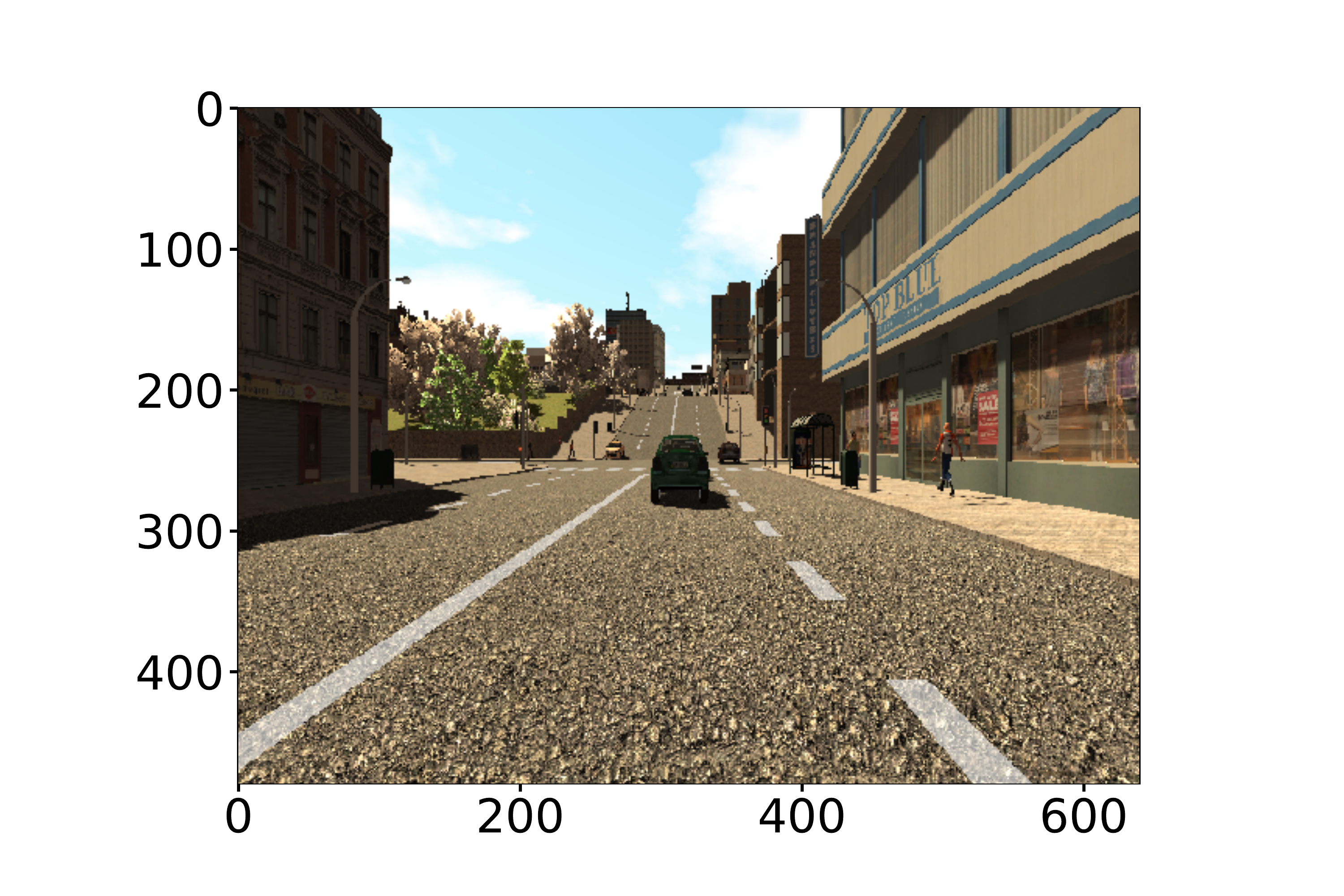} 
        \caption{RGB input image}
    \end{subfigure}
    \hfill
    \begin{subfigure}[b]{.49\textwidth}
        \centering
        \includegraphics[trim={1.5cm 1.5cm 3.5cm 1.5cm}, clip, width=\textwidth]{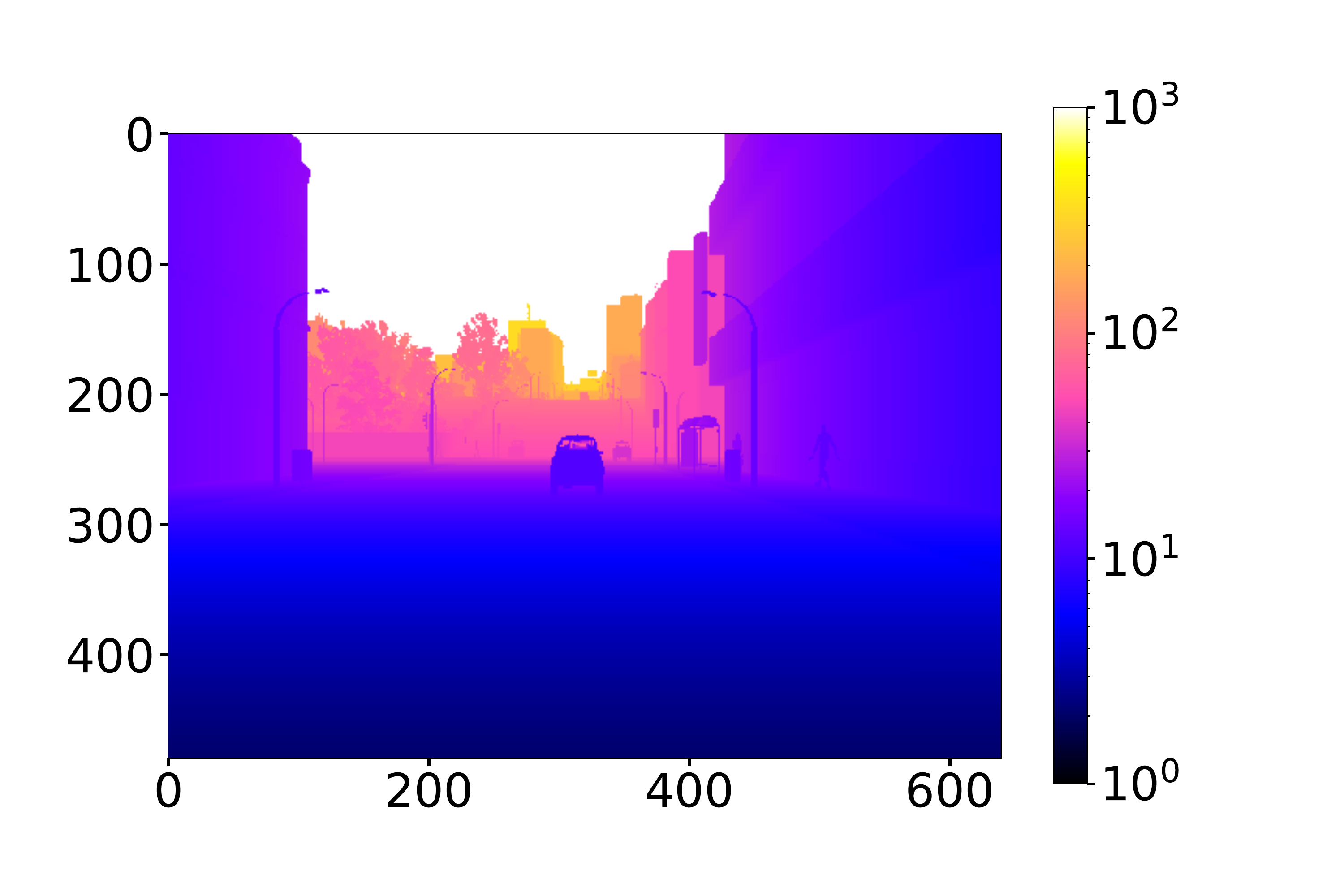} 
        \caption{Depth ground truth }
    \end{subfigure}
    \hfill
    \begin{subfigure}[b]{.49\textwidth}
        \centering
        \includegraphics[trim={1.5cm 1.5cm 3.5cm 1.5cm}, clip, width=\textwidth]{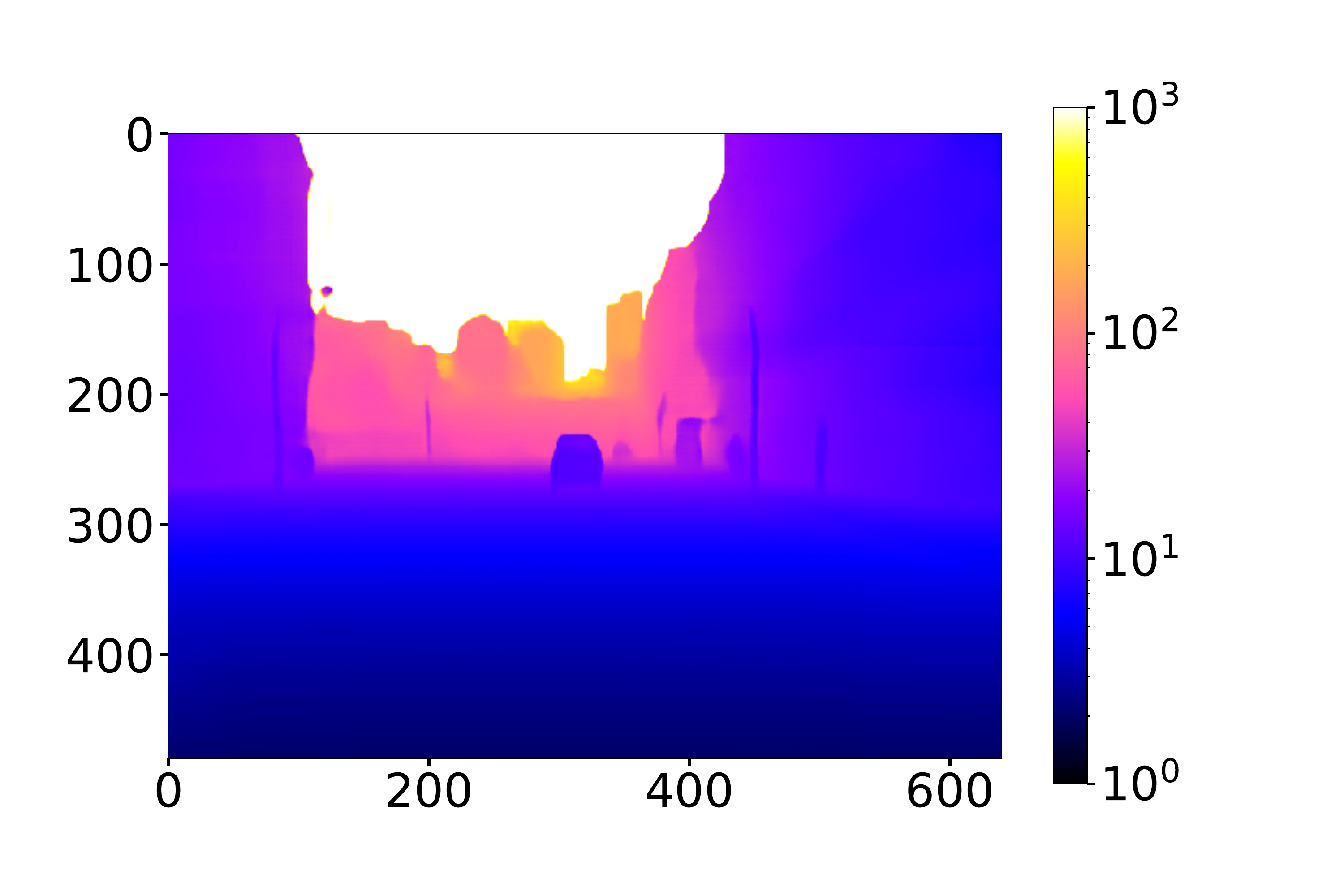} 
        \caption{Depth estimation by M18}
    \end{subfigure}
    \hfill
    \begin{subfigure}[b]{.49\textwidth}
        \centering
        \includegraphics[trim={1.5cm 1.5cm 3.5cm 1.5cm}, clip, width=\textwidth]{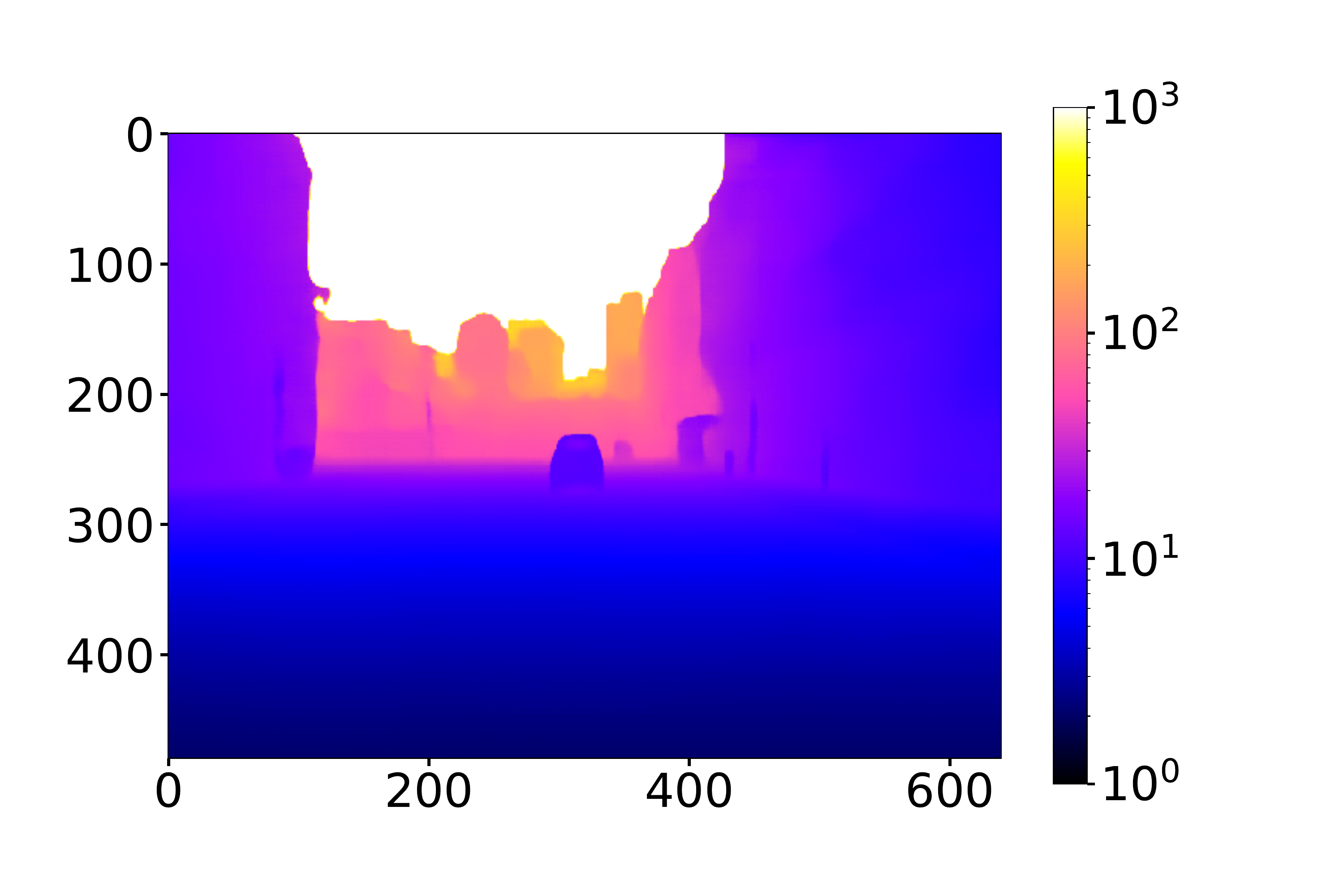} 
        \caption{Depth estimation by M20}
    \end{subfigure}
    
    \caption[Depth estimation samples of M18 and M20]{Depth estimation samples of M18 and M20}
    \label{fig:DepthEstimationSamples}
    \end{figure}

\section{Supplementary Data for Second Experiment}

% \subsection{Training Dataset: Cityscapes}
\begin{figure}[h]
    \centering
    \includegraphics[trim={0 0 0 1cm},clip,width=0.9\linewidth]{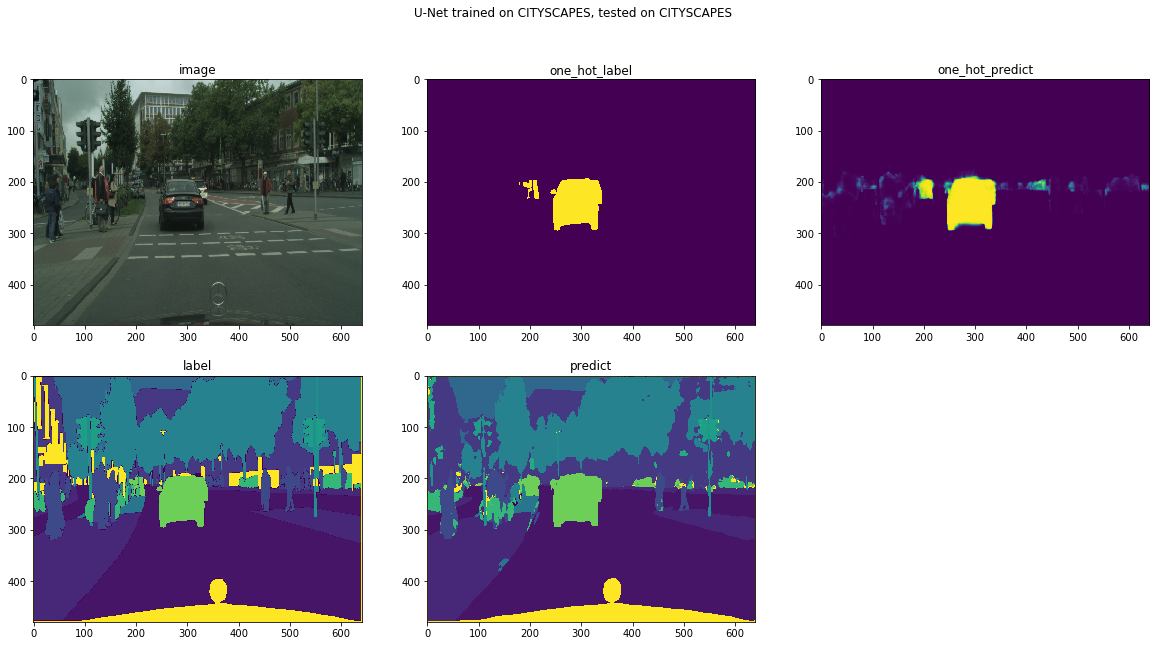}
    \caption[Perception sample of Cityscapes U-Net on \emph{Cityscapes}]{Perception sample of Cityscapes U-Net on \emph{Cityscapes}. The top right image is the input RGB image, the other two top images are ground truth and prediction in a single semantic class and two bottom images are overall ground truth and prediction.}
    \label{fig:visual_Cityscapes_Cityscapes}
    \end{figure}

\begin{figure}
    \centering
    \includegraphics[trim={0 0 0 1cm},clip,width=0.9\linewidth]{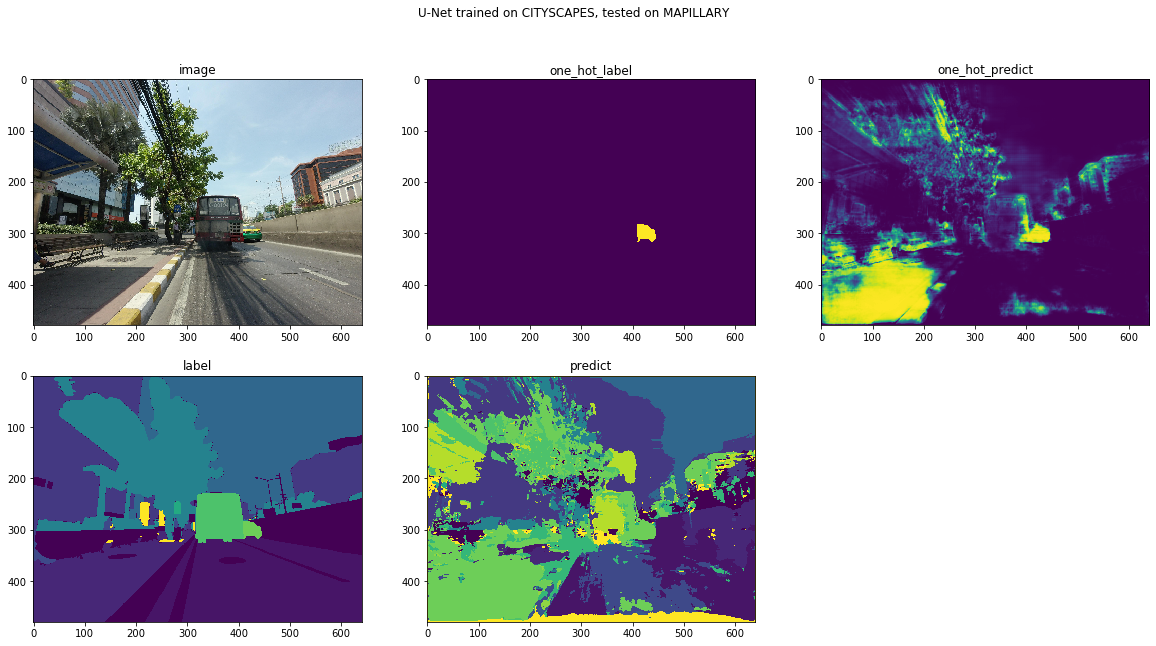}
    \caption[Perception sample of Cityscapes U-Net on \emph{Mapillary}]{Perception sample of Cityscapes U-Net on \emph{Mapillary}. The top right image is the input RGB image, the other two top images are ground truth and prediction in a single semantic class and two bottom images are overall ground truth and prediction.}
    \label{fig:visual_Cityscapes_Mapillary}
    \end{figure}

\begin{figure}
    \centering
    \includegraphics[trim={0 0 0 1cm},clip,width=0.9\linewidth]{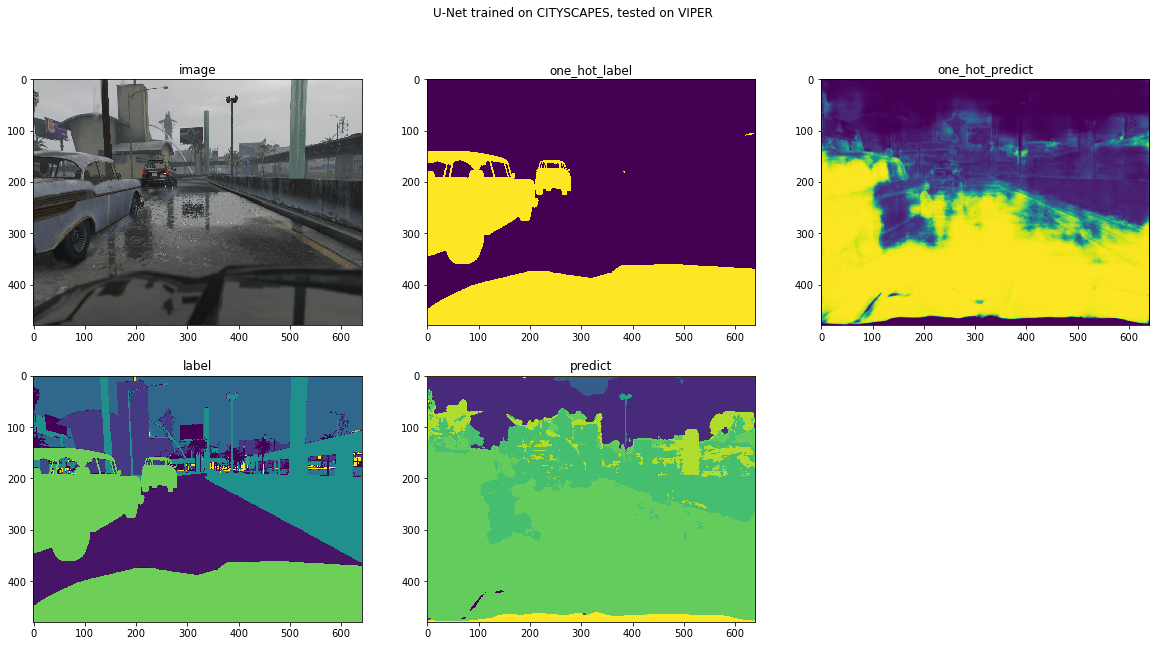}
    \caption[Perception sample of Cityscapes U-Net on \emph{VIPER}]{Perception sample of Cityscapes U-Net on \emph{VIPER}. The top right image is the input RGB image, the other two top images are ground truth and prediction in a single semantic class and two bottom images are overall ground truth and prediction.}
    \label{fig:visual_Cityscapes_VIPER}
    \end{figure}

\begin{figure}
    \centering
    \includegraphics[trim={0 0 0 1cm},clip,width=0.9\linewidth]{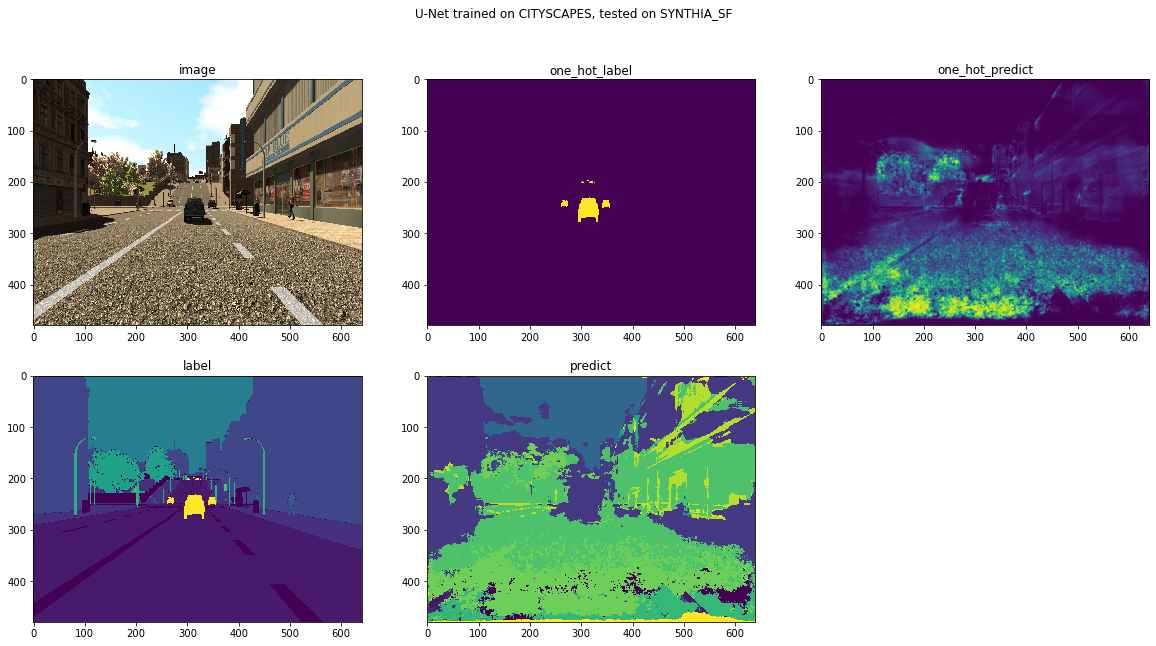}
    \caption[Perception sample of Cityscapes U-Net on \emph{SYNTHIA-SF}]{Perception sample of Cityscapes U-Net on \emph{SYNTHIA-SF}. The top right image is the input RGB image, the other two top images are ground truth and prediction in a single semantic class and two bottom images are overall ground truth and prediction.}
    \label{fig:visual_Cityscapes_SYNTHIA-SF}
    \end{figure}

% \subsection{Training Dataset: Mapillary}
\begin{figure}
    \centering
    \includegraphics[trim={0 0 0 1cm},clip,width=0.9\linewidth]{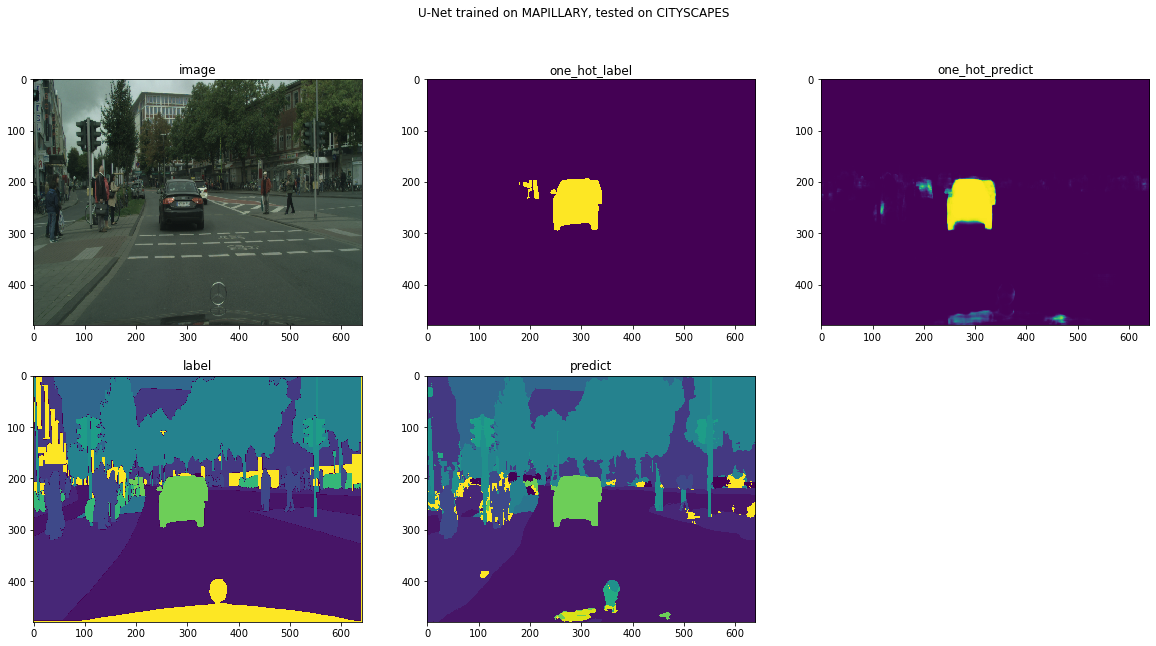}
    \caption[Perception sample of Mapillary U-Net on \emph{Cityscapes}]{Perception sample of Mapillary U-Net on \emph{Cityscapes}. The top right image is the input RGB image, the other two top images are ground truth and prediction in a single semantic class and two bottom images are overall ground truth and prediction.}
    \label{fig:visual_Mapillary_Cityscapes}
    \end{figure}

\begin{figure}
    \centering
    \includegraphics[trim={0 0 0 1cm},clip,width=0.9\linewidth]{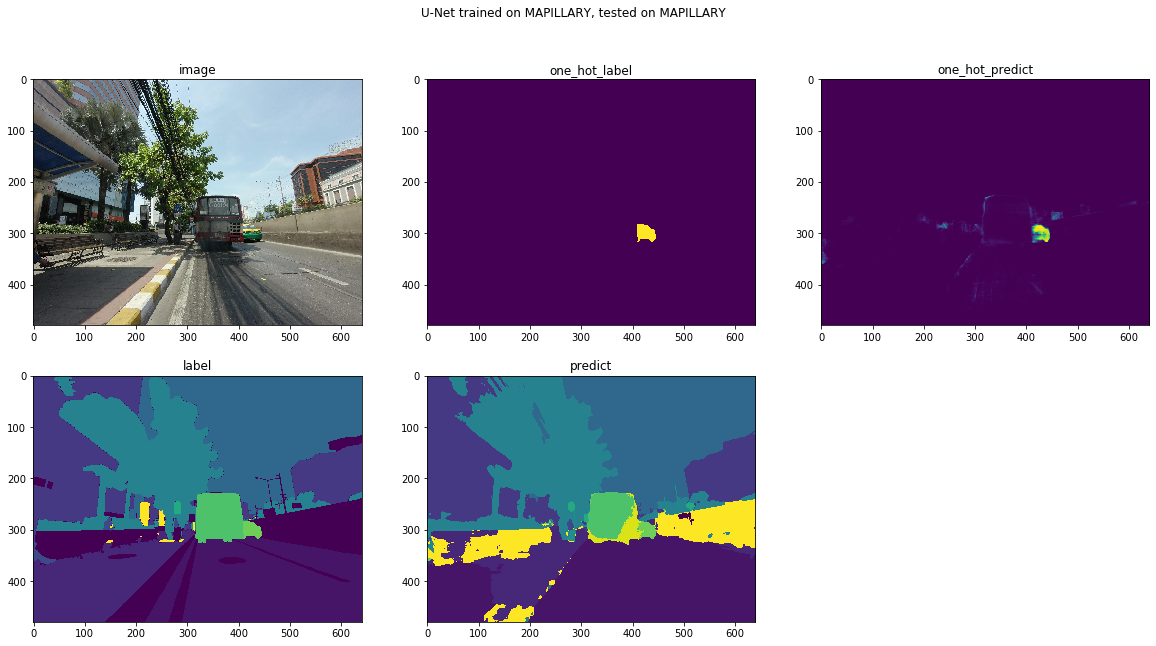}
    \caption[Perception sample of Mapillary U-Net on \emph{Mapillary}]{Perception sample of Mapillary U-Net on \emph{Mapillary}. The top right image is the input RGB image, the other two top images are ground truth and prediction in a single semantic class and two bottom images are overall ground truth and prediction.}
    \label{fig:visual_Mapillary_Mapillary}
    \end{figure}

\begin{figure}
    \centering
    \includegraphics[trim={0 0 0 1cm},clip,width=0.9\linewidth]{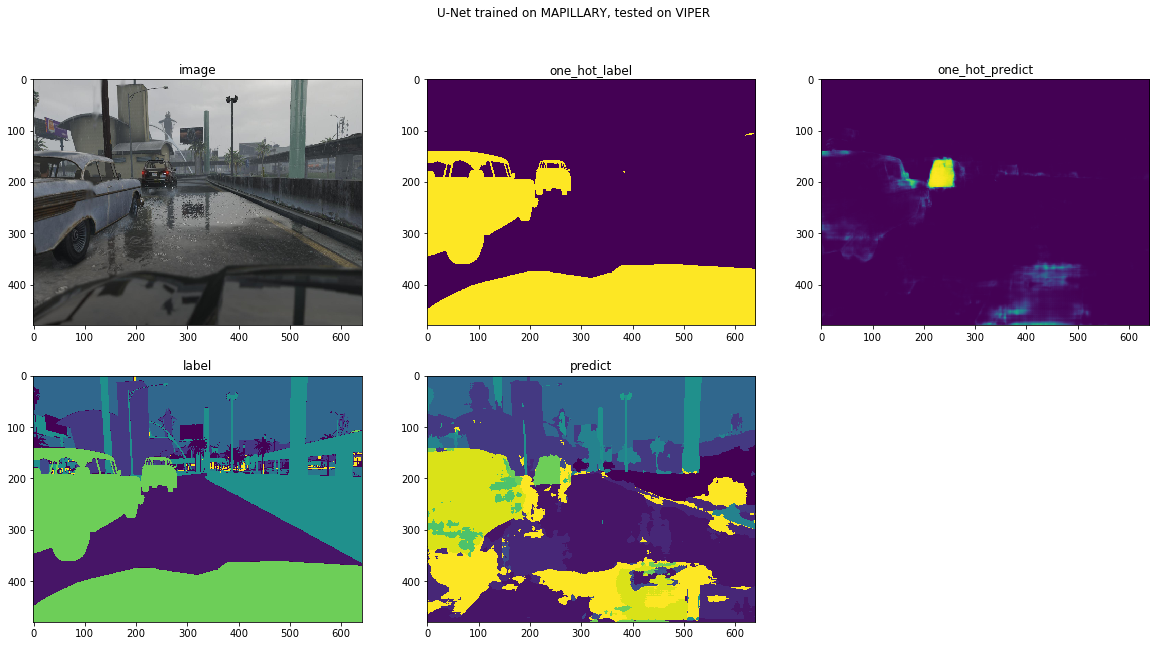}
    \caption[Perception sample of Mapillary U-Net on \emph{VIPER}]{Perception sample of Mapillary U-Net on \emph{VIPER}. The top right image is the input RGB image, the other two top images are ground truth and prediction in a single semantic class and two bottom images are overall ground truth and prediction.}
    \label{fig:visual_Mapillary_VIPER}
    \end{figure}

\begin{figure}
    \centering
    \includegraphics[trim={0 0 0 1cm},clip,width=0.9\linewidth]{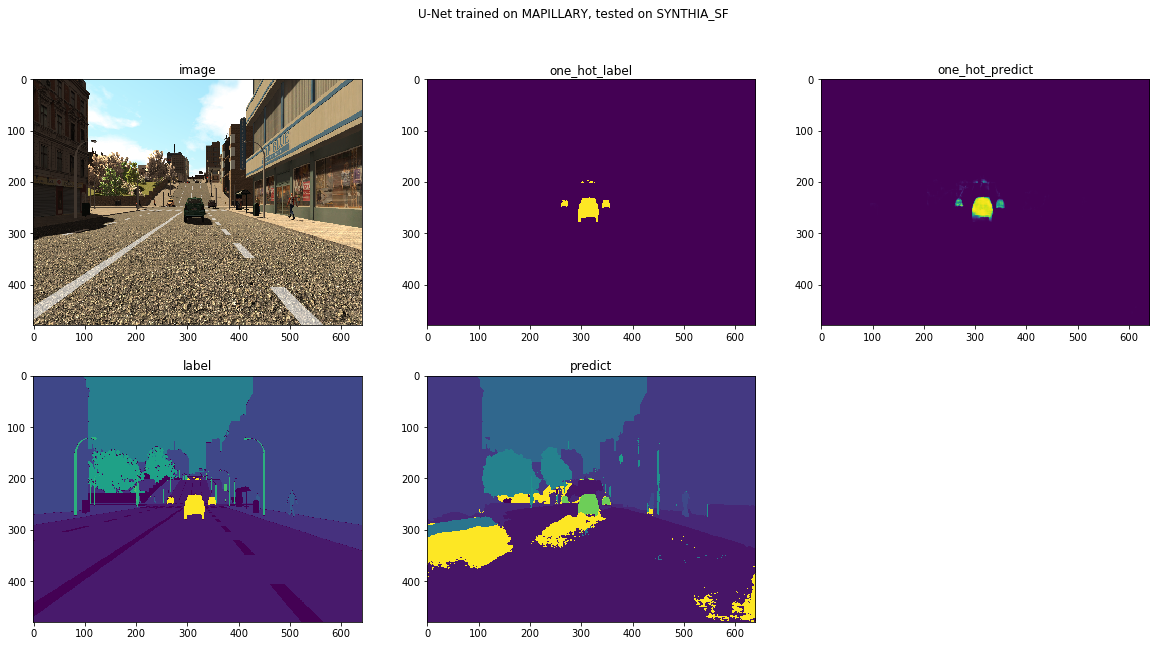}
    \caption[Perception sample of Mapillary U-Net on \emph{SYNTHIA-SF}]{Perception sample of Mapillary U-Net on \emph{SYNTHIA-SF}. The top right image is the input RGB image, the other two top images are ground truth and prediction in a single semantic class and two bottom images are overall ground truth and prediction.}
    \label{fig:visual_Mapillary_SYNTHIA-SF}
    \end{figure}

% \subsection{Training Dataset: SYNTHIA-SF}
\begin{figure}
    \centering
    \includegraphics[trim={0 0 0 1cm},clip,width=0.9\linewidth]{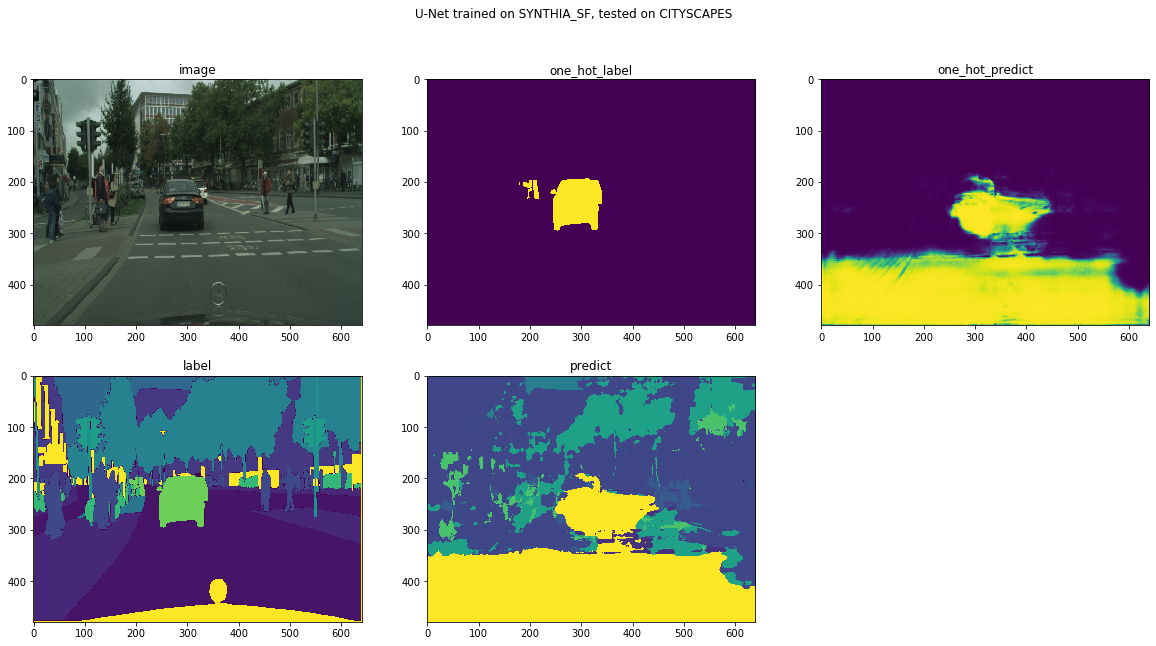}
    \caption[Perception sample of SYNTHIA-SF U-Net on \emph{Cityscapes}]{Perception sample of SYNTHIA-SF U-Net on \emph{Cityscapes}. The top right image is the input RGB image, the other two top images are ground truth and prediction in a single semantic class and two bottom images are overall ground truth and prediction.}
    \label{fig:visual_SYNTHIA-SF_Cityscapes}
    \end{figure}

\begin{figure}
    \centering
    \includegraphics[trim={0 0 0 1cm},clip,width=0.9\linewidth]{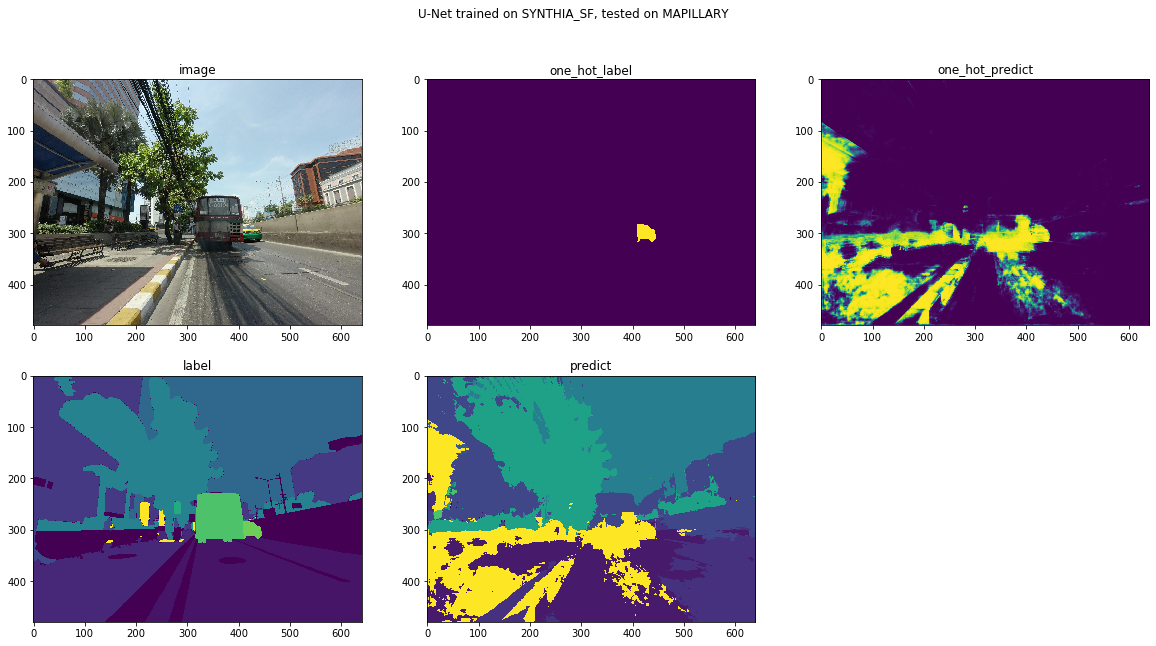}
    \caption[Perception sample of SYNTHIA-SF U-Net on \emph{Mapillary}]{Perception sample of SYNTHIA-SF U-Net on \emph{Mapillary}. The top right image is the input RGB image, the other two top images are ground truth and prediction in a single semantic class and two bottom images are overall ground truth and prediction.}
    \label{fig:visual_SYNTHIA-SF_Mapillary}
    \end{figure}

\begin{figure}
    \centering
    \includegraphics[trim={0 0 0 1cm},clip,width=0.9\linewidth]{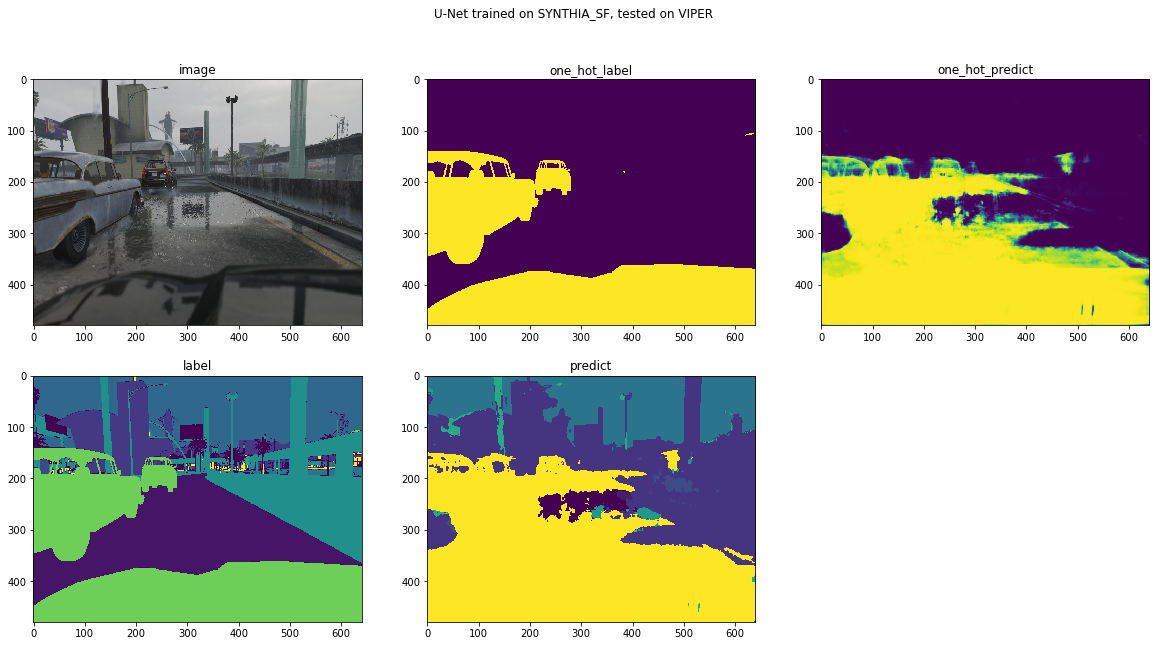}
    \caption[Perception sample of SYNTHIA-SF U-Net on \emph{VIPER}]{Perception sample of SYNTHIA-SF U-Net on \emph{VIPER}. The top right image is the input RGB image, the other two top images are ground truth and prediction in a single semantic class and two bottom images are overall ground truth and prediction.}
    \label{fig:visual_SYNTHIA-SF_VIPER}
    \end{figure}

\begin{figure}
    \centering
    \includegraphics[trim={0 0 0 1cm},clip,width=0.9\linewidth]{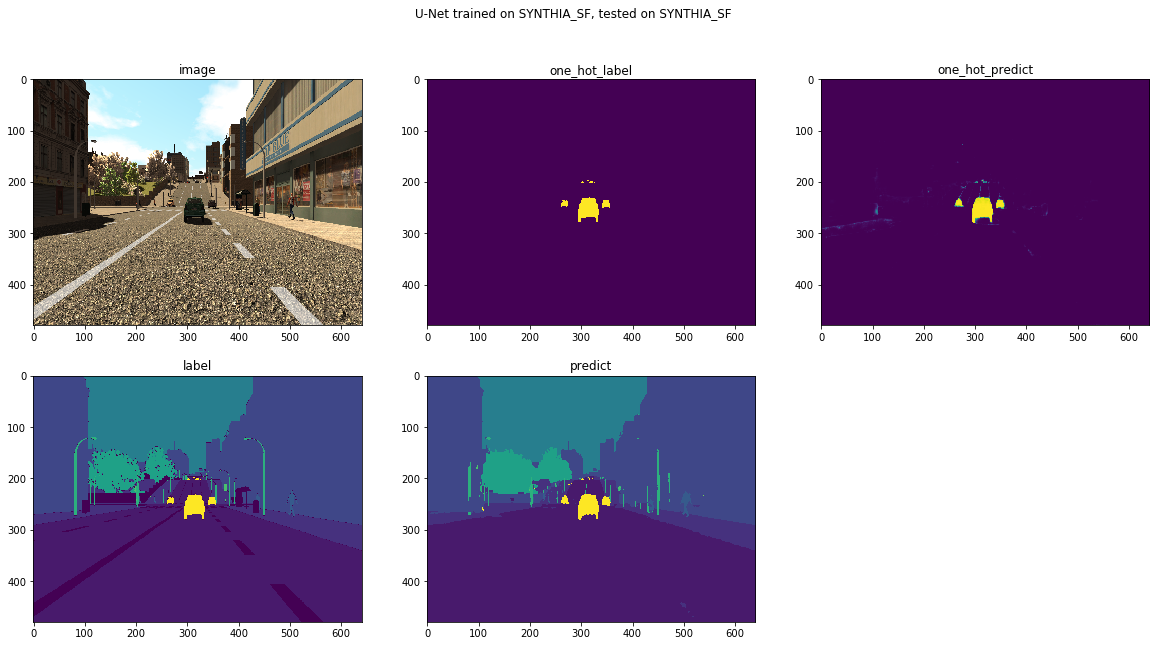}
    \caption[Perception sample of SYNTHIA-SF U-Net on \emph{SYNTHIA-SF}]{Perception sample of SYNTHIA-SF U-Net on \emph{SYNTHIA-SF}. The top right image is the input RGB image, the other two top images are ground truth and prediction in a single semantic class and two bottom images are overall ground truth and prediction.}
    \label{fig:visual_SYNTHIA-SF_SYNTHIA-SF}
    \end{figure}

% \subsection{Training Dataset: VIPER}
\begin{figure}
    \centering
    \includegraphics[trim={0 0 0 1cm},clip,width=0.9\linewidth]{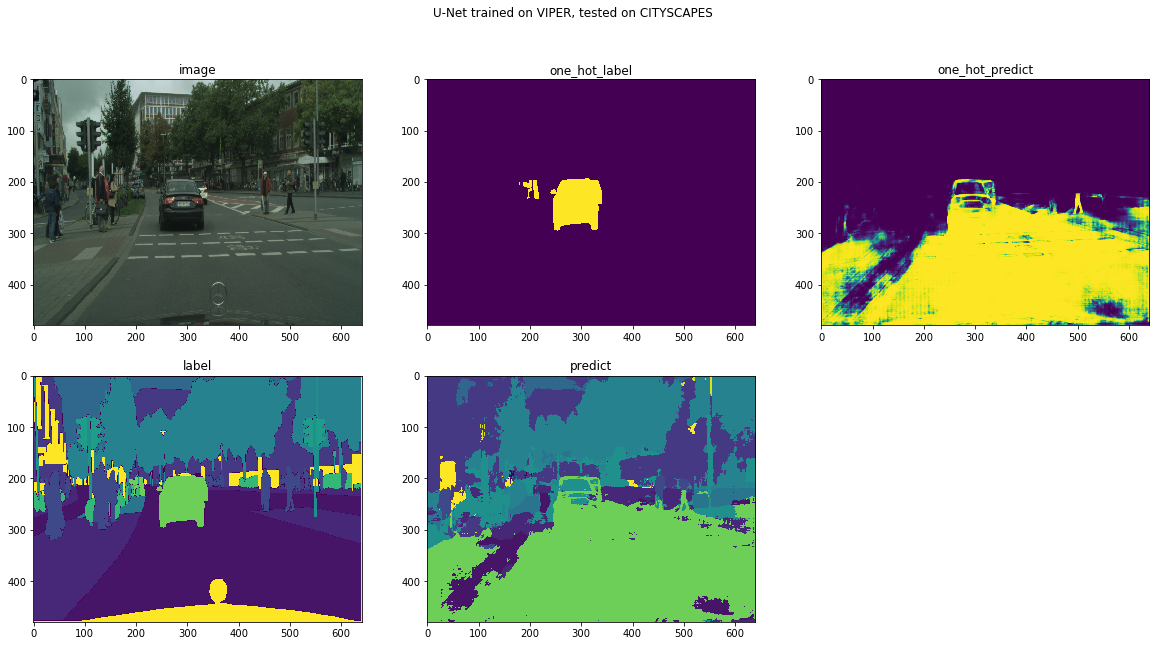}
    \caption[Perception sample of VIPER U-Net on \emph{Cityscapes}]{Perception sample of VIPER U-Net on \emph{Cityscapes}. The top right image is the input RGB image, the other two top images are ground truth and prediction in a single semantic class and two bottom images are overall ground truth and prediction.}
    \label{fig:visual_VIPER_Cityscapes}
    \end{figure}

\begin{figure}
    \centering
    \includegraphics[trim={0 0 0 1cm},clip,width=0.9\linewidth]{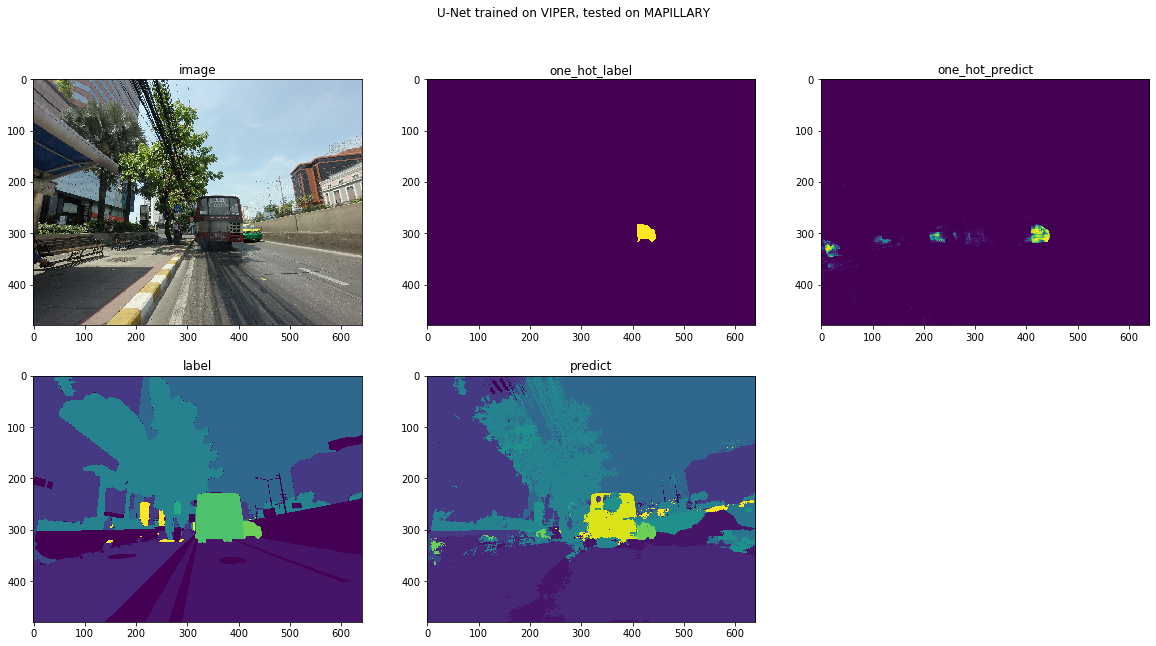}
    \caption[Perception sample of VIPER U-Net on \emph{Mapillary}]{Perception sample of VIPER U-Net on \emph{Mapillary}. The top right image is the input RGB image, the other two top images are ground truth and prediction in a single semantic class and two bottom images are overall ground truth and prediction.}
    \label{fig:visual_VIPER_Mapillary}
    \end{figure}

\begin{figure}
    \centering
    \includegraphics[trim={0 0 0 1cm},clip,width=0.9\linewidth]{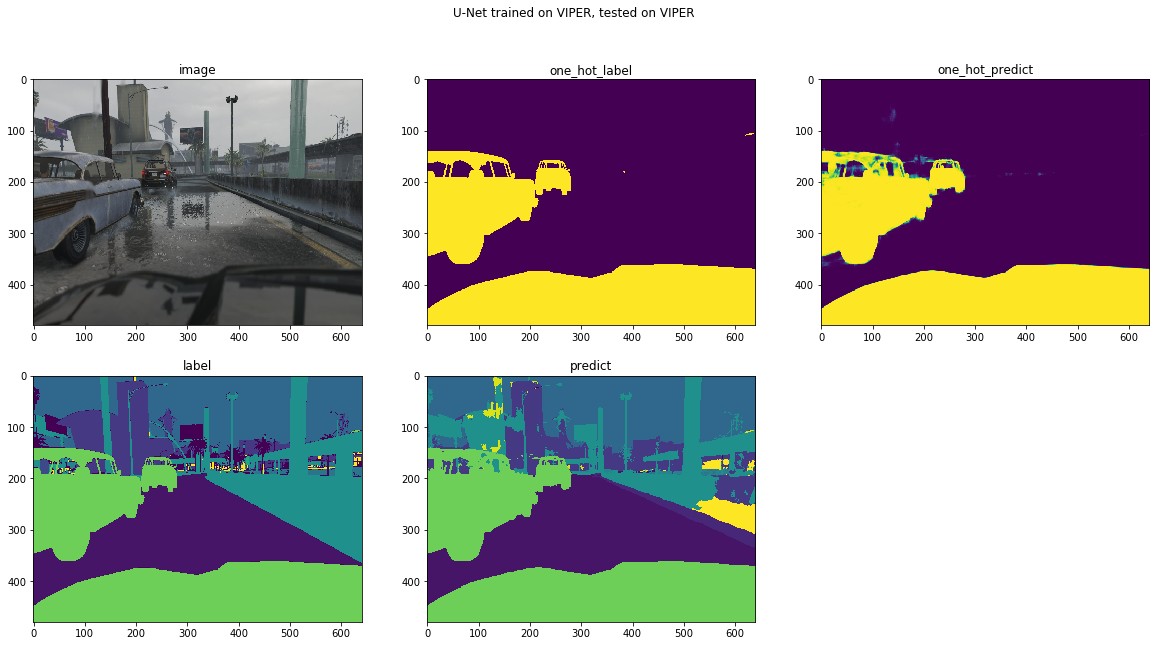}
    \caption[Perception sample of VIPER U-Net on \emph{VIPER}]{Perception sample of VIPER U-Net on \emph{VIPER}. The top right image is the input RGB image, the other two top images are ground truth and prediction in a single semantic class and two bottom images are overall ground truth and prediction.}
    \label{fig:visual_VIPER_VIPER}
    \end{figure}

\begin{figure}
    \centering
    \includegraphics[trim={0 0 0 1cm},clip,width=0.9\linewidth]{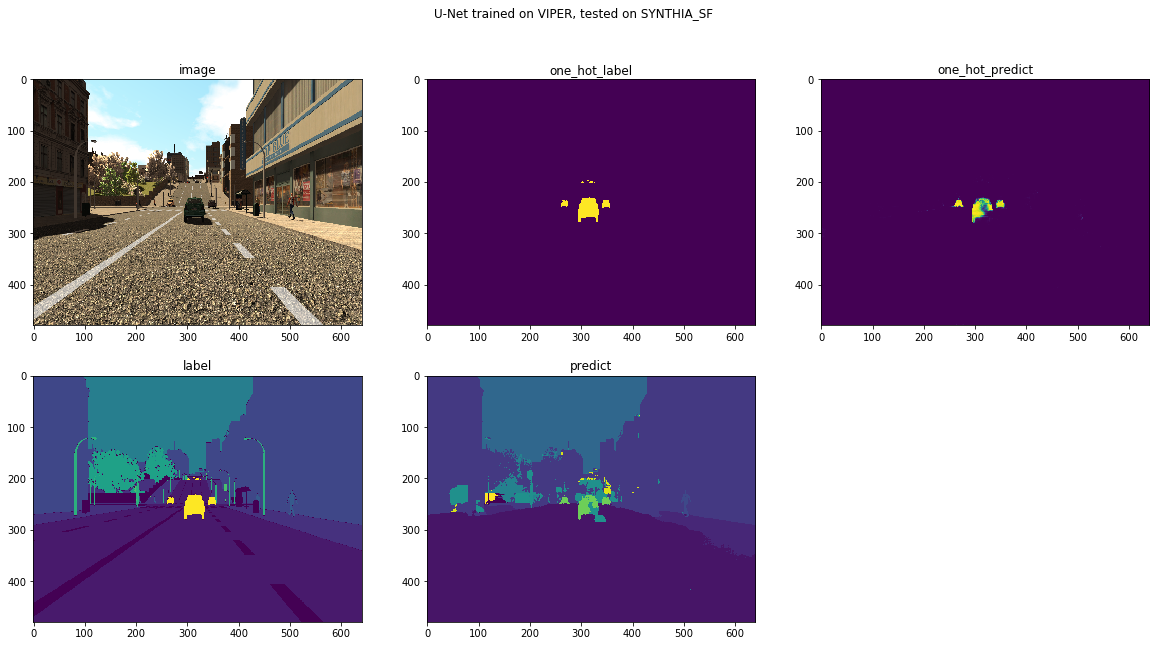}
    \caption[Perception sample of VIPER U-Net on \emph{SYNTHIA-SF}]{Perception sample of VIPER U-Net on \emph{SYNTHIA-SF}. The top right image is the input RGB image, the other two top images are ground truth and prediction in a single semantic class and two bottom images are overall ground truth and prediction.}
    \label{fig:visual_VIPER_SYNTHIA-SF}
    \end{figure}

%% Bibliography
    \printbibliography[heading=bibintoc, title={Bibliography}]

\end{document}